\documentclass[letterpaper]{article} 
\usepackage{aaai25}  
\usepackage{times}  
\usepackage{helvet}  
\usepackage{courier}  
\usepackage[hyphens]{url}  
\usepackage{graphicx} 
\usepackage{natbib}  
\usepackage{caption} 
\usepackage{algorithm}
\usepackage{algorithmic}

\usepackage{newfloat}
\usepackage{listings}
\DeclareCaptionStyle{ruled}{labelfont=normalfont,labelsep=colon,strut=off} 
\floatstyle{ruled}
\newfloat{listing}{tb}{lst}{}
\floatname{listing}{Listing}
\newcommand{\vct}[1]{\bm{#1}}
\newcommand{\x}{\vct{x}}

\usepackage{amsmath,amsfonts,amssymb,bm,subcaption,multirow}
\usepackage{xspace}
\newcommand{\name}{\emph{TimePFN}\xspace}

\title{TimePFN: Effective Multivariate Time Series Forecasting with Synthetic Data}
\author{
    Ege Onur Taga,
    M. Emrullah Ildiz,
    Samet Oymak
}
\affiliations{

    University of Michigan, Ann Arbor\\
    egetaga@umich.edu, eildiz@umich.edu, oymak@umich.edu  \\
}

\begin{document}

\maketitle

\begin{abstract}
    The diversity of time series applications and scarcity of domain-specific data highlight the need for time-series models with strong few-shot learning capabilities.
    In this work, we propose a novel training scheme and a transformer-based architecture, collectively referred to as \name,
    for multivariate time-series (MTS) forecasting. \name is based on the concept of Prior-data Fitted Networks (PFN),
    which aims to approximate Bayesian inference. Our approach consists of (1) generating synthetic MTS data through diverse Gaussian process kernels and the linear coregionalization method, and (2) a novel MTS architecture capable of utilizing both 
    temporal and cross-channel dependencies across all input patches. We evaluate \name on several benchmark datasets and demonstrate that
    it outperforms the existing state-of-the-art models for MTS forecasting in both zero-shot and few-shot settings. Notably, fine-tuning \name with as few as 500 data points nearly matches full dataset training error, and even 50 data points yield competitive results. We also find that \name exhibits strong univariate forecasting performance, attesting to its generalization ability. Overall, this work unlocks the power of synthetic data priors for MTS forecasting and facilitates strong zero- and few-shot forecasting performance.
\end{abstract}


%
\begin{links}
      \link{Code}{https://github.com/egetaga/TimePFN}
\end{links}

\section{Introduction}
\label{sec:intro}

Natural language processing has achieved remarkable success driven by advances in neural architectures and data pipelines. These advances underlie modern language and vision-language models that exhibit remarkable zero-shot and few-shot learning capabilities. Inspired by these, researchers have started exploring whether such methods and ideas could be extended to time series forecasting. For instance, a notable line of work \cite{haoyietal-informer-2021, wu2021autoformer, zhou2022fedformer, zhang2023crossformer} examine the use of transformer architecture \cite{attention_is_all_you_need} in time-series forecasting. More recently, there is also a push toward building foundation models for time series tasks \cite{ansari2024chronos}. However, the heterogeneous nature of time series data brings additional complications. As shown by \cite{Zeng2022AreTE}, even simple linear models are shown to outperform most existing transformer-based models in univariate and multivariate time-series forecasting. This could be attributed to the heterogeneous nature of time-series data and relatively naive tokenization methods, underscoring the need for richer datasets as well as more effective architectures that can capture both temporal and cross-channel dependencies.

\begin{figure}[t!] 
   \centering    \includegraphics[width= 0.47\textwidth]{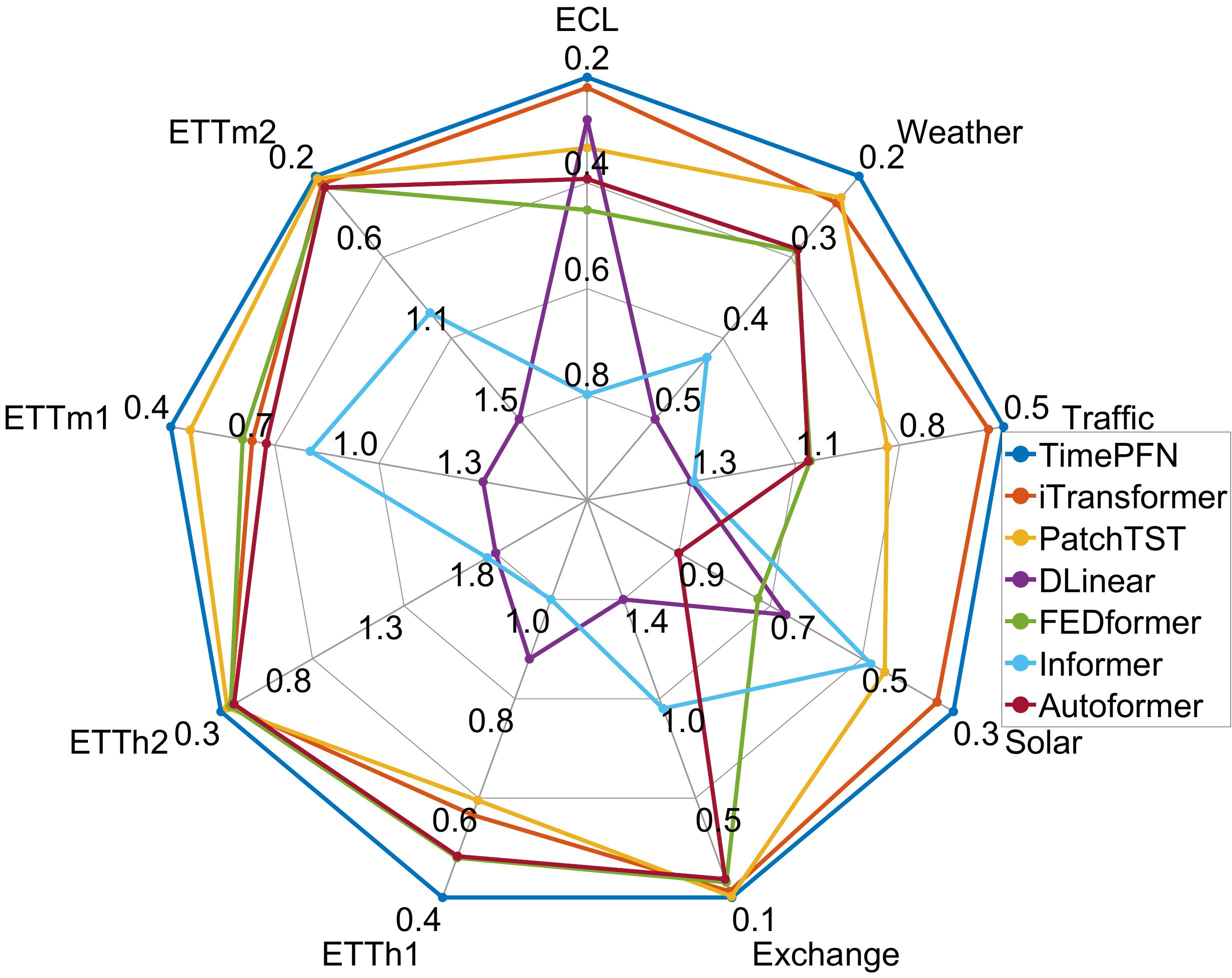}
   \caption{Average forecasting performance (MSE) of TimePFN. MSE values are averaged over all data budgets.}
\end{figure}

In language models, the discrete nature of the problem makes the tokenization fairly straightforward, which is in contrast to the continuous time series data. Additionally, the scalar value of a time series datapoint have no clear meaning, unlike words, where vector embeddings can capture semantic similarity. To address these problems, PatchTST \cite{Yuqietal-2023-PatchTST} proposed using patching with overlapping strides and demonstrated its benefit for univariate forecasting. While PatchTST treats multivariate forecasting as multiple univariate problems, iTransformer \cite{liu2023itransformer} proposes representing each channel as a single token, resulting in an architecture that intuitively augments simple linear layers with a transformer architecture.  

In this work, we approach MTS forecasting from a data-centric perspective. While various architectural considerations have been incorporated into the forecasting process, we argue that the data aspect is relatively underappreciated. Existing transformer-based MTS approaches focus on the classical learning setup where a model is trained and tested on the same task. Although this often results in satisfactory performance for large datasets, it is likely to underperform in real-world applications where the training set is small or test set is out-of-distribution. This is especially so for modern sequence/transformer models that involve complex architectures and naturally require a substantial amount of data to operate at optimal performance. 

Our approach \name brings two key innovations: (1) Generating realistic and diverse large-scale multivariate time series data, where inter- and intra-channel dependencies are common, and (2) Developing an architecture capable of extracting time series features from this large-scale synthetic dataset. The architecture also allows for transfer learning to novel tasks with arbitrary number of channels. Overall, empowered by large amount of synthetic data (on the order of millions of samples), \name facilitates state-of-the-art zero-shot and few-shot accuracy on benchmark datasets. 

\begin{figure}[t!] 
   \centering    \includegraphics[width= 0.47\textwidth]{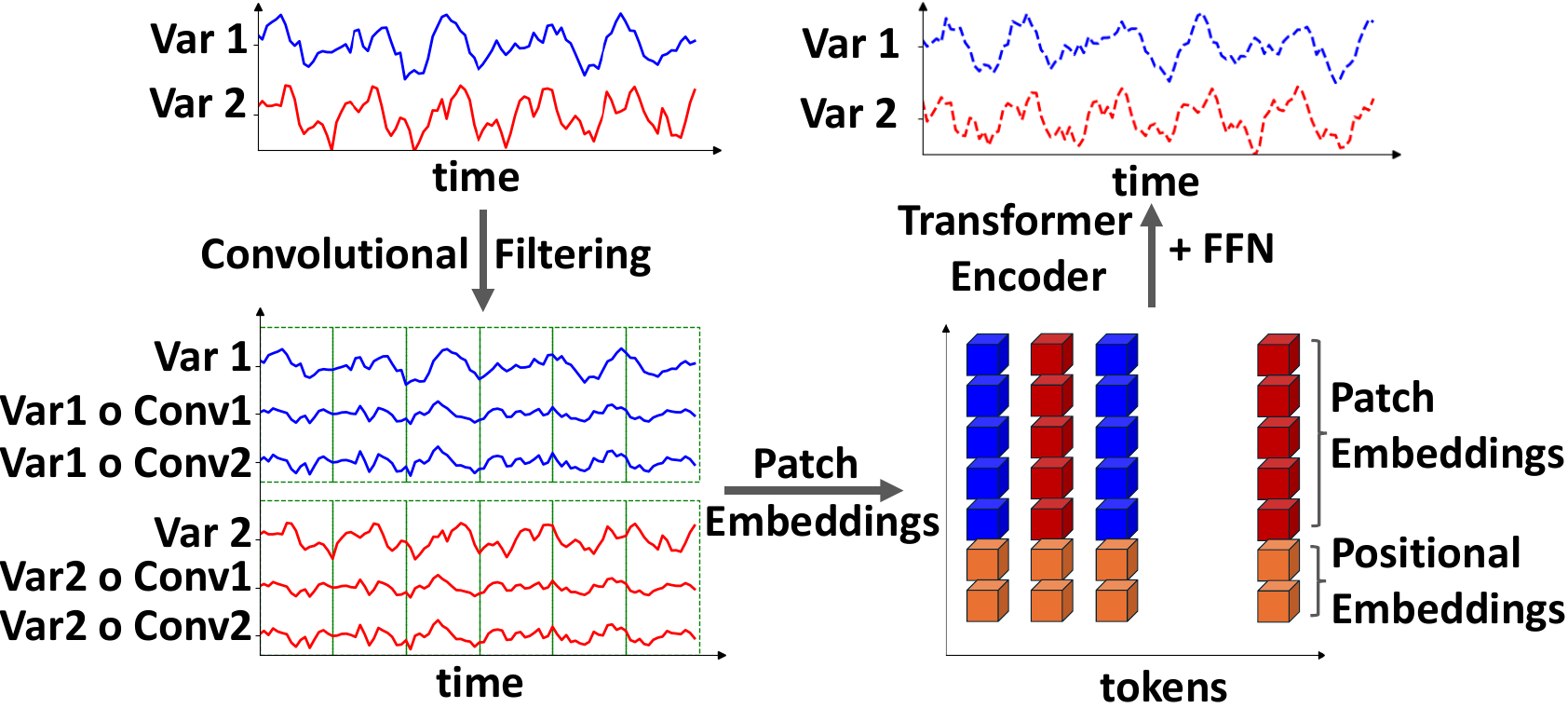}
   \caption{Illustration of the architecture of TimePFN. Variates are filtered with 1D convolutions, to be patched with overlapping strides, following \cite{Yuqietal-2023-PatchTST}. They are then fed into transformer encoder with channel mixing, with the final forecast coming from the final feedforward network.}
\end{figure}

The strong zero-shot performance of our model, along with its superior performance in few-shot settings, supports the importance of the data-centric perspective. Evaluations demonstrate that our model, when fine-tuned on as few as 50 to 500 samples, is competitive with the performance of alternative methods trained on the entire dataset. More specifically, we make the following contributions:
\begin{itemize}
    \item We present a new method to generate synthetic multivariate time series data using Gaussian processes with kernel compositions and a linear coregionalization model.
    \item We propose a variation of PatchTST \cite{Yuqietal-2023-PatchTST} for multivariate forecasting. Unlike PatchTST, our architecture incorporates channel mixing and employs a convolutional embedding module for patch embeddings. This allows it to effectively extract cross-channel relations and generate more representative embeddings, as demonstrated by experiments. 
    \item \name is the first multivariate time-series PFN. Notably, \name demonstrates strong zero-shot and few-shot performance and consistently outperforms comparable models/methods across various benchmarks.
    \item We find that \name also exhibits strong univariate forecasting performance, although it is explicitly trained with synthetic multivariate data. This attests to the flexibility and generalization capability of our approach. 
\end{itemize}

\begin{algorithm}[tb]
\caption{LMC-Synth}
\label{alg:algorithm}
\textbf{Input}: Number of variates $N$, time-series length $T$, Weibull shape parameter $\beta$, Weibull scale parameter $\lambda$, (min, max) value of dirichlet concentration parameter $(d_{min}, d_{max})$, minimum number of latent functions $m$, maximum number of kernel composition in KernelSynth $J$  \\
\textbf{Output}: Synthetic MTS $\vct{C}$ with $N$ variates and length $T$
\begin{algorithmic}[1] 
\STATE $L \sim max(min(Weibull(\beta, \alpha), N), m)$
\STATE $d \sim U(d_{min}, d_{max})$
\FOR{j $\in \{1 \dots L\} $}
\STATE $l_j(\vct{t}) \leftarrow KernelSynth(J,T)$ 
\ENDFOR
\FOR{i $\in \{1 \dots N\} $}
\STATE $[\alpha_{i,1} \dots \alpha_{i,L}] \sim Dir(d)$
\STATE  $C_i(\vct{t}) \leftarrow \sum_{j=1}^L \alpha_{i,j} l_j(\vct{t})$
\ENDFOR
\STATE \textbf{return} $\{C_i(\vct{t})\}_{i=1}^N$
\end{algorithmic}
\label{alg:SynthLMC}
\end{algorithm}
\section{Related Work}
\label{sec:related}

Transformers \cite{attention_is_all_you_need} have revolutionized NLP, significantly advancing zero-shot and few-shot capabilities in language and vision models. This has led researchers to explore the application of transformers to time-series forecasting, leading to a substantial body of work including but not limited to \cite{ haoyietal-informer-2021, wu2021autoformer, zhou2022fedformer,logtrans, Yuqietal-2023-PatchTST, liu2022pyraformer, zhang2023crossformer}.  Informer by \cite{liu2023itransformer} introduces the ProbSparse attention mechanism, which alleviates the quadratic complexity of the naive transformer to log-linear complexity, to mitigate the scalability issues in long sequence time-series forecasting (LSTF). \cite{zhou2022fedformer} uses the sparsity of the time-series in the fourier domain to enhance the performance in LSTF. PatchTST \cite{Yuqietal-2023-PatchTST} uses patching with overlapping strides as a tokenization mechanism to address issues associated with naive tokenization of time-series data. This approach yields patch-based tokens that are interpretable while maintaining channel independence, treating each channel as univariate but facilitating joint learning across all channels through the same set of shared weights. Our architecture deviates from PatchTST by incorporating convolutional layers before patching and using channel-mixing to capture interactions between tokens from different channels. The advantages of convolutions are highlighted in the speech literature by \cite{wav2vec,hsu_conv}. On the other hand, iTransformer \cite{liu2023itransformer} treats each variate as a single token, showing the potential benefits of utilizing inter-channel relationships.

In zero-shot forecasting, a line of work has emerged \cite{orozco, Oreshkin_Carpov_Chapados_Bengio_2021, domain-adapt, dooley2023forecastpfn, ansari2024chronos}. More recently, \cite{ansari2024chronos} has developed novel tokenization methods, employed quantization, and made time series data resemble language, enabling the training of LLM architectures for probabilistic univariate forecasting in a framework called Chronos. Chronos employs a data augmentation technique called KernelSynth, which generates synthetic time-series data using Gaussian processes to improve the generalization capability of the model. Meanwhile, another line of work, ForecastPFN \cite{dooley2023forecastpfn}, is trained entirely on a synthetic dataset following the framework of Prior-data Fitted Networks (PFNs). Initially proposed by \cite{muller2022transformers}, PFNs are designed to approximate Bayesian inference. Another study \cite{verdenius2024lat} integrates Joint-Embedding Predictive Architectures with PFNs for zero-shot forecasting. In addition to the mentioned models, Mamba4Cast \cite{bhethanabhotla2024mamba4castefficientzeroshottime} is also trained entirely on synthetic data using the Mamba architecture as its backbone \cite{gu2024mambalineartimesequencemodeling}. While the mentioned literature addresses univariate settings, our work introduces the first multivariate Prior-data Fitted Network, to the best of our knowledge, featuring an architecture that enables strong zero-shot and few-shot performances on MTS forecasting. 

\section{Proposed Method}
\label{sec:method}

This work relies on two key aspects: a multivariate synthetic time series data generation mechanism that encapsulates inter- and intra-channel dependencies common across real time series data, and an architecture capable of generalization to real datasets when trained on such a dataset. 

In the following section, we introduce the main concept behind our synthetic MTS data generation and training mechanism: Prior-data Fitted Networks (PFNs).

\subsection{Prior-data Fitted Networks for MTS Forecasting}


Let $\mathcal{D} := \{t, X_t\}_{t=1}^T$ represent an N-channel multivariate time series data spanning a time horizon \( T \), where \( X_t := [x_{t,1}, \dots, x_{t,N}] \). Each \( x_{t,i} \) is potentially causally dependent on previous time steps and on one another. Given the data $\{t, X_t\}_{t=1}^{\widetilde{T}}$ where \( \widetilde{T}<T \), the task is to forecast \( X_{\widetilde{T}+1}, \dots, X_T \). We tackle this problem using a Bayesian framework. Assuming a hypothesis space $\Omega$ with a prior distribution $p(\omega)$, each hypothesis $\omega \in \Omega$ models a multivariate time series (MTS) generating process, i.e., \( X_t = \omega(t) \). For example, $\Omega$ could represent the space of hypotheses for vector autoregression (VAR) models, where a particular instance $\omega \in \Omega$ corresponds to a specific VAR process, such as VAR(2), and data $\mathcal{D}$ can be generated via this process. Now, given a data $\mathcal{D} := \{t, X_t\}_{t=1}^{\widetilde{T}}$ where \( \widetilde{T}<T \), the posterior predictive distribution (PPD) of $\x \in \mathbb{R}^N $ at time \( T \) is \( p(\cdot \mid T, \mathcal{D}) \). By Bayes' theorem,

\begin{equation}
    p(\x \mid T, \mathcal{D})  \propto\int_\Omega p(\x \mid T,\omega)p(D\mid\omega)p(\omega)d\omega 
\end{equation}

As shown by \cite{muller2022transformers, hollmann2023tabpfn, dooley2023forecastpfn}, the posterior predictive distribution (PPD) is approximated using prior fitting networks (PFNs) as follows: We iteratively sample a hypothesis $\omega$ from the hypothesis space $\Omega$ according to the probability $p(\omega)$. Next, we generate a prior dataset $\mathcal{D}$ from this hypothesis, denoted as $\mathcal{D} \sim p(\mathcal{D} \mid \omega)$. We then optimize the parameters of the PFN on these generated datasets using standard methods. The time series dataset is divided into input and output parts, where $\mathcal{D}_{input}:= \{t, X_t\}_{t=1}^{\widetilde{T}}$ and $\mathcal{D}_{output}:= \{t, X_t\}_{t=\widetilde{T}+1}^{T}$. Subsequently, we train the PFN to forecast $\mathcal{D}_{output}$ from $\mathcal{D}_{input}$ using standard time-series transformer training techniques, aiming to minimize the mean-squared error loss as our optimization objective, following the setting of \cite{dooley2023forecastpfn}.

In our work, we define the hypothesis space $\Omega$ as consisting of single-input, multi-output Gaussian processes represented by the linear model of coregionalization (LMC) \cite{geostatistics_LMC}. Our choice is driven by the representational power of Gaussian processes and their ability to generate a diverse range of time series through the LMC framework.  


\subsection{Synthetic MTS Data Generation}
In synthetic MTS (multivariate time series) data generation, our goal is twofold. First, we strive to create variates that are realistic, exhibiting periodic patterns, trends, and other common features found in real-world data. Second, we aim for these variates to be correlated with one another, which better represents MTS data characteristics. Fortunately, our first goal is addressed by a method called KernelSynth. Chronos \cite{ansari2024chronos} uses KernelSynth to enrich its training corpus by randomly composing kernels using binary operators (such as addition and multiplication) to generate diverse, univariate synthetic time-series data. This method is essentially the inverse of the kernel composition approach described in  \cite{automatic_statistician}, where kernel compositions are used for structure discovery in nonparametric regression. In contrast, KernelSynth focuses on generating realizations from these kernels. For example, combining a linear kernel with a periodic kernel results in a pattern that exhibits both a linear trend and sinusoidal seasonality. Similarly, multiplying a squared-exponential kernel with a periodic kernel creates locally periodic patterns \cite{automatic_statistician}. Chronos aggregates kernels of various types—such as Linear, Periodic, Squared-Exponential, Rational, and Quadratic—and with different parameters (such as daily, weekly, and monthly periodic kernels) in a kernel bank, composing them as described above to define the Gaussian processes.

However, generating a MTS time-series data is yet to be addressed. To address the second goal, generating variates that are correlated in a realistic manner, we use a generative Gaussian modelling, called linear model of coregionalization (LMC), which is developed initially in the field of geostatistics \cite{geostatistics_LMC}. For ease of understanding, we adopt the time-series notation we used above. 
In LMC, the outputs are obtained as linear combinations of independent latent random functions. In other words, given $\vct{t} \in \mathbb{R}^T$, the outputs in each channel $\{C_i(\vct{t})\}_{i=1}^N$ is the linear combination of $L$ latent functions
\begin{equation}
    C_i(\vct{t}) = \sum_{j=1}^L \alpha_{i,j} l_j(\vct{t}) \label{eqq:lin-coreg}
\end{equation}
Observe that, since latent functions are independent with zero-mean the resulting output covariance is a well-defined PSD function with zero-mean \cite{kernels_for_vector_valued}. 
In our synthetic data generation algorithm, to avoid scaling issues, we restrict ourselves to convex combinations. Thus, for each $i$, we have $\alpha_{i,1}+\dots \alpha_{i,L} = 1 \text{ with } \alpha_{i,j} \geq 0$, meaning that the outputs lie in the convex hull of latent functions $l_j$'s. We generate the latent functions based on KernelSynth's algorithm due to it's extensive descriptive value. Note that the LMC formulation encapsulates the cases where the correlations between different variates are small or nonexistent. Specifically, the case where each variate is independent from the rest corresponds to $L=N$ with $C_i(\vct{t}) =  l_i(\vct{t})$. Such a modelling is important, as some MTS data have strong correlation between different variates, whereas others have small or non-existent correlation.  

In our algorithm, LMC-Synth, we sample the number of latent functions from a Weibull distribution and $[\alpha_{i,1} \dots \alpha_{i,L}]$ from a Dirichlet distribution. To avoid highly skewed cases, we impose upper and lower bounds on the possible number of latent functions. Since the uncorrelated setting of $L=N$ with $C_i(\vct{t}) =  l_i(\vct{t})$ is crucial for modeling MTS problems with low correlation among variates, we also generate data under this setting. Incorporating this extra setting is shown to yield the strongest performance in zero-shot settings.

\subsection{Architecture for TimePFN}
In designing the architecture, our main principle was to create a system capable of extracting time-series features useful for MTS forecasting. Through this, we aimed for the architecture to achieve better generalization when applied to real-world datasets. The primary advantage of the PFN framework in our case is that, since synthesizing large-scale synthetic MTS data is feasible with LMC-Synth, we are no longer constrained by data scarcity. Previous MTS models were compelled to balance model complexity with their datasets due to limited data, often restricting the use of certain components or their quantity (such as the number of transformer layers) to avoid overfitting. However, with access to large-scale MTS data, we can expand our architecture and freely incorporate additional components that we believe will improve forecasting performance on new datasets. In light of this, we proceed to explain our architecture and design choices. 

The \name model resembles PatchTST \cite{Yuqietal-2023-PatchTST} in several aspects when processing MTS data, but it differs significantly in two areas: our convolutional filtering of the variates prior to patching and channel-mixing. 

\textbf{Convolutional Filtering.} Before patching, consider an MTS dataset \( X = [x_1 \dots x_N] \) in \( \mathbb{R}^{L \times N} \), where \( L \) is the length and \( N \) is the number of variates. We apply learnable 1D convolution operations to each variate, with convolutional weights shared across all variates. After convolutions, we apply 1D magnitude max pooling to each newly generated variate, followed by a new set of 1D convolutions. In \name, each $x_i  \in \mathbb{R}^L$ is transformed into $\bar{x}_i \in \mathbb{R}^{(C+1) \times L}$, where $C$ rows come from 1D convolutional operations and magnitude max pooling, whereas one row is the original $x_i$. We keep the original $x_i$ to not loose any information, analogous to skip connections used in NLP \cite{he2015deepresiduallearningimage}. In practice, we used $C=9$. Filtering with convolutions is a valuable tool in time-series analysis. Many operations, such as differencing to de-trend data, can be effectively represented by convolutions. We utilized this approach to extract common time-series features across various datasets, thereby improving the generalization capability of our model.

\begin{table*}[t]
\centering
\small
\setlength{\tabcolsep}{0.5mm}
\begin{tabular}{ll|lr|rl|rl|rl|rl|rl|rl|ll|ll}
\hline
\multicolumn{2}{c|}{Dataset} & \multicolumn{2}{c|}{ECL} & \multicolumn{2}{c|}{Weather} & \multicolumn{2}{c|}{Traffic} & \multicolumn{2}{c|}{Solar} & \multicolumn{2}{c|}{Exchange} & \multicolumn{2}{c|}{ETTh1} & \multicolumn{2}{c|}{ETTh2} & \multicolumn{2}{c|}{ETTm1} & \multicolumn{2}{c}{ETTm2} \\ \cline{3-20} 
\multicolumn{2}{c|}{Models} & MSE & MAE & MSE & MAE & MSE & MAE & MSE & MAE & MSE & MAE & MSE & MAE & MSE & MAE & MSE & MAE & MSE & MAE \\ \hline
\parbox[t]{5mm}{\multirow{4}{*}{\rotatebox[origin=c]{90}{z.s. }}} & TimePFN & \textbf{0.315} & \textbf{0.383} & \textbf{0.209} & \textbf{0.255} & \textbf{1.108} & \textbf{0.613} & 0.941 & \textbf{0.730} & 0.105 & 0.229 & \textbf{0.453} & \textbf{0.439} & \textbf{0.328} & \textbf{0.362} & \textbf{0.637} & \textbf{0.512} & \textbf{0.212}  & \textbf{0.291} \\
 & Naive & 1.587 & 0.945 & 0.259 & 0.254 & 2.714 & 1.077 & 1.539 & 0.815 & \textbf{0.081} & \textbf{0.196} & 1.294 & 0.713 & 0.431 & 0.421  & 1.213 & 0.664 & 0.266 & 0.327 \\ 
  & SeasonalN. & 1.618 & 0.964 & 0.268 & 0.263 & 2.774 & 1.097 & 1.599 & 0.844 & 0.086 & 0.204 & 1.325 & 0.727 & 0.445 & 0.431  & 1.227 & 0.673 & 0.274 & 0.334 \\ 
  & Mean & 0.845 & 0.761 & 0.215 & 0.271 & 1.410 & 0.804 & \textbf{0.910} & 0.734 & 0.139 & 0.269 & 0.700 & 0.558 & 0.352 & 0.387 & 0.693 & 0.547 & 0.229 & 0.307 \\ \hline \hline
\parbox[t]{5mm}{\multirow{7}{*}{\rotatebox[origin=c]{90}{Budget =50 }}} & TimePFN & \textbf{0.235} & \textbf{0.322} & \textbf{0.190} & \textbf{0.235} & \textbf{0.746} & \textbf{0.468} & \textbf{0.429} & \textbf{0.450} & \textbf{0.096} & \textbf{0.218}  & \textbf{0.438} & \textbf{0.429} & \textbf{0.324} & \textbf{0.359} &  \textbf{0.419} & \textbf{0.418} & \textbf{0.195} & \textbf{0.276} \\
 & iTransformer & 0.278 & 0.360 & 0.237 & 0.278 & 0.801 & 0.499 & 0.513 & 0.479  & 0.145 & 0.275 & 0.838 & 0.617 & 0.410 & 0.422  & 0.884   & 0.608  & 0.268 & 0.337 \\
 & PatchTST & 0.667 & 0.646 & 0.221 & 0.269 & 1.295 & 0.746 & 0.810 & 0.669 & 0.127 & 0.255 & 0.778 & 0.587 & 0.372 & 0.401 & 0.656 & 0.528 & 0.231 & 0.310 \\
 & DLinear & 0.406 & 0.463 & 0.742 & 0.612 & 1.888 & 0.937 & 0.956 & 0.813 & 3.432 & 1.349 & 1.404 & 0.881 & 3.928 & 1.383 & 1.332 & 0.846 & 3.484 & 1.290 \\
 & FEDFormer & 0.908 & 0.758 & 0.306 & 0.381 & 1.587 & 0.874 & 0.972 & 0.757 & 0.165 & 0.300 & 0.676 & 0.570 & 0.424 & 0.468  & 0.745 & 0.589 & 0.291 & 0.387 \\ 
& Informer & 1.226 & 0.896 & 0.464 & 0.511 & 1.714 & 0.901 & 0.887 & 0.783 & 1.470 & 1.007 & 1.172 & 0.819 & 2.045 & 1.093  & 1.003 & 0.745 & 1.590 & 0.995 \\ 
  & Autoformer & 0.729 & 0.675 & 0.322 & 0.401 & 1.600 & 0.883 & 1.065 & 0.808 & 0.213 & 0.351 & 0.607 & 0.560 & 0.492 & 0.506 & 0.763 & 0.592 & 0.316 & 0.407 \\ \hline \hline

\parbox[t]{5mm}{\multirow{7}{*}{\rotatebox[origin=c]{90}{Budget =500 }}} & TimePFN & \textbf{0.190} & \textbf{0.283} & \textbf{0.178} & \textbf{0.222} & \textbf{0.487} & \textbf{0.335} & \textbf{0.269} & \textbf{0.305} & 0.083 & 0.203 & \textbf{0.401} & \textbf{0.412} & \textbf{0.311} & \textbf{0.352} & \textbf{0.360} & \textbf{0.386} & \textbf{0.185} & \textbf{0.268} \\
 & iTransformer & 0.200 & 0.284 & 0.211 & 0.248 & 0.514 & 0.354  & 0.307 & 0.334 & 0.113 & 0.239 & 0.489 & 0.470 & 0.361 & 0.394 & 0.569 & 0.494 & 0.231 & 0.310  \\
 & PatchTST & 0.236 & 0.320 & 0.210 & 0.246 & 0.740 & 0.455 & 0.321 & 0.353 & \textbf{0.081} & \textbf{0.198} & 0.596 & 0.515 & 0.358 & 0.392 & 0.369  & 0.386 & 0.190 & 0.275  \\
 & DLinear & 0.235 & 0.328 & 0.335 & 0.394 & 1.312 & 0.727 & 0.622 & 0.656 & 0.655 & 0.551 & 0.749 & 0.609 & 1.098 & 0.712 & 0.817 & 0.621 & 0.870 & 0.626 \\
 & FEDformer & 0.317 & 0.407 & 0.265 & 0.341 & 0.888 & 0.548 & 0.821 & 0.706 & 0.157 & 0.288 & 0.444 & 0.452 & 0.358 & 0.401  & 0.674 & 0.542 & 0.238 & 0.322 \\ 
   & Informer & 0.869 & 0.760 & 0.320 & 0.393 & 1.411 & 0.774 & 0.318 & 0.385 & 0.699 & 0.694 & 0.913 & 0.713 & 1.311 & 0.940  & 0.704 & 0.595 & 1.121 & 0.803 \\ 
  & Autoformer & 0.303 & 0.396 & 0.237 & 0.312 & 0.896 & 0.549 & 0.950 & 0.787 & 0.158 & 0.290 & 0.456 & 0.456 & 0.339 & 0.384 & 0.672 & 0.534 & 0.223 & 0.308 \\ \hline \hline

 \parbox[t]{5mm}{\multirow{6}{*}{\rotatebox[origin=c]{90}{Budget = All}}} & TimePFN & \textbf{0.138} & \textbf{0.137} & \textbf{0.166} & \textbf{0.208} & \textbf{0.392} & \textbf{0.260} & 0.203 & 0.219 & 0.100 & 0.223 & 0.402 & 0.417 & \textbf{0.293} & \textbf{0.343} & 0.392 & 0.402 & 0.180 & 0.262 \\
 & iTransformer & 0.147 & 0.239 & 0.175 & 0.215  & 0.393 & 0.268 & 0.201 & 0.233 & 0.086& 0.206  & \textbf{0.387} & 0.405 & 0.300 & 0.349 & 0.342 & 0.376 & 0.185 & 0.272 \\
 & PatchTST & 0.185 & 0.267 & 0.177 & 0.218 & 0.517 & 0.334 & 0.222 & 0.267 & \textbf{0.080} & \textbf{0.196} & 0.392 & \textbf{0.404} & \textbf{0.293} & \textbf{0.343} & \textbf{0.318} & \textbf{0.357} & \textbf{0.177} & \textbf{0.260} \\
 & DLinear & 0.195 & 0.278 & 0.341 & 0.412 & 0.690 & 0.432 & 0.286 & 0.375 & 0.101 & 0.237 & 0.400 & 0.412 & 0.357 & 0.406  & 0.344 & 0.371 & 0.195 & 0.293 \\
 & FEDformer & 0.196 & 0.310 & 0.227 & 0.313 & 0.573 & 0.357 & 0.242 & 0.342 & 0.148 & 0.280 & 0.380 & 0.417 & 0.340 & 0.386  & 0.363 & 0.408 & 0.191 & 0.286 \\ 
& Informer & 0.327 & 0.413 & 0.455 & 0.481 & 0.735 & 0.409 & \textbf{0.190} & \textbf{0.216} & 0.921 & 0.774 & 0.930 & 0.763 & 2.928 & 1.349  & 0.623 & 0.559 & 0.396 & 0.474 \\ 
  & Autoformer & 0.214 & 0.327 & 0.273 & 0.344 & 0.605 & 0.376 & 0.455 & 0.480 & 0.141 & 0.271 & 0.440 & 0.446 & 0.364 & 0.408 & 0.520 & 0.490 & 0.233 & 0.311 \\ \hline 
\multicolumn{2}{c|}{\# of Variates} & \multicolumn{2}{c|}{321} & \multicolumn{2}{c|}{21} & \multicolumn{2}{c|}{862} & \multicolumn{2}{c|}{137} & \multicolumn{2}{c|}{8} & \multicolumn{2}{c|}{7} & \multicolumn{2}{c|}{7} & \multicolumn{2}{c|}{7} & \multicolumn{2}{c}{7} \\ 

\end{tabular}
\caption{MTS forecasting results of TimePFN and comparable architectures with best results in bold. Input and forecast lengths are set to be 96. SeasonalN. stands for Seasonal Naive. \name demonstrates remarkable performance in budget-limited settings, as well as with the full dataset, particularly in scenarios involving a large number of variates.}
\label{tbl:mse_main}
\end{table*}

\textbf{Patch Embeddings.} Given $\bar{x}_i \in \mathbb{R}^{(C+1) \times L}$, we extract overlapping patches of size $P$ with a stride of $S$, following the settings described in \cite{Yuqietal-2023-PatchTST}. Each patch thus has dimensions $\mathbb{R}^{(C+1) \times P}$, and a total of $\left\lfloor \frac{L-P}{S} + 2 \right\rfloor$ patches are extracted from a single variate. In total, we get $N \times \left\lfloor \frac{L-P}{S} + 2 \right\rfloor$ patches. Each patch is then flattened and fed into a 2-layer feedforward neural network to be mapped to embedding dimension $D$.  We add  2D sinusoidal positional encodings \cite{wang2019translating} to the embeddings to correctly capture channel-wise and temporal information. In practice, we used $(P=16, S=8)$, similar to \cite{Yuqietal-2023-PatchTST}.

\textbf{Channel-mixing.} Unlike PatchTST \cite{Yuqietal-2023-PatchTST}, where the tokens from each channel are fed independently into a transformer encoder, we input all tokens into the transformer encoder after applying the positional encodings described above. Consequently, tokens from different variates can attend to each other.

\textbf{Transformer Encoder.} We employ a naive multihead transformer encoder, incorporating layer normalizations \cite{ba2016layernormalization} and skip connections \cite{he2015deepresiduallearningimage} to improve training stability. After feeding the tokens into the multilayer encoder, we rearrange them into their respective channels and apply a channel-wise flattening operation. This is followed by a two-layer feedforward network that processes the flattened variate representations using shared weights (single FFN is applied to all variates).

\textbf{Normalization.} We normalize each variate $x_i$ to have zero mean and unit standard deviation prior to any other process described above, as recommended by \cite{kim2021reversible}, to alleviate the impact of distribution shifts between our synthetic dataset and test examples \cite{liu2023itransformer, Yuqietal-2023-PatchTST}.  Before forecasting, we revert the time series to its original scale by de-normalizing. 

\textbf{Architectural Details.} Due to our architectural specifications, \name has fixed input sequence and forecasting lengths. However, it can accept an arbitrary number of variates. Thus, although we trained \name with a synthetic dataset of a fixed channel size ($C=160$), it can forecast with both fewer and more channels than those used in its training data. When forecasting with a number of channels $\bar{C} \leq C$, we directly input the data to \name. To mitigate the effects of distribution shifts when forecasting with more channels at test time, we process the data by splitting it into non-overlapping channels of size at most $C$. If the test data has $\bar{C}$ channels, we divide it into $\left\lfloor \frac{\bar{C}}{C} \right\rfloor + 1$ segments, input them separately, and then stack them afterwards. 

\begin{table*}[ht]
\small
\setlength{\tabcolsep}{0.9mm}
\begin{center}

\begin{tabular}{ll|lr|rl|rl|rl|rl|rl|rl|ll|}
\hline
\multicolumn{2}{c|}{Dataset} & \multicolumn{2}{c|}{TimePFN-36} & \multicolumn{2}{c|}{TimePFN-96} & \multicolumn{2}{c|}{ForecastPFN} & \multicolumn{2}{c|}{Chronos-s}  & \multicolumn{2}{c|}{SeasonalN.} & \multicolumn{2}{c|}{Naive} & \multicolumn{2}{c|}{Mean} \\ \cline{3-16} 
\multicolumn{2}{c|}{Models} & MSE & MAE & MSE & MAE & MSE & MAE & MSE & MAE & MSE & MAE & MSE & MAE & MSE & MAE  \\ \hline
\multicolumn{2}{l|}{ECL} & \textbf{0.752} & \textbf{0.703} & 0.509 & 0.549 & 1.416 & 0.958 & 1.152 & 0.792  & 1.559  & 0.995 & 1.211 & 0.829  &0.963  & 0.805   \\
\multicolumn{2}{l|}{Weather$\times 10^{2}$} & 0.042 &   1.381 & 0.046 & 1.477 & 0.084 & 1.999 & 0.036 & 1.136   &0.045 & 1.352  & \textbf{0.035}  & \textbf{1.123}  & 0.069  & 1.893  \\
\multicolumn{2}{l|}{Traffic} & \textbf{1.503} & \textbf{1.032} & 0.414 & 0.503 &4.521 & 1.742 & 3.103 & 1.364   &4.301  & 1.771 & 3.330  &1.463  & 2.125 & 1.256   \\ 
\multicolumn{2}{l|}{Exchange} & 0.027 & 0.125 & 0.034 & 0.139 & 0.057 & 0.180 & 0.049 & 0.113 & 0.028 & 0.128 & \textbf{0.022} & \textbf{0.107} & 0.040 & 0.156   \\ 
\multicolumn{2}{l|}{ETTh1} & \textbf{0.029} & 0.130 & 0.030 & 0.133 & 0.102 & 0.237 & 0.061 & 0.155  &  0.039& 0.151 & 0.031 & \textbf{0.128} & 0.040 & 0.154  \\ 
\multicolumn{2}{l|}{ETTh2} & \textbf{0.126} & \textbf{0.273} & 0.086 & 0.224 & 0.434 & 0.517  & 0.207 & 0.321  &0.279& 0.408 &0.215  & 0.336 &0.168  & 0.321  \\ \hline 

\end{tabular}

\end{center}
\caption{Zero-shot results of TimePFN on univariate time-series forecasting with input length = 36. TimePFN-96 has input length of 96. The errors are averaged over forecasting lengths of $\{6,8,14,18,24,36,48\}$. Chronos-s stands for Chronos-small. The best results are in bold (excluding TimePFN-96). } 
\label{tbl:univariate_res}
\end{table*}

\section{Experiments}
\label{sec:exp}
In all MTS evaluations, our primary objective is to forecast a horizon of 96 time steps using an MTS input of 96 time steps. We trained a single \name model on a large-scale, multivariate synthetic dataset generated by LMC-Synth and conducted all experiments using this model. We generated 15,000 synthetic datasets with a length of 1024 and a channel size of 160 from LMC-Synth, further augmenting with datasets having independent variates as in the case $C_i(\vct{t}) = l_i(\vct{t})$. The independent data comprises approximately 25\% of the purely correlated data. During training, we extracted time-series input and output pairs using a sliding window of size 192 (96 for input, 96 for output), resulting in approximately 1.5 million synthetic data points. We trained the model to forecast the MTS output based on the given input using MSE loss with our 160 channel synthetic dataset. Training a single \name of 8 transformer layers takes around $~10$ hours on L40S GPU.

In the few-shot evaluations, we fine-tuned \name using the specified data budget. We did not perform any hyperparameter tuning on \name, and the same set of hyperparameters was used in all few-shot settings. Details about the model hyperparameters are provided in the appendix.

\textbf{Benchmark Datasets.} We evaluated \name on nine widely-used, real-world benchmark datasets for MTS forecasting. These datasets include ETTh1, ETTh2, ETTm1, ETTm2 (collectively referred to as ETT, representing Electricity Transformer Temperature), Weather, Solar Energy, ECL (Electricity Consuming Load), Exchange, and Traffic. The Solar Energy dataset was introduced by \cite{LSTNet}, while the others were introduced by \cite{wu2021autoformer}. We provide the specifications of these datasets in the appendix section \textit{Datasets}. 

\textbf{Baselines.} Since no MTS PFN is available, we compared \name with state-of-the-art transformer-based MTS forecasting models, including FEDformer \cite{zhou2022fedformer}, Autoformer \cite{wu2021autoformer}, Informer \cite{haoyietal-informer-2021}, PatchTST \cite{Yuqietal-2023-PatchTST}, and iTransformer \cite{liu2023itransformer}. We evaluated these models across the entire dataset and at various data budgets, including 50, 100, 500, and 1000 data points. For instance, at a data budget of 500, the model is trained using 500 MTS input and output pairs. Additionally, we included DLinear \cite{Zeng2022AreTE}, a linear model, as part of our baseline. Given its lower complexity, we consider it a strong baseline for smaller data budgets. 

For smaller data budgets and our zero-shot evaluations, we incorporated three algorithmic baselines as suggested by \cite{dooley2023forecastpfn} and \cite{ansari2024chronos}: Mean, Naive, and Seasonal Naive. These baselines are applied independently to each variate. The Mean baseline forecasts by repeating the mean value of the input variate. The Naive approach forecasts by repeating the last value of the input variate. In the Seasonal Naive method, we assume a periodicity of seven.

Although \name is specifically trained for multivariate time-series forecasting, we also evaluated its performance on univariate forecasting (C=1) to demonstrate its robust generalization capabilities. In this context, we compared it with ForecastPFN \cite{dooley2023forecastpfn} and Chronos \cite{ansari2024chronos}, two state-of-the-art univariate zero-shot forecasters. ForecastPFN utilizes an input sequence length of 36, while \name operates with a sequence length of 96. To accommodate this discrepancy, we padded the additional 60 time steps with the mean value of the input sequence when running \name. These models are evaluated over forecast lengths of 6, 8, 14, 18, 24, 36, and 48. We used the smaller version of Chronos. The full results are detailed in the appendix, while in the main text, we report the averaged MSE and MAE values across these forecast lengths. Furthermore, to showcase the complete forecasting performance of \name, we also conducted runs with a non-padded sequence length of 96. We evaluated all results ourselves. 

\textbf{Experimental Setting.} When comparing \name with the aforementioned baselines, we use the hyperparameters reported in their official codebases. We re-run the experiments with limited budgets and by utilizing the entire training dataset. \cite{liu2023itransformer} presents the forecasting results for the mentioned transformer-based MTS architectures using the full training dataset. We re-run all the experiments and selected the best results from both our run and their report to ensure that the performance of other architectures is not underreported when we use the entire training set. Our unaltered results are included in the appendix. 

In \name, we use a single model with fixed hyperparameters that is trained only once on our large-scale multivariate synthetic dataset. In few-shot evaluations, we fine-tune \name with a given data budget, maintaining the same hyperparameters across different datasets. In all evaluations except for univariate cases, we report the forecasting errors for the next 96 time steps, given a multivariate time series (MTS) of sequence length 96. Our implementation details and further experimental settings such as hyperparemeters are reported in the Appendix: \textit{Implementation Details}.   

\subsection{Main Results}
In MTS forecasting, we compared \name with various baselines in zero-shot settings, as well as with different data budgets, and by utilizing the entire dataset. Table 1 presents our results for zero-shot settings, data budgets of 50 and 500, and scenarios using the entire dataset. Our comprehensive results, which also include data budgets of 100 and 1000, can be found in the Appendix under the section \textit{Extended Results}. With a data budget of 50, \name outperforms all transformer-based architectures and DLinear. However, with a data budget of 500, it surpasses all baselines except for PatchTST \cite{Yuqietal-2023-PatchTST} in the exchange dataset, closely competing with it. When utilizing the entire dataset, \name achieves the best results in four datasets, equaling the performance of PatchTST. Given that we fine-tuned \name with fixed hyperparameters across all datasets, and selected the best results from the baselines and the findings reported in \cite{liu2023itransformer}, the performance of \name is noteworthy. We observe that \name excels in datasets with a greater number of variates and a more multivariate nature, while PatchTST primarily excels in ETT datasets and Exchange. This outcome is anticipated, as \name is designed to incorporate channel mixing, whereas PatchTST is designed with channel independence. Indeed, the lower forecasting performance of PatchTST on Traffic dataset supports this hypothesis.

\begin{table*}[t]
\centering
\small
\setlength{\tabcolsep}{0.4mm}
\begin{tabular}{ll|lr|rl|rl|rl|rl|rl|rl|ll|ll}
\hline
\multicolumn{2}{c|}{Dataset} & \multicolumn{2}{c|}{ECL} & \multicolumn{2}{c|}{Weather} & \multicolumn{2}{c|}{Traffic} & \multicolumn{2}{c|}{Solar} & \multicolumn{2}{c|}{Exchange} & \multicolumn{2}{c|}{ETTh1} & \multicolumn{2}{c|}{ETTh2} & \multicolumn{2}{c|}{ETTm1} & \multicolumn{2}{c}{ETTm2} \\ \cline{3-20} 
\multicolumn{2}{c|}{Models} & MSE & MAE & MSE & MAE & MSE & MAE & MSE & MAE & MSE & MAE & MSE & MAE & MSE & MAE & MSE & MAE & MSE & MAE \\ \hline
\parbox[t]{2mm}{\multirow{3}{*}{\rotatebox[origin=c]{90}{z.s. }}} & TimePFN & \textbf{0.315} & \textbf{0.383} & \textbf{0.209} & \textbf{0.255} & \textbf{1.108} & \textbf{0.613} & \textbf{0.941} & \textbf{0.730} & \textbf{0.105} & \textbf{0.229} & \textbf{0.453} & \textbf{0.439} & 0.328 & \textbf{0.362} & \textbf{0.637} & \textbf{0.512} & \textbf{0.212}  & \textbf{0.291} \\
 &TimePFN-w/o Conv & 0.653 & 0.637 & 0.221 & 0.271 & 1.287 & 0.757 & 1.197 & 0.829 & 0.111 & 0.237 & 0.608 & 0.517 & 0.338 & 0.374  & 0.771 & 0.565 & 0.224 & 0.307 \\ 
  & PatchTST-PFN & 0.470 & 0.522 & 0.212 & 0.262 & 1.172 & 0.702 & 1.014 & 0.787 & 0.108 & 0.231 & 0.554 & 0.501 & \textbf{0.322} & 0.366  & 0.746 & 0.560 & 0.215 & 0.301 \\  \hline

\end{tabular}
\caption{In TimePFN-w/o-Convolution, we eliminate the convolutional operator that is normally applied to the initial input variates. In PatchTST-PFN, we train a PatchTST model to evaluate the significance of channel-mixing and the appropriateness of our architecture for PFNs. Both the sequence length and the forecasting length are set to 96. }
\label{tbl:arch-ablation}
\end{table*}

\begin{table*}[t]
\centering
\small
\setlength{\tabcolsep}{0.6mm}
\begin{tabular}{ll|lr|rl|rl|rl|rl|rl|rl|ll|ll}
\hline
\multicolumn{2}{c|}{Dataset} & \multicolumn{2}{c|}{ECL} & \multicolumn{2}{c|}{Weather} & \multicolumn{2}{c|}{Traffic} & \multicolumn{2}{c|}{Solar} & \multicolumn{2}{c|}{Exchange} & \multicolumn{2}{c|}{ETTh1} & \multicolumn{2}{c|}{ETTh2} & \multicolumn{2}{c|}{ETTm1} & \multicolumn{2}{c}{ETTm2} \\ \cline{3-20} 
\multicolumn{2}{c|}{Models} & MSE & MAE & MSE & MAE & MSE & MAE & MSE & MAE & MSE & MAE & MSE & MAE & MSE & MAE & MSE & MAE & MSE & MAE \\ \hline
\parbox[t]{2mm}{\multirow{2}{*}{\rotatebox[origin=c]{90}{z.s. }}} & TimePFN & \textbf{0.315} & \textbf{0.383} & \textbf{0.209} & \textbf{0.255} & \textbf{1.108} & \textbf{0.613} & \textbf{0.941} & \textbf{0.730} & \textbf{0.105} & \textbf{0.229} & \textbf{0.453} & \textbf{0.439} & 0.328 & \textbf{0.362} & \textbf{0.637} & \textbf{0.512} & \textbf{0.212}  & \textbf{0.291} \\
 &TimePFN-Ind & 0.350 & 0.416 & 0.214 & 0.260 & 1.180 & 0.651 & 1.197 & 0.829 & 0.113 & 0.238 & 0.468 & 0.447 & \textbf{0.326} & 0.363  & 0.761 & 0.542 & 0.215 & 0.295 \\  \hline

\end{tabular}
\caption{TimePFN-ind is the model trained using only independent variates, while the other model is our standard one, which we used throughout the experiments. Both models have a sequence and forecasting length of 96. }
\label{tbl:data-ablation}
\end{table*}

In zero-shot settings, \name outperforms all zero-shot baselines except on the Solar-Energy and Exchange datasets, with Solar-Energy being in close proximity. In fact, the Exchange dataset is highly non-trivial, as simply using the last value as the forecast outperforms all baselines, except for PatchTST in the entire budget case. We observed that the Solar-Energy data exhibits sudden spikes e.g.~as a function of sun rising or going down. Our model, based on its training data from the LMC-Synth prior, fails to anticipate such sudden spikes. However, these spiky behavior is well within the capabilities of changepoint kernels in Gaussian processes, suggesting a clear path for future improvements.

\subsection{Univariate Time-Series Forecasting}
Although \name was specifically trained for MTS forecasting using a synthetic dataset with a channel size of 160, we also tested it in a zero-shot scenario for univariate time-series forecasting where \(C=1\). Moreover, we used the sequence length of 36 that ForecastPFN \cite{dooley2023forecastpfn} was specifically trained on. To accommodate this sequence length, we padded the remaining \(96-36 = 60\) sequence lengths with the mean value of the input time-series to mitigate any scaling issues, and named this model configuration \name-36. To demonstrate the full performance of our model, we included results for \name without padding using a sequence length of 96, referred to as \name-96 in Table 2. All other results were reported with a sequence length of 36. 

As demonstrated in Table 2, \name outperforms models that were specifically trained for univariate time series forecasting, which attests to its robust generalization and zero-shot performance. Our extensive evaluations, which detail the errors for different sequence lengths, can be found in the appendix under the section \textit{Extended Results}.

\subsection{Ablation Study}

Training a single \name model requires approximately 10 hours on a single L40S GPU, which limited our capacity for ablation studies. Nevertheless, we conducted two types of key ablations: the first type focused on the architecture, while the second type focused on the synthetic data generation.

\textbf{Architectural Ablation.} In the first part, we first aim to understand the impact of our 1D convolutional operation applied to time-series variates before any patching. To do this, we remove the operation and train a \name-convolutionless model, then report the zero-shot results in Table 3. We observe that without the convolutional operation, the zero-shot performance significantly decreases. Additionally, since our architecture differs from that of \cite{Yuqietal-2023-PatchTST} particularly in terms of channel mixing, we trained the PatchTST architecture to assess the impact of channel mixing on zero-shot forecasting performance. As seen in Table 3, both of our ablation experiments supports our model design principles and underscores the usefulness of \name's architecture for synthetic time series learning. 

\textbf{Synthetic Dataset Ablation.} To understand whether the synthetic data generation algorithm LMC-Synth gives any benefits over just using the variates generated by KernelSynth \cite{ansari2024chronos} independently in each channel, we trained \name with using data where each channel is generated independently. This case, as we described previously, corresponds to the case where $C_i(\vct{t}) =  l_i(\vct{t})$ with number of channels equaling to number of latent functions.  We see in Table 4 that using generative coregionalization provides clear benefits.


\section{Conclusion}
\label{sec:conclusion}
In this work, we demonstrate that with large-scale synthetic training and a suitable architecture for extracting useful time series features, fine-tuning with as few as 50 to 500 examples are sufficient to achieve competitive performance in multivariate time series forecasting. To this end, we present a novel method for generating large-scale synthetic MTS data with realistic intra- and inter-channel dependencies, called LMC-Synth, utilizing Gaussian processes and linear coregionalization model. Simultaneously, we developed an architecture capable of transfer learning, utilizing 1D convolutions applied to time series variates and channel-mixed patching. \name exhibits strong zero-shot performance, and although it is explicitly trained for MTS forecasting, it also excels in zero-shot univariate forecasting, demonstrating the flexibility and generality of our framework. To the best of our knowledge, \name is the first multivariate time-series PFN. For future work, we aim to improve our synthetic data-generation mechanism to better model sudden changes and multi-scale challenges that are prevalent in many time-series tasks. Additionally, integrating time series PFNs with tabular models presents an intriguing avenue for research. Moreover, we plan to extend our efforts into developing foundation models for multivariate time series.

\section*{Acknowledgements}
This work was supported in part by the NSF grants CCF-2046816, CCF-2403075, the Office of Naval Research grant N000142412289, and gifts by Open Philanthropy and Google Research.


\bibliography{aaai25}
\clearpage

\appendix

\newpage
\section{Extended Results}

\begin{table*}[t]
\centering
\small
\setlength{\tabcolsep}{0.6mm}
\begin{tabular}{ll|lr|rl|rl|rl|rl|rl|rl|ll|ll}
\hline
\multicolumn{2}{c|}{Dataset} & \multicolumn{2}{c|}{ECL} & \multicolumn{2}{c|}{Weather} & \multicolumn{2}{c|}{Traffic} & \multicolumn{2}{c|}{Solar} & \multicolumn{2}{c|}{Exchange} & \multicolumn{2}{c|}{ETTh1} & \multicolumn{2}{c|}{ETTh2} & \multicolumn{2}{c|}{ETTm1} & \multicolumn{2}{c}{ETTm2} \\ \cline{3-20} 
\multicolumn{2}{c|}{Models} & MSE & MAE & MSE & MAE & MSE & MAE & MSE & MAE & MSE & MAE & MSE & MAE & MSE & MAE & MSE & MAE & MSE & MAE \\ \hline
\parbox[t]{2mm}{\multirow{4}{*}{\rotatebox[origin=c]{90}{z.s. }}} & TimePFN & \textbf{0.315} & \textbf{0.383} & \textbf{0.209} & \textbf{0.255} & \textbf{1.108} & \textbf{0.613} & 0.941 & \textbf{0.730} & 0.105 & 0.229 & \textbf{0.453} & \textbf{0.439} & \textbf{0.328} & \textbf{0.362} & \textbf{0.637} & \textbf{0.512} & \textbf{0.212}  & \textbf{0.291} \\
 & Naive & 1.587 & 0.945 & 0.259 & 0.254 & 2.714 & 1.077 & 1.539 & 0.815 & \textbf{0.081} & \textbf{0.196} & 1.294 & 0.713 & 0.431 & 0.421  & 1.213 & 0.664 & 0.266 & 0.327 \\ 
  & SeasonalNaive & 1.618 & 0.964 & 0.268 & 0.263 & 2.774 & 1.097 & 1.599 & 0.844 & 0.086 & 0.204 & 1.325 & 0.727 & 0.445 & 0.431  & 1.227 & 0.673 & 0.274 & 0.334 \\ 
  & Mean & 0.845 & 0.761 & 0.215 & 0.271 & 1.410 & 0.804 & \textbf{0.910} & 0.734 & 0.139 & 0.269 & 0.700 & 0.558 & 0.352 & 0.387 & 0.693 & 0.547 & 0.229 & 0.307 \\ \hline \hline
\parbox[t]{2mm}{\multirow{7}{*}{\rotatebox[origin=c]{90}{Budget =50 }}} & TimePFN & \textbf{0.235} & \textbf{0.322} & \textbf{0.190} & \textbf{0.235} & \textbf{0.746} & \textbf{0.468} & \textbf{0.429} & \textbf{0.450} & \textbf{0.096} & \textbf{0.218}  & \textbf{0.438} & \textbf{0.429} & \textbf{0.324} & \textbf{0.359} &  \textbf{0.419} & \textbf{0.418} & \textbf{0.195} & \textbf{0.276} \\
 & iTransformer & 0.278 & 0.360 & 0.237 & 0.278 & 0.801 & 0.499 & 0.513 & 0.479  & 0.145 & 0.275 & 0.838 & 0.617 & 0.410 & 0.422  & 0.884   & 0.608  & 0.268 & 0.337 \\
 & PatchTST & 0.667 & 0.646 & 0.221 & 0.269 & 1.295 & 0.746 & 0.810 & 0.669 & 0.127 & 0.255 & 0.778 & 0.587 & 0.372 & 0.401 & 0.656 & 0.528 & 0.231 & 0.310 \\
 & DLinear & 0.406 & 0.463 & 0.742 & 0.612 & 1.888 & 0.937 & 0.956 & 0.813 & 3.432 & 1.349 & 1.404 & 0.881 & 3.928 & 1.383 & 1.332 & 0.846 & 3.484 & 1.290 \\
 & FEDFormer & 0.908 & 0.758 & 0.306 & 0.381 & 1.587 & 0.874 & 0.972 & 0.757 & 0.165 & 0.300 & 0.676 & 0.570 & 0.424 & 0.468  & 0.745 & 0.589 & 0.291 & 0.387 \\ 
& Informer & 1.226 & 0.896 & 0.464 & 0.511 & 1.714 & 0.901 & 0.887 & 0.783 & 1.470 & 1.007 & 1.172 & 0.819 & 2.045 & 1.093  & 1.003 & 0.745 & 1.590 & 0.995 \\ 
  & Autoformer & 0.729 & 0.675 & 0.322 & 0.401 & 1.600 & 0.883 & 1.065 & 0.808 & 0.213 & 0.351 & 0.607 & 0.560 & 0.492 & 0.506 & 0.763 & 0.592 & 0.316 & 0.407 \\ \hline \hline

\parbox[t]{2mm}{\multirow{7}{*}{\rotatebox[origin=c]{90}{Budget =100 }}} & TimePFN & \textbf{0.221} & \textbf{0.309} & \textbf{0.187} & \textbf{0.232} & \textbf{0.644} & \textbf{0.424} & \textbf{0.351} & \textbf{0.383} & \textbf{0.083} & \textbf{0.205}  & \textbf{0.441} & \textbf{0.429} & \textbf{0.322} & \textbf{0.356} &  \textbf{0.412} & \textbf{0.411} & \textbf{0.196} & \textbf{0.273} \\
 & iTransformer & 0.253 & 0.337 & 0.220 & 0.263 & 0.740 & 0.468 & 0.369 & 0.387  & 0.138 & 0.268 & 0.728 & 0.574 & 0.401 & 0.418  & 0.816   & 0.586  & 0.260 & 0.331 \\
 & PatchTST & 0.361 & 0.432 & 0.216 & 0.256 & 0.982 & 0.592 & 0.575 & 0.524 & 0.102 & 0.227 & 0.757 & 0.579 & 0.371 & 0.400 & 0.502 & 0.461 & 0.215 & 0.298 \\
 & DLinear & 0.332 & 0.409 & 0.636 & 0.562 & 1.770 & 0.897 & 0.887 & 0.784 & 2.712 & 1.172 & 1.256 & 0.826 & 3.237 & 1.246 & 1.214 & 0.799 & 2.810 & 1.140 \\
 & FEDformer & 0.597 & 0.598 & 0.264 & 0.344 & 1.350 & 0.775 & 0.951 & 0.752 & 0.158 & 0.291 & 0.562 & 0.513 & 0.362 & 0.407  & 0.724 & 0.572 & 0.290 & 0.387 \\ 
  & Informer & 1.056 & 0.837 & 0.369 & 0.432 & 1.609 & 0.860 & 0.731 & 0.702 & 0.883 & 0.774 & 1.038 & 0.757 & 1.279 & 0.916  & 0.883 & 0.664 & 1.040 & 0.802 \\ 
  & Autoformer & 0.468 & 0.519 & 0.251 & 0.330 & 1.344 & 0.773 & 0.960 & 0.762 & 0.187 & 0.321 & 0.550 & 0.524 & 0.379 & 0.422 & 0.704 & 0.554 & 0.258 & 0.351 \\ \hline \hline

\parbox[t]{2mm}{\multirow{7}{*}{\rotatebox[origin=c]{90}{Budget =500 }}} & TimePFN & \textbf{0.190} & \textbf{0.283} & \textbf{0.178} & \textbf{0.222} & \textbf{0.487} & \textbf{0.335} & \textbf{0.269} & \textbf{0.305} & 0.083 & 0.203 & \textbf{0.401} & \textbf{0.412} & \textbf{0.311} & \textbf{0.352} & \textbf{0.360} & \textbf{0.386} & \textbf{0.185} & \textbf{0.268} \\
 & iTransformer & 0.200 & 0.284 & 0.211 & 0.248 & 0.514 & 0.354  & 0.307 & 0.334 & 0.113 & 0.239 & 0.489 & 0.470 & 0.361 & 0.394 & 0.569 & 0.494 & 0.231 & 0.310  \\
 & PatchTST & 0.236 & 0.320 & 0.210 & 0.246 & 0.740 & 0.455 & 0.321 & 0.353 & \textbf{0.081} & \textbf{0.198} & 0.596 & 0.515 & 0.358 & 0.392 & 0.369  & 0.386 & 0.190 & 0.275  \\
 & DLinear & 0.235 & 0.328 & 0.335 & 0.394 & 1.312 & 0.727 & 0.622 & 0.656 & 0.655 & 0.551 & 0.749 & 0.609 & 1.098 & 0.712 & 0.817 & 0.621 & 0.870 & 0.626 \\
 & FEDformer & 0.317 & 0.407 & 0.265 & 0.341 & 0.888 & 0.548 & 0.821 & 0.706 & 0.157 & 0.288 & 0.444 & 0.452 & 0.358 & 0.401  & 0.674 & 0.542 & 0.238 & 0.322 \\ 
   & Informer & 0.869 & 0.760 & 0.320 & 0.393 & 1.411 & 0.774 & 0.318 & 0.385 & 0.699 & 0.694 & 0.913 & 0.713 & 1.311 & 0.940  & 0.704 & 0.595 & 1.121 & 0.803 \\ 
  & Autoformer & 0.303 & 0.396 & 0.237 & 0.312 & 0.896 & 0.549 & 0.950 & 0.787 & 0.158 & 0.290 & 0.456 & 0.456 & 0.339 & 0.384 & 0.672 & 0.534 & 0.223 & 0.308 \\ \hline \hline

\parbox[t]{2mm}{\multirow{7}{*}{\rotatebox[origin=c]{90}{Budget = 1000 }}} & TimePFN & \textbf{0.173} & \textbf{0.268} & \textbf{0.175} & \textbf{0.219} & \textbf{0.452} & \textbf{0.310} & 0.243 & 0.288 & 0.084 & 0.204 & \textbf{0.405} & \textbf{0.415} & \textbf{0.304} & \textbf{0.351} & \textbf{0.344} & \textbf{0.378} & \textbf{0.180} & \textbf{0.262} \\
 & iTransformer & 0.184 & 0.271 & 0.206 & 0.242 & 0.469 & 0.324  & 0.276 & 0.309 & 0.100 & 0.223 & 0.433 & 0.436 & 0.336 & 0.379 & 0.464 & 0.444 & 0.211 & 0.294  \\
 & PatchTST & 0.219 & 0.304 & 0.198 & 0.237 & 0.683 & 0.420 & 0.280 & 0.324 & \textbf{0.082} & \textbf{0.200} & 0.490 & 0.467 & 0.337 & 0.378 & 0.353  & 0.375 & 0.187 & 0.272  \\
 & DLinear & 0.218 & 0.310 & 0.254 & 0.331 & 1.076 & 0.627 & 0.488 & 0.569 & 0.193 & 0.330 & 0.562 & 0.513 & 0.528 & 0.507 & 0.629 & 0.528 & 0.380 & 0.437 \\
 & FEDformer & 0.284 & 0.379 & 0.269 & 0.341 & 0.806 & 0.486 & 0.545 & 0.546 & 0.157 & 0.287 & 0.402 & 0.435 & 0.341 & 0.383  & 0.436 & 0.456 & 0.228 & 0.312 \\ 
   & Informer & 0.693 & 0.647 & 0.341 & 0.413 & 1.231 & 0.678 & \textbf{0.229} & \textbf{0.294} & 0.689 & 0.666 & 0.887 & 0.710 & 1.357 & 0.939  & 0.682 & 0.596 & 0.615 & 0.596 \\ 
  & Autoformer & 0.270 & 0.367 & 0.239 & 0.314 & 0.787 & 0.492 & 0.926 & 0.742 & 0.156 & 0.285 & 0.427 & 0.442 & 0.341 & 0.383 & 0.617 & 0.521 & 0.218 & 0.301 \\ \hline \hline

 \parbox[t]{2mm}{\multirow{7}{*}{\rotatebox[origin=c]{90}{Budget = All }}} & TimePFN & \textbf{0.138} & \textbf{0.137} & \textbf{0.166} & \textbf{0.208} & \textbf{0.392} & \textbf{0.260} & 0.203 & 0.219 & 0.100 & 0.223 & 0.402 & 0.417 & \textbf{0.293} & \textbf{0.343} & 0.392 & 0.402 & 0.180 & 0.262 \\
 & iTransformer & 0.147 & 0.239 & 0.175 & 0.215  & 0.393 & 0.268 & 0.201 & 0.233 & 0.086& 0.206  & \textbf{0.387} & 0.405 & 0.300 & 0.349 & 0.342 & 0.376 & 0.185 & 0.272 \\
 & PatchTST & 0.185 & 0.267 & 0.177 & 0.218 & 0.517 & 0.334 & 0.222 & 0.267 & \textbf{0.080} & \textbf{0.196} & 0.392 & \textbf{0.404} & \textbf{0.293} & \textbf{0.343} & \textbf{0.318} & \textbf{0.357} & \textbf{0.177} & \textbf{0.260} \\
 & DLinear & 0.195 & 0.278 & 0.341 & 0.412 & 0.690 & 0.432 & 0.286 & 0.375 & 0.101 & 0.237 & 0.400 & 0.412 & 0.357 & 0.406  & 0.344 & 0.371 & 0.195 & 0.293 \\
 & FEDformer & 0.196 & 0.310 & 0.227 & 0.313 & 0.573 & 0.357 & 0.242 & 0.342 & 0.148 & 0.280 & 0.380 & 0.417 & 0.340 & 0.386  & 0.363 & 0.408 & 0.191 & 0.286 \\ 
& Informer & 0.327 & 0.413 & 0.455 & 0.481 & 0.735 & 0.409 & \textbf{0.190} & \textbf{0.216} & 0.921 & 0.774 & 0.930 & 0.763 & 2.928 & 1.349  & 0.623 & 0.559 & 0.396 & 0.474 \\ 
  & Autoformer & 0.214 & 0.327 & 0.273 & 0.344 & 0.605 & 0.376 & 0.455 & 0.480 & 0.141 & 0.271 & 0.440 & 0.446 & 0.364 & 0.408 & 0.520 & 0.490 & 0.233 & 0.311 \\ \hline \hline

 \parbox[t]{2mm}{\multirow{7}{*}{\rotatebox[origin=c]{90}{Avg Acc. Budgets }}} & TimePFN & \textbf{0.191}	& \textbf{0.264}	&\textbf{0.179}	&\textbf{0.223}&	\textbf{0.544}	&\textbf{0.359}	&\textbf{0.299}	&\textbf{0.329}&	\textbf{0.089}&	\textbf{0.211}&	\textbf{0.417}&	\textbf{0.420}&	\textbf{0.311}	&\textbf{0.352}&	\textbf{0.385}	&\textbf{0.399}&	\textbf{0.187}&	\textbf{0.268 } \\
&iTransformer &0.212	&0.298&	0.210&	0.249	&0.583	&0.383&	0.333	&0.348	&0.116&0.242&	0.575	&0.500	&0.362&	0.392&	0.615&	0.502	&0.231&	0.309 \\
&PatchTST & 0.334	&0.394	&0.204&	0.245&	0.843&	0.509&	0.442&	0.427	&0.094&	0.215	&0.603&	0.510&	0.346	&0.383&	0.440&	0.421	&0.200	&0.283 \\
& DLinear& 0.277 & 0.358	&0.462&	0.462&	1.347&	0.724&	0.648	&0.639	&1.419	&0.728&	0.874&	0.648&	1.830&	0.851&	0.867&	0.633&	1.548	&0.757 \\
& FEDformer &0.460&	0.490&	0.266	&0.344	&1.041&	0.608&	0.706	&0.621&	0.157&	0.289&	0.493&	0.477&	0.365&	0.409&	0.588	&0.513&	0.248&	0.339 \\
&Informer &0.834&	0.711	&0.390&	0.446	&1.340	&0.724&	0.471&	0.476	&0.932	&0.783&	0.988&	0.752&	1.784&	1.047&	0.779&	0.632&	0.952&	0.734 \\
&Autoformer& 0.397&	0.457	&0.264&	0.340	&1.046&	0.615&	0.871	&0.716	&0.171&	0.304&	0.496&	0.486&	0.383&	0.421&	0.655&	0.538&	0.250&	0.336 \\
 \hline 
\multicolumn{2}{c|}{\# of Variates} & \multicolumn{2}{c|}{321} & \multicolumn{2}{c|}{21} & \multicolumn{2}{c|}{862} & \multicolumn{2}{c|}{137} & \multicolumn{2}{c|}{8} & \multicolumn{2}{c|}{7} & \multicolumn{2}{c|}{7} & \multicolumn{2}{c|}{7} & \multicolumn{2}{c}{7} \\ 
 
\end{tabular}
\caption{Results of \name on various benchmarks, compared to baseline models. \name has been fine-tuned using specified data budgets, with MSE and MAE scores reported. The best results are highlighted in bold, and both input and prediction lengths are set at 96. \name demonstrates remarkable performance in budget-limited settings, as well as with the full dataset, particularly in scenarios involving a large number of variates. }
\label{tbl:mse_main_app}
\end{table*}

\begin{table*}[t]
\centering
\small
\setlength{\tabcolsep}{0.6mm}
\begin{tabular}{ll|lr|rl|rl|rl|rl|rl|rl|ll|}
\hline
\multicolumn{2}{c|}{Prediction Length} & \multicolumn{2}{c|}{6} & \multicolumn{2}{c|}{8} & \multicolumn{2}{c|}{14} & \multicolumn{2}{c|}{18} & \multicolumn{2}{c|}{24} & \multicolumn{2}{c|}{36} & \multicolumn{2}{c|}{48} & \multicolumn{2}{c|}{Average} \\ \cline{3-18} 
\multicolumn{2}{c|}{Metric} & MSE & MAE & MSE & MAE & MSE & MAE & MSE & MAE & MSE & MAE & MSE & MAE & MSE & MAE & MSE & MAE \\ \hline
{\multirow{6}{*}{\rotatebox[origin=c]{90}{TimePFN-96 }}} & Exchange & 0.015 & 0.094 & 0.017 & 0.102 & 0.026 & 0.124 & 0.030 & 0.136 & 0.037 & 0.149 & 0.052 & 0.174 & 0.065 & 0.195 & 0.034 & 0.139 \\
 & Weather & 0.020 & 1.006 & 0.023 & 1.068 & 0.034 & 1.280 & 0.041 & 1.411 & 0.051 & 1.582 & 0.072 & 1.885 & 0.087 & 2.108 & 0.046 & 1.477 \\
 & Traffic & 0.388 & 0.489 & 0.397 & 0.496 & 0.409 & 0.506 & 0.408 & 0.499 & 0.413 & 0.499 & 0.436 & 0.516 & 0.449 & 0.520 & 0.414 & 0.503 \\
 & ECL & 0.368 & 0.471 & 0.410 & 0.497 & 0.497 & 0.550 & 0.518 & 0.560 & 0.539 & 0.571 & 0.602 & 0.600 & 0.629 & 0.600 & 0.509 & 0.549 \\
 & ETTh1 & 0.017 & 0.101 & 0.020 & 0.109 & 0.026 & 0.126 & 0.030 & 0.134 & 0.034 & 0.143 & 0.041 & 0.156 & 0.045 & 0.163 & 0.030 & 0.133 \\
 & ETTh2 & 0.059 & 0.185 & 0.067 & 0.198 & 0.082 & 0.220 & 0.087 & 0.227 & 0.091 & 0.234 & 0.104 & 0.249 & 0.112 & 0.260 & 0.086 & 0.224 \\ \hline

 \hline \parbox[t]{2mm}{\multirow{6}{*}{\rotatebox[origin=c]{90}{TimePFN-36 }}} & Exchange & 0.012 & 0.087 & 0.014 & 0.094 & 0.020 & 0.112 & 0.024 & 0.122 & 0.030 & 0.134 & 0.041 & 0.155 & 0.052 & 0.174 & 0.027 & 0.125 \\
 & Weather $\times 10^{2}$ & 0.017 & 0.932 & 0.020 & 0.991 & 0.029 & 1.174 & 0.036 & 1.301 & 0.046 & 1.470 & 0.065 & 1.775 & 0.081 & 2.024 & 0.042 & 1.381 \\
 & Traffic & 1.393 & 1.008 & \textbf{1.528} & \textbf{1.051} & \textbf{1.644} & \textbf{1.084} & \textbf{1.520} & \textbf{1.031} & \textbf{1.403} & \textbf{0.988} & \textbf{1.538} & \textbf{1.039} & \textbf{1.495} & \textbf{1.027} & \textbf{1.503} & \textbf{1.032} \\
 & ECL & 0.585 & 0.621 & \textbf{0.640} & 0.649 & \textbf{0.745} & \textbf{0.701} & \textbf{0.747} & \textbf{0.702} & \textbf{0.760} & \textbf{0.712} & \textbf{0.878} & \textbf{0.764} & \textbf{0.909} & \textbf{0.772} & \textbf{0.752} & \textbf{0.703} \\
 & ETTh1 & 0.018 & 0.100 & 0.020 & 0.107 & 0.025 & 0.121 & \textbf{0.028} & \textbf{0.128} & \textbf{0.032} & \textbf{0.137} & \textbf{0.040} & \textbf{0.153} & \textbf{0.045} & \textbf{0.164} & \textbf{0.029} & 0.130 \\
 & ETTh2 & 0.100 & 0.241 & \textbf{0.110} & 0.253 & \textbf{0.126} & \textbf{0.274} & \textbf{0.125} & \textbf{0.274} & \textbf{0.126} & \textbf{0.275} & \textbf{0.145} & \textbf{0.295} & \textbf{0.152} & \textbf{0.302} & \textbf{0.126} & \textbf{0.273} \\ \hline

 \hline \parbox[t]{2mm}{\multirow{6}{*}{\rotatebox[origin=c]{90}{ForecastPFN }}} & Exchange & 0.041 & 0.154 & 0.042 & 0.158 & 0.049 & 0.169 & 0.054 & 0.177 & 0.061 & 0.187 & 0.072 & 0.201 & 0.084 & 0.215 & 0.057 & 0.180 \\
 & Weather $\times 10^{2}$ & 0.062 & 1.668 & 0.065 & 1.719 & 0.074 & 1.865 & 0.080 & 1.952 & 0.089 & 2.073 & 0.103 & 2.278 & 0.115 & 2.443 & 0.084 & 1.999 \\
 & Traffic & 4.690 & 1.779 & 4.712 & 1.790 & 4.572 & 1.765 & 4.428 & 1.724 & 4.348 & 1.698 & 4.504 & 1.735 & 4.394 & 1.703 & 4.521 & 1.742 \\
 & ECL & 1.430 & 0.962 & 1.444 & 0.969 & 1.406 & 0.955 & 1.360 & 0.935 & 1.356 & 0.935 & 1.453 & 0.973 & 1.467 & 0.977 & 1.416 & 0.958 \\
 & ETTh1 & 0.085 & 0.216 & 0.087 & 0.220 & 0.093 & 0.228 & 0.097 & 0.232 & 0.104 & 0.239 & 0.119 & 0.256 & 0.131 & 0.270 & 0.102 & 0.237 \\
 & ETTh2 & 0.409 & 0.504 & 0.418 & 0.510 & 0.424 & 0.513 & 0.421 & 0.509 & 0.426 & 0.511 & 0.462 & 0.532 & 0.481 & 0.540 & 0.434 & 0.517 \\ \hline

\hline \parbox[t]{2mm}{\multirow{6}{*}{\rotatebox[origin=c]{90}{Chronos-Small }}} & Exchange & 0.026 & 0.072 & 0.048 & 0.081 & 0.079 & 0.104 & 0.020 & 0.107 & 0.059 & 0.124 & \textbf{0.034} & 0.141 & 0.075 & 0.165 & 0.049 & 0.113 \\
 & Weather $\times 10^{2}$ & 0.013 & 0.623 & \textbf{0.014} & 0.703 & \textbf{0.023} & 0.920 & 0.030 & 1.055 & 0.041 & 1.244 & \textbf{0.058} & 1.568 & 0.078 & 1.842 & 0.036 & 1.136 \\
 & Traffic & \textbf{1.298} & \textbf{0.819} & 1.997 & 1.056 & 3.738 & 1.530 & 4.063 & 1.642 & 3.545 & 1.502 & 3.434 & 1.482 & 3.646 & 1.519 & 3.103 & 1.364 \\
 & ECL & \textbf{0.473} & \textbf{0.488} & 0.698 & \textbf{0.601} & 1.313 & 0.856 & 1.443 & 0.914 & 1.310 & 0.865 & 1.371 & 0.893 & 1.458 & 0.931 & 1.152 & 0.792 \\
 & ETTh1 & 0.045 & 0.114 & 0.044 & 0.121 & 0.062 & 0.151 & 0.065 & 0.159 & 0.065 & 0.168 & 0.073 & 0.184 & 0.076 & 0.194 & 0.061 & 0.155 \\
 & ETTh2 & \textbf{0.089} & \textbf{0.188} & 0.134 & \textbf{0.238} & 0.227 & 0.337 & 0.251 & 0.365 & 0.235 & 0.358 & 0.250 & 0.374 & 0.266 & 0.393 & 0.207 & 0.321 \\ \hline

  \hline \parbox[t]{2mm}{\multirow{6}{*}{\rotatebox[origin=c]{90}{SeasonalNaive }}} & Exchange & 0.015 & 0.096 & 0.016 & 0.100 & 0.021 & 0.114 & 0.025 & 0.124 & 0.030 & 0.135 & 0.039 & 0.154 & 0.050 & 0.172 & 0.028 & 0.128 \\
 & Weather $\times 10^{2}$ & 0.021 & 0.907 & 0.023 & 0.965 & 0.031 & 1.137 & 0.039 & 1.278 & 0.048 & 1.445 & 0.067 & 1.740 & 0.084 & 1.989 & 0.045 & 1.352 \\
 & Traffic & 4.354 & 1.850 & 4.581 & 1.891 & 5.263 & 2.016 & 4.416 & 1.784 & 3.756 & 1.614 & 4.104 & 1.691 & 3.631 & 1.548 & 4.301 & 1.771 \\
 & ECL & 1.427 & 0.962  & 1.523 & 0.994 & 1.810 & 1.092 & 1.590 & 1.004 & 1.427 & 0.942 & 1.600 & 1.001 & 1.533 & 0.973 & 1.559 & 0.995 \\
 & ETTh1 & 0.027 & 0.126 & 0.029 & 0.131 & 0.035 & 0.145 & 0.037 & 0.149 & 0.040 & 0.156 & 0.049 & 0.171 & 0.055 & 0.181 & 0.039 & 0.151 \\
 & ETTh2 &0.272 & 0.394 & 0.283 & 0.405 & 0.313 & 0.437 & 0.278 & 0.409 & 0.254 & 0.390 & 0.279 & 0.413  & 0.273 & 0.406 & 0.279 & 0.408 \\ \hline

 \hline \parbox[t]{2mm}{\multirow{6}{*}{\rotatebox[origin=c]{90}{Naive }}} & Exchange & \textbf{0.008} & \textbf{0.064} & \textbf{0.010} & \textbf{0.073} & \textbf{0.015} & \textbf{0.093} & \textbf{0.019} & \textbf{0.104} & \textbf{0.024} & \textbf{0.118} & \textbf{0.034} & \textbf{0.140} & \textbf{0.045} & \textbf{0.160} & \textbf{0.022} & \textbf{0.107} \\
 & Weather $\times 10^{2}$ & \textbf{0.011} & \textbf{0.598} & \textbf{0.014} & \textbf{0.685} & \textbf{0.023} & \textbf{0.910} & \textbf{0.029} & \textbf{1.044} & \textbf{0.038} & \textbf{1.232} & \textbf{0.058} & \textbf{1.561} & \textbf{0.075} & \textbf{1.834} & \textbf{0.035} & \textbf{1.123} \\
 & Traffic & 1.759 & 1.041 & 2.495 & 1.263 & 4.090 & 1.661 & 4.245 & 1.716 & 3.524 & 1.504 & 3.622 & 1.536 & 3.574 & 1.517 & 3.330 & 1.463 \\
 & ECL & 0.586 & 0.560 & 0.827 & 0.672 & 1.400 & 0.904 & 1.479 & 0.941 & 1.309 & 0.874 & 1.406 & 0.914 & 1.471 & 0.938 & 1.211 & 0.829 \\
 & ETTh1 & \textbf{0.014} & \textbf{0.084} & \textbf{0.018} & \textbf{0.095} & \textbf{0.027} & \textbf{0.120} & 0.031 & 0.131 & 0.034 & 0.139 & 0.043 & 0.157 & 0.050 & 0.171 & 0.031 & \textbf{0.128} \\
 & ETTh2 &0.114&0.226& 0.157 & 0.272 & 0.240 & 0.357 & 0.256 & 0.376 & 0.229 & 0.357 & 0.248 & 0.377  & 0.259 & 0.390 & 0.215 & 0.336 \\ \hline
 
 \hline \parbox[t]{2mm}{\multirow{6}{*}{\rotatebox[origin=c]{90}{Mean }}} & Exchange & 0.027 & 0.131 & 0.028 & 0.135 & 0.033 & 0.146 & 0.037 & 0.152 & 0.042 & 0.161 & 0.052 & 0.177 & 0.062 & 0.192 & 0.040 & 0.156 \\
 & Weather $\times 10^{2}$ & 0.047 & 1.546 & 0.050 & 1.599 & 0.059 & 1.775 & 0.064 & 1.840 & 0.074 & 1.966 & 0.089 & 2.178 & 0.101 & 2.345 & 0.069 & 1.893 \\
 & Traffic & 2.293 & 1.332 & 2.350 & 1.343 & 2.233 & 1.287 & 2.049 & 1.221 & 1.920 & 1.183 & 2.078 & 1.234 & 1.955 & 1.192 & 2.125 & 1.256 \\
 & ECL & 0.923 & 0.793 & 0.955 & 0.805 & 0.960 & 0.805 & 0.929 & 0.790 & 0.923 & 0.788 & 1.025 & 0.828 & 1.029 & 0.827 & 0.963 & 0.805 \\
 & ETTh1 & 0.031 & 0.135 & 0.033 & 0.138 & 0.036 & 0.146 & 0.038 & 0.151 & 0.041 & 0.157 & 0.048 & 0.170 & 0.052 & 0.178 & 0.040 & 0.154 \\
 & ETTh2 & 0.161 & 0.314 & 0.166 & 0.320 & 0.167 & 0.321 & 0.162 & 0.315 & 0.162 & 0.314 & 0.179 & 0.330 & 0.182 & 0.333 & 0.168 & 0.321 \\ \hline
 
\end{tabular}
\caption{Zero-shot results of TimePFN on univariate time-series forecasting with input length = 36. TimePFN-96 has input length of 96. All other baselines have input length 36. Meta-N-Beats is not included as it is not our implementation. }
\label{tbl:uni_zero_shot}
\end{table*}

In addition to the budget scenarios presented in the main body, we also conducted experiments with data budgets of 100 and 1,000 to fully characterize our experimental results. Furthermore, the average accuracy across these data budgets is provided for reference. Table 5 showcases all these evaluations. In Table 6, we present the raw results of the univariate forecasting task for zero-shot forecasting.

\subsection{Multivariate Forecasting}
As shown in Table 5, \name consistently achieves the best results with a data budget of 100 and significantly outperforms all other models with a budget of 1,000, leading in 7 out of 9 datasets. \name excels particularly in datasets with a multivariate nature. Consider that PatchTST \cite{Yuqietal-2023-PatchTST} assumes channel independence, whereas iTransformer \cite{liu2023itransformer} treats each variate as a token, demonstrating extreme channel dependence. In the full budget scenario, where the entire dataset is utilized, the difference in forecasting performance between iTransformer and PatchTST is revealing, particularly in detecting datasets with high inter-channel dependencies. For instance, in the ECL and Traffic datasets, which have a large number of variates (which does not mean high channel dependence by itself), iTransformer shows superior forecasting performance compared to PatchTST. Conversely, in the ETT datasets, PatchTST performs comparatively better. Extrapolating from there, we realize that \name excels in datasets with a high multivariate nature, even in full budget scenarios, and also yields good and competitive performance in datasets with comparatively low multivariate characteristic in full budget scenarios. With limited budgets, we see that \name is the leading model among the baselines.  

\subsection{Univariate Forecasting}
Although \name is specifically designed for multivariate time series forecasting, we also assessed its performance in zero-shot univariate forecasting, compared to similar models. See Table 6 for more details. On average, \name-36 is the most successful model among other models, and uniformly better than all other deep-learning based baselines in our setting. Generally, Chronos-small \cite{ansari2024chronos} outperforms \name-36 with shorter prediction lengths, while TimePFN-36 excels at longer prediction lengths, outperforming the other models. This outcome is expected, as \name is specifically trained to handle an input length of 96 and predict the same distance ahead. For these comparisons, we trimmed \name's predictions to match the given prediction lengths. Given \name's focus on longer prediction horizons, it's no surprise that Chronos-small performs better at shorter lengths. For \name-36, we padded the first 60 sequences of the 96-sequence input with the average of a 36-sequence input to minimize distribution shift. We also included results for \name-96, which uses the full 96-sequence input length without padding, to demonstrate our model’s complete performance.

\section{Datasets}
As datasets, we used 9 benchmark datasets which are commonly used in multivariate time-series forecasting. These consist of four ETT datasets \cite{haoyietal-informer-2021} (ETTh1, ETTh2, ETTm1, ETTm2), ECL (Electricity Consuming Load), Exchange, Traffic, Weather and Solar Energy.   Except for the Solar Energy datasets, the others are benchmarked by \cite{wu2021autoformer}, while the Solar is introduced by \cite{LSTNet}. We splitted the training, validation and test sets in a chronological way deterministically, following \cite{Yuqietal-2023-PatchTST, liu2023itransformer}.  We will briefly explain the features of these datasets in this section. 

\textbf{ETT Datasets.} The abbreviation ETT refers to Electricity Transformer Temperature \cite{haoyietal-informer-2021}. All ETT datasets consist of seven variates. The ETTh1 and ETTh2 datasets are sampled hourly, while the ETTm1 and ETTm2 datasets are sampled every 15 minutes. Specifically, the ETTh datasets contain 8545, 2881, and 2881 data points in the training, validation, and test sets, respectively. In contrast, the ETTm datasets comprise 34465, 11521, and 11521 data points in the training, validation, and test sets, respectively \cite{liu2023itransformer}.

\textbf{ECL Dataset.} The abbreviation ECL refers to the electricity consumption load of 321 users \cite{wu2021autoformer}. It is recorded in hourly intervals, resulting in a dataset with 321 variates. The ECL dataset contains 18317, 2633, and 5261 data points in the training, validation, and test sets, respectively \cite{liu2023itransformer}.

\textbf{Exchange Dataset.} This dataset provides daily exchange rates for eight countries \cite{wu2021autoformer}, comprising eight variates. The Exchange dataset includes 5120, 665, and 1422 data points in the training, validation, and test sets, respectively \cite{liu2023itransformer}. Some works, such as PatchTST \cite{Yuqietal-2023-PatchTST}, avoid using this dataset as a benchmark because simple naive predictions (using the last observed value) often outperform more complex methods. However, for completeness, we have included it in our analysis.

\textbf{Traffic Dataset.} This dataset includes hourly road occupancy rates from 862 locations \cite{wu2021autoformer}, resulting in 862 variates. The traffic dataset contains 12185, 1757, and 3509 data points in the training, validation, and test sets, respectively \cite{liu2023itransformer}. It is by far the most high-dimensional dataset in our evaluation. 

\textbf{Weather Dataset.} This dataset includes 21 meteorological factors collected every 10 minutes \cite{wu2021autoformer}, resulting in 21 variates. The weather dataset contains 36792, 5271, and 10540 data points in the training, validation, and test sets, respectively \cite{liu2023itransformer}.

\textbf{Solar-Energy Dataset.} This dataset includes power production values from 137 solar power plants, sampled every 10 minutes \cite{LSTNet}, resulting in 137 variables. The solar energy dataset contains 36601, 5161, and 10417 data points in the training, validation, and test sets, respectively \cite{liu2023itransformer}.  

\section{Additional Ablations}

\begin{table*}[t]
\centering
\small
\setlength{\tabcolsep}{0.3mm}
\begin{tabular}{ll|lr|rl|rl|rl|rl|rl|rl|ll|ll}
\hline
\multicolumn{2}{c|}{Dataset} & \multicolumn{2}{c|}{ECL} & \multicolumn{2}{c|}{Weather} & \multicolumn{2}{c|}{Traffic} & \multicolumn{2}{c|}{Solar-Energy} & \multicolumn{2}{c|}{Exchange} & \multicolumn{2}{c|}{ETTh1} & \multicolumn{2}{c|}{ETTh2} & \multicolumn{2}{c|}{ETTm1} & \multicolumn{2}{c}{ETTm2} \\ \cline{3-20} 
\multicolumn{2}{c|}{Models} & MSE & MAE & MSE & MAE & MSE & MAE & MSE & MAE & MSE & MAE & MSE & MAE & MSE & MAE & MSE & MAE & MSE & MAE \\ \hline
\parbox[t]{2mm}{\multirow{2}{*}{\rotatebox[origin=c]{90}{50}}} & TimePFN & \textbf{0.235} & \textbf{0.322} & \textbf{0.190} & \textbf{0.235} & \textbf{0.746} & \textbf{0.468} & \textbf{0.429} & 0.450 & 
\textbf{0.096} & \textbf{0.218}  & \textbf{0.438} & \textbf{0.429} & \textbf{0.324} & \textbf{0.359} &  \textbf{0.419} & \textbf{0.418} & \textbf{0.195} & \textbf{0.276} \\
 &TimePFN-w/o-synthetic & 0.314 & 0.391 & 0.213 & 0.255 & 0.966 & 0.580 & 0.448 & \textbf{0.440} & 0.110 & 0.237 & 0.520 & 0.482 & 0.360 & 0.392  & 0.468 & 0.445 & 0.234 & 0.312 \\ \hline

 \parbox[t]{2mm}{\multirow{2}{*}{\rotatebox[origin=c]{90}{100}}} & TimePFN & \textbf{0.221} & \textbf{0.309} & \textbf{0.187} & \textbf{0.232} & \textbf{0.644} & \textbf{0.424} & 0.351 & 0.383 & \textbf{0.083} & \textbf{0.205}  & \textbf{0.441} & \textbf{0.429} & \textbf{0.322} & \textbf{0.356} &  0.412 & \textbf{0.411} & \textbf{0.196} & \textbf{0.273} \\
 &TimePFN-w/o-synthetic & 0.259 & 0.344 & 0.221 & 0.257 & 1.008 & 0.596 & \textbf{0.324} & \textbf{0.347} & 0.104 & 0.230 & 0.505 & 0.475 & 0.360 & 0.391  & \textbf{0.404} & \textbf{0.411} & 0.231 & 0.308 \\ \hline

 \parbox[t]{2mm}{\multirow{2}{*}{\rotatebox[origin=c]{90}{500}}} & TimePFN & \textbf{0.190} & \textbf{0.283} & \textbf{0.178} & \textbf{0.222} & \textbf{0.487} & \textbf{0.335} & \textbf{0.269} & \textbf{0.305} & \textbf{0.083} & \textbf{0.203} & \textbf{0.401} & \textbf{0.412} & \textbf{0.311} & \textbf{0.352} & 0.360 & 0.386 & \textbf{0.185} & \textbf{0.268} \\
 &TimePFN-w/o-synthetic & 0.220 & 0.311 & 0.191 & 0.235 & 0.914 & 0.559 & 0.303 & 0.318 & 0.105 & 0.232 & 0.423 & 0.431 & 0.354 & 0.387  & \textbf{0.357} & \textbf{0.383} & 0.229 & 0.307 \\ \hline

 \parbox[t]{2mm}{\multirow{2}{*}{\rotatebox[origin=c]{90}{1000}}} & TimePFN & \textbf{0.173} & \textbf{0.268} & \textbf{0.175} & \textbf{0.219} & \textbf{0.452} & \textbf{0.310} & \textbf{0.243} & \textbf{0.288} & \textbf{0.084} & \textbf{0.204} & \textbf{0.405} & \textbf{0.415} & \textbf{0.304} & \textbf{0.351} & 0.344 & 0.378 & \textbf{0.180} & \textbf{0.262} \\
 &TimePFN-w/o-synthetic & 0.187 & 0.285 & 0.182 & 0.224 & 0.907 & 0.556 & 0.278 & 0.302 & 0.109 & 0.233 & 0.409 & 0.425 & 0.352 & 0.387  & \textbf{0.341} & \textbf{0.374} & 0.206 & 0.289 \\ \hline

 \parbox[t]{2mm}{\multirow{2}{*}{\rotatebox[origin=c]{90}{All}}} & TimePFN & 0.138 & 0.137 & \textbf{0.166} & \textbf{0.208} & \textbf{0.392} & \textbf{0.260} & 0.203 & 0.219 & \textbf{0.100} & \textbf{0.223} & \textbf{0.402} & \textbf{0.417} & \textbf{0.293} & \textbf{0.343} & \textbf{0.392} & \textbf{0.402} & \textbf{0.180} & \textbf{0.262} \\
 &TimePFN-w/o-synthetic & \textbf{0.137} & \textbf{0.136} & 0.214 & 0.269 & 1.408 & 0.800 & \textbf{0.196} & \textbf{0.210} & 0.101 & 0.225 & 0.412 & 0.423 & 0.352 & 0.386  & 0.437 & 0.414 & 0.229 & 0.307 \\ \hline

\end{tabular}
\caption{In TimePFN-w/o-synthetic, we do not pretrain the TimePFN architecture. Instead, we directly train the architecture with given data budgets and compare it with the pretrained and fine-tuned TimePFN. Both the sequence lengths and the forecasting length are set to 96. }
\label{tbl:abl_TimePFN_synthetic}
\end{table*}

\begin{table*}[t]
\centering
\small
\setlength{\tabcolsep}{0.3mm}
\begin{tabular}{ll|lr|rl|rl|rl|rl|rl|rl|ll|ll}
\hline
\multicolumn{2}{c|}{Dataset} & \multicolumn{2}{c|}{ECL} & \multicolumn{2}{c|}{Weather} & \multicolumn{2}{c|}{Traffic} & \multicolumn{2}{c|}{Solar-Energy} & \multicolumn{2}{c|}{Exchange} & \multicolumn{2}{c|}{ETTh1} & \multicolumn{2}{c|}{ETTh2} & \multicolumn{2}{c|}{ETTm1} & \multicolumn{2}{c}{ETTm2} \\ \cline{3-20} 
\multicolumn{2}{c|}{Models} & MSE & MAE & MSE & MAE & MSE & MAE & MSE & MAE & MSE & MAE & MSE & MAE & MSE & MAE & MSE & MAE & MSE & MAE \\ \hline
\parbox[t]{2mm}{\multirow{4}{*}{\rotatebox[origin=c]{90}{z.s. }}} & TimePFN & \textbf{0.315} & \textbf{0.383} & \textbf{0.209} & \textbf{0.255} & \textbf{1.108} & \textbf{0.613} & \textbf{0.941} & \textbf{0.730} & \textbf{0.105} & \textbf{0.229} & \textbf{0.453} & \textbf{0.439} & 0.328 & \textbf{0.362} & \textbf{0.637} & \textbf{0.512} & \textbf{0.212}  & \textbf{0.291} \\
&TimePFN-w/o Conv & 0.653 & 0.637 & 0.221 & 0.271 & 1.287 & 0.757 & 1.197 & 0.829 & 0.111 & 0.237 & 0.608 & 0.517 & 0.338 & 0.374  & 0.771 & 0.565 & 0.224 & 0.307 \\ 
  & PatchTST-PFN & 0.470 & 0.522 & 0.212 & 0.262 & 1.172 & 0.702 & 1.014 & 0.787 & 0.108 & 0.231 & 0.554 & 0.501 & \textbf{0.322} & 0.366  & 0.746 & 0.560 & 0.215 & 0.301 \\  
  & iTransformer-PFN & 0.609 & 0.605 & 0.213 & 0.261 & 1.321 & 0.761 & 1.075 & 0.791 & 0.111 & 0.235 & 0.522 & 0.481 & 0.327 & 0.367  & 0.765 & 0.563 & 0.220 & 0.305 \\  \hline

\end{tabular}
\caption{Extended ablation results involving iTransformer-PFN. In iTransformer-PFN, we pre-train an iTransformer model to assess the suitability of our architecture for PFNs, using a sequence length and forecasting length of 96. In TimePFN-w/o-Convolution, we remove the convolutional operator typically applied to the initial input variables. In PatchTST-PFN, we train a PatchTST model to evaluate the importance of channel mixing and the suitability of our architecture for PFNs, also using a sequence length and forecasting length of 96. We evaluate the zero-shot performance. }
\label{tbl:itransformer-ablation}
\end{table*}

\begin{table*}[t]
\centering
\small
\setlength{\tabcolsep}{0.3mm}
\begin{tabular}{ll|lr|rl|rl|rl|rl|rl|rl|ll|ll}
\hline
\multicolumn{2}{c|}{Dataset} & \multicolumn{2}{c|}{ECL} & \multicolumn{2}{c|}{Weather} & \multicolumn{2}{c|}{Traffic} & \multicolumn{2}{c|}{Solar-Energy} & \multicolumn{2}{c|}{Exchange} & \multicolumn{2}{c|}{ETTh1} & \multicolumn{2}{c|}{ETTh2} & \multicolumn{2}{c|}{ETTm1} & \multicolumn{2}{c}{ETTm2} \\ \cline{3-20} 
\multicolumn{2}{c|}{Models} & MSE & MAE & MSE & MAE & MSE & MAE & MSE & MAE & MSE & MAE & MSE & MAE & MSE & MAE & MSE & MAE & MSE & MAE \\ \hline
\parbox[t]{2mm}{\multirow{2}{*}{\rotatebox[origin=c]{90}{50}}} & iTransformer-PFN & \textbf{0.255} & \textbf{0.344} & \textbf{0.215} & \textbf{0.250} & 1.292 & 0.725 & \textbf{0.392} & \textbf{0.412} & 
\textbf{0.102} & \textbf{0.230}  & \textbf{0.491} & \textbf{0.465} & \textbf{0.370} & \textbf{0.393} &  \textbf{0.466} & \textbf{0.440} & \textbf{0.217} & \textbf{0.301} \\
 & iTransformer & 0.278 & 0.360 & 0.237 & 0.278 & \textbf{0.801} & \textbf{0.499} & 0.513 & 0.479  & 0.145 & 0.275 & 0.838 & 0.617 & 0.410 & 0.422  & 0.884   & 0.608  & 0.268 & 0.337 \\ \hline

 \parbox[t]{2mm}{\multirow{2}{*}{\rotatebox[origin=c]{90}{100}}} & iTransformer-PFN & \textbf{0.227} & \textbf{0.318} & \textbf{0.213} & \textbf{0.249} & 1.216 & 0.702 & \textbf{0.336} & \textbf{0.355} & \textbf{0.101} & \textbf{0.226}  & \textbf{0.478} & \textbf{0.455} & \textbf{0.370} & \textbf{0.393} &  \textbf{0.430} & \textbf{0.424} & \textbf{0.204} & \textbf{0.287} \\
 & iTransformer & 0.253 & 0.337 & 0.220 & 0.263 & \textbf{0.740} & \textbf{0.468} & 0.369 & 0.387  & 0.138 & 0.268 & 0.728 & 0.574 & 0.401 & 0.418  & 0.816   & 0.586  & 0.260 & 0.331 \\ \hline

 \parbox[t]{2mm}{\multirow{2}{*}{\rotatebox[origin=c]{90}{500}}} & iTransformer-PFN & \textbf{0.184} & \textbf{0.276} & \textbf{0.200} & \textbf{0.241} & 1.201 & 0.698 & 0.309 & \textbf{0.333} & \textbf{0.096} & \textbf{0.223} & \textbf{0.452} & \textbf{0.444} & \textbf{0.319} & \textbf{0.362} & \textbf{0.375} & \textbf{0.394} & \textbf{0.199} & \textbf{0.284} \\
  & iTransformer & 0.200 & 0.284 & 0.211 & 0.248 & \textbf{0.514} & \textbf{0.354} & \textbf{0.307} & 0.334 & 0.113 & 0.239 & 0.489 & 0.470 & 0.361 & 0.394 & 0.569 & 0.494 & 0.231 & 0.310  \\ \hline

 \parbox[t]{2mm}{\multirow{2}{*}{\rotatebox[origin=c]{90}{1000}}} & iTransformer-PFN & \textbf{0.170} & \textbf{0.263} & \textbf{0.195} & \textbf{0.238} & 1.239 & 0.696 & 0.288 & 0.308 & \textbf{0.097} & \textbf{0.223} & \textbf{0.431} & \textbf{0.432} & \textbf{0.304} & \textbf{0.352} & \textbf{0.365} & \textbf{0.391} & \textbf{0.196} & \textbf{0.279} \\
  & iTransformer & 0.184 & 0.271 & 0.206 & 0.242 & \textbf{0.469} & \textbf{0.324}  & \textbf{0.276} & \textbf{0.309} & 0.100 & \textbf{0.223} & 0.433 & 0.436 & 0.336 & 0.379 & 0.464 & 0.444 & 0.211 & 0.294  \\ \hline

 \parbox[t]{2mm}{\multirow{2}{*}{\rotatebox[origin=c]{90}{All}}} & iTransformer-PFN & \textbf{0.147} & 0.240 & 0.209 & 0.259 & 1.408 & 0.800 & 0.231 & 0.262 & 0.104 & 0.233 & 0.424 & 0.428 & 0.350 & 0.385 & 0.666 & 0.533 & 0.223 & 0.302\\
& iTransformer & \textbf{0.147} & \textbf{0.239} & \textbf{0.175} & \textbf{0.215}  & \textbf{0.393} & \textbf{0.268} & \textbf{0.201} & \textbf{0.233} & \textbf{0.086}& \textbf{0.206}  & \textbf{0.387} & \textbf{0.405} & \textbf{0.300} & \textbf{0.349} & \textbf{0.342} & \textbf{0.376} & \textbf{0.185} & \textbf{0.272} \\ \hline

\end{tabular}
\caption{In the iTransformer-PFN, we pre-trained an iTransformer model on a large-scale synthetic dataset. The data budget evaluations involve fine-tuning the model using these data budgets. For the iTransformer evaluations, we utilized the official hyperparameters of the model. Both the sequence lengths and the forecasting lengths are set to 96. }
\label{tbl:iTransformer_PFN}
\end{table*}

In addition to the ablation studies we provided, we include three additional case studies, with one focusing on the performance of the architecture of \name when it is not pretrained synthetic data, and the second one focuses on the performace of iTransformer \cite{liu2023itransformer} when it is used as the PFN backbone in zero-shot setting. In the third case study, we evaluate this iTransformer-PFN compared to iTransformer in various data budgets to demonstrate the generality of the framework of large-scale synthetic training and to demonstrate the architectural novelties of \name. Table 7, Table 8  and Table 9 contain the results of those ablation studies, respectively. 

\textbf{Pretraining TimePFN with LMC-Synth.} To better understand the impact of synthetic training in \name, we removed the synthetic training component and directly trained the architecture, which we referred to as \name-w/o-synthetic. As shown in Table 7, the forecasting performance of \name is significantly better than that of \name-w/o-synthetic, demonstrating its contribution in both full budget and limited budget settings.

\textbf{iTransformer as PFN.} To evaluate the performance of other architectures with prior-data-fitting, we trained an iTransformer architecture with LMC-Synth in addition to PatchTST-PFN. As shown in Table 8, compared to other variations, \name demonstrates significantly better zero-shot forecasting capability, with uniformly better results than those of the competing architectures. This supports our architectural design principles, involving 1D convolutions and channel-mixing.

\textbf{iTransformer-PFN with Data Budgets.} To demonstrate the behavior of another architecture with synthetic training, we pre-trained the iTransformer \cite{liu2023itransformer} with data generated by LMC-Synth and fine-tuned it using specified data budgets. As shown in Table 9, synthetic pre-training improves the performance of the model in most cases, demonstrating the generality of the framework. However, the contribution is limited when using the entire dataset, in contrast to those instances with \name, thereby highlighting \name's superior performance.
\section{Baseline Details}
In addition to the main body, we would like to elaborate on how we aggregated the forecasting results from Chronos-small \cite{ansari2024chronos} and ForecastPFN \cite{dooley2023forecastpfn}. We chose Chronos-small because its parameter size is somewhat closer to \name's compared to the larger variations of Chronos. We opted not to use Chronos-Tiny to maintain a stronger baseline. \name has approximately 8.5 million parameters, whereas Chronos-small has 46 million parameters, significantly exceeding the parameter count of \name. As Chronos is a probabilistic model, we performed inference on each time-series data point three times with Chronos and averaged the results to obtain the final point forecast, except for the Exchange dataset. Our evaluation of Chronos on the Exchange dataset yielded unstable results with three inference iterations; therefore, we conducted five runs for this dataset. We aggregated the ForecastPFN results using their published model weights.
\section{Implementation Details}
We implemented \name entirely in PyTorch. We optimized the pretraining task using a synthetic dataset generated with LMC-Synth, employing the Adam optimizer \cite{KingBa15} and adhering to a one-cycle learning rate policy with a maximum learning rate of $lr=0.0005$ \cite{onecycleLR}. In the few-shot evaluations, we fine-tuned the \name with maximum $lr=0.0002$ using AdamW optimizer \cite{loshchilov2019decoupledweightdecayregularization} with one-cycle learning rate policy. In training \name with synthetic dataset, we observed that making model see the independently generated channels first, corresponding to the case with $C_i(\vct{t}) =  l_i(\vct{t})$, then introducing the inter-channel dependent data, significanly improves the learning speed. The explanation is simple, with the case with $C_i(\vct{t}) =  l_i(\vct{t})$, the model sees much more time-series patterns, as the time-series channels are all independently generated by Gaussian processes. Thus, after the model learns to make channel-independent decisions, we introduced the channel-dependent data, similar to curriculum learning scenario \cite{curriculum_learning}. Moreover, while training on synthetic data, we added a multiplicative Gaussian noise to each MTS data point as a regularization, with $\sigma = 0.1, mean = 1$.  In the end, \name is trained on 1.5 million MTS data points with 160 channels. 

In \name, we used the token embedding dimension of 256, and the latent space dimension of 1024, while the feed-forward network dimension is set to 512. We did not do any hyperparameter tuning and chose those values from checking similar works such as \cite{Yuqietal-2023-PatchTST, liu2023itransformer}. In fine-tuning, we always used the same learning rate with the same number of epochs accross different datasets (0.0002 and 8 epochs). Thus, we run the evaluations once. While doing all those, we fixed the seed to a random value of 2023.  

Throughout the experiments, we used a single L40S GPU. In addition the GPU, we had access to 128 GB of RAM and a 32-core CPU, which facilitated the acceleration of synthetic time-series data generation. Our codebase is developed based on \cite{liu2023itransformer}. Overall, we used approximately 300 GPU hours, as we conducted benchmarks not only for our own model but also for many others. We are providing the full source code for \name, including the synthetic data generation, architecture, and the training and evaluation scripts. Furthermore, we are providing the model weights of \name.

\section{Visualizations}

\textbf{Visualizations for LMC-Synth.} We provide visualizations for the multivariate synthetic data generated by LMC-Synth. For clarity, we have limited the number of channels to 5 and the sequence length to 96. Figures 3-6 showcase the MTS data generated using LMC-Synth.

\textbf{The Forecasts of TimePFN.} We provide forecasts from \name under various data budgets and datasets, including zero-shot and full-data scenarios. Figures 7-18 display the forecasts of \name in these settings. As shown, with an increasing data budget, \name's forecasts align more closely with the ground truth.

\begin{figure*}
    \centering
    \includegraphics[width=1\linewidth]{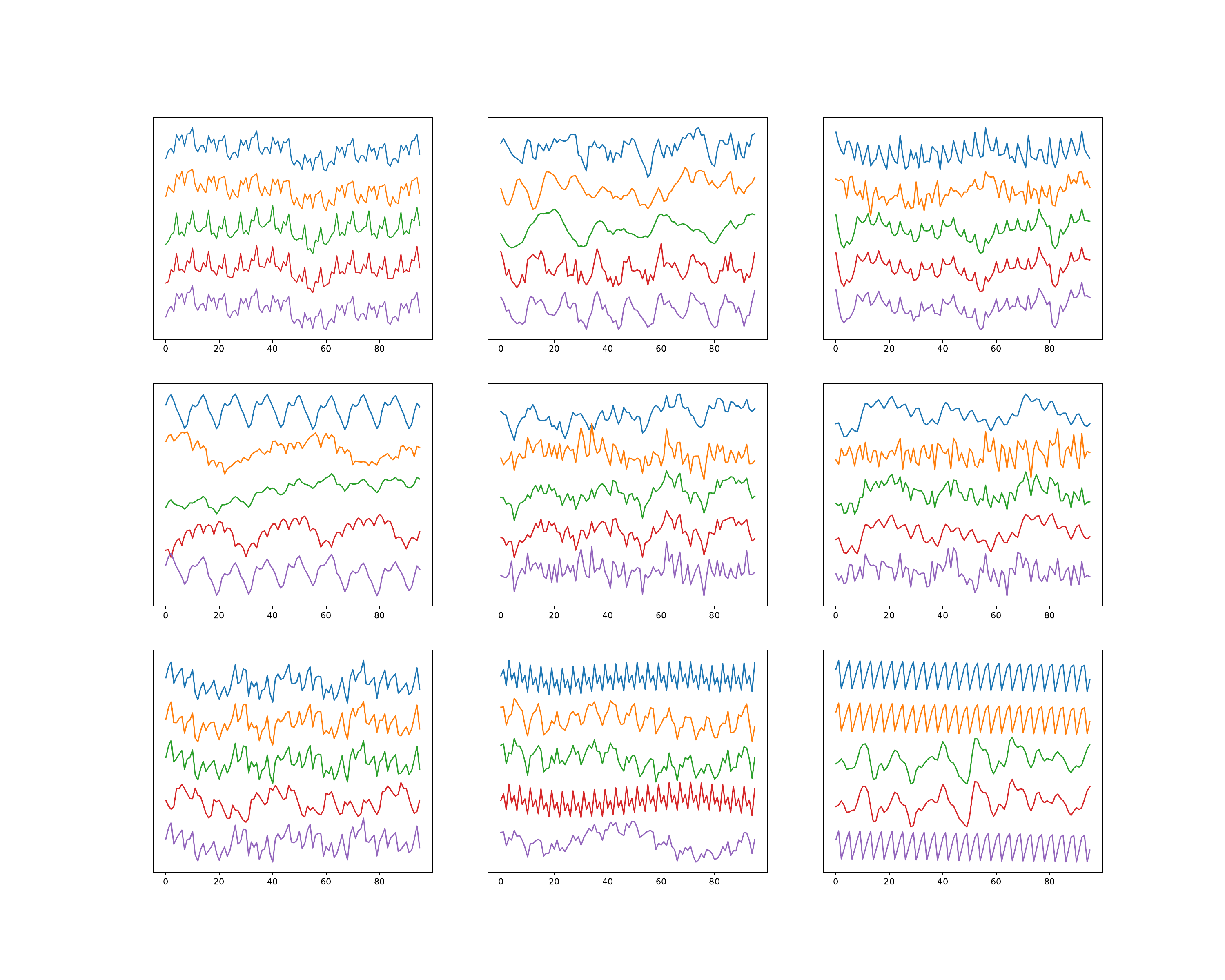}
    \caption{Examples of synthetic multivariate time time-series data generated by LMC-Synth. For the ease of understanding, we took C=5 and sequence lenght = 96. Dirichlet concentration parameter controls the diversity of variates from one another. }
    \label{fig:LMC-synth1}
\end{figure*}

\begin{figure*}
    \centering
    \includegraphics[width=1\linewidth]{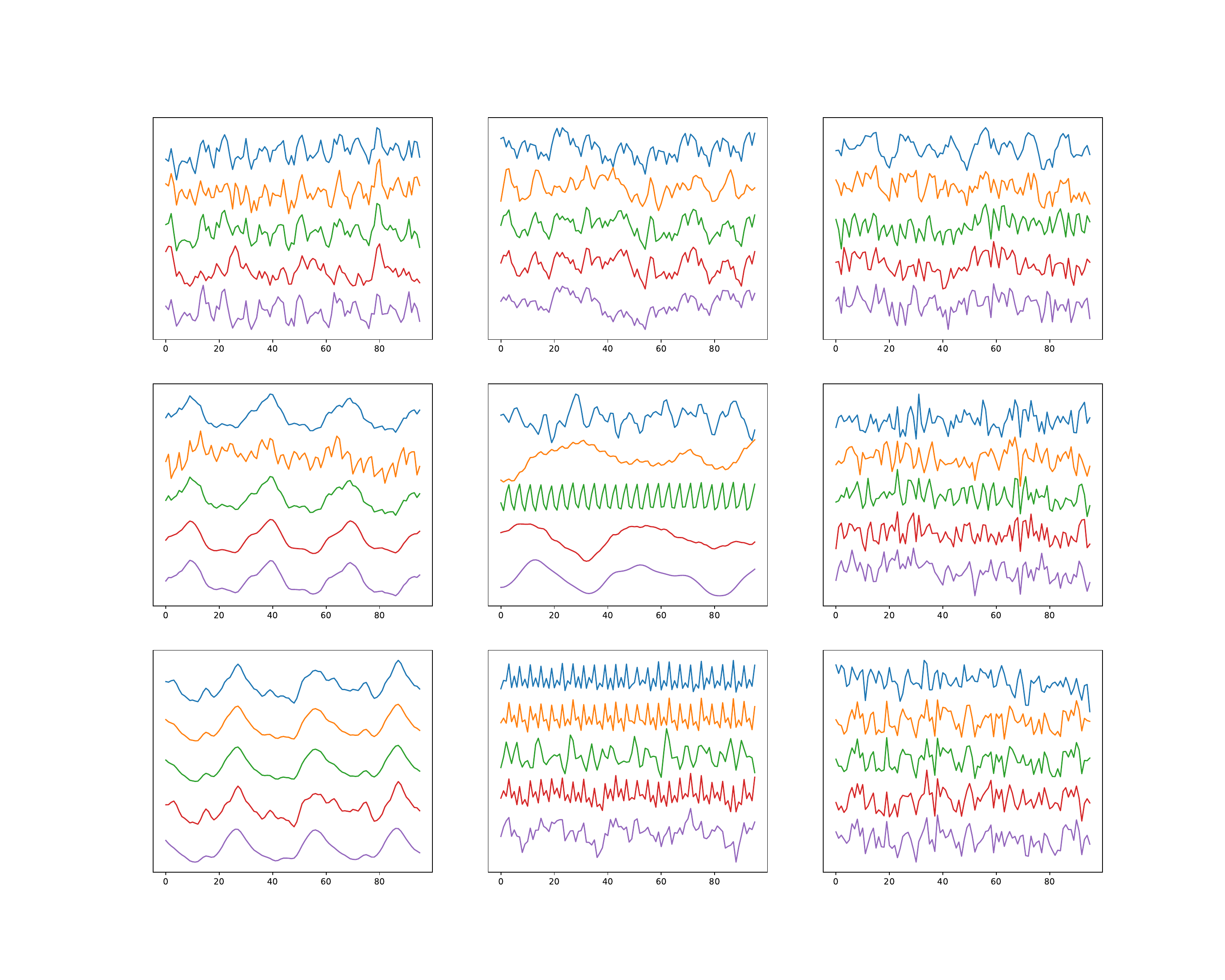}
    \caption{Examples of synthetic multivariate time time-series data generated by LMC-Synth. For the ease of understanding, we took C=5 and sequence lenght = 96. Dirichlet concentration parameter controls the diversity of variates from one another. }    \label{fig:LMC-synth2}
\end{figure*}

\begin{figure*}
    \centering
    \includegraphics[width=1\linewidth]{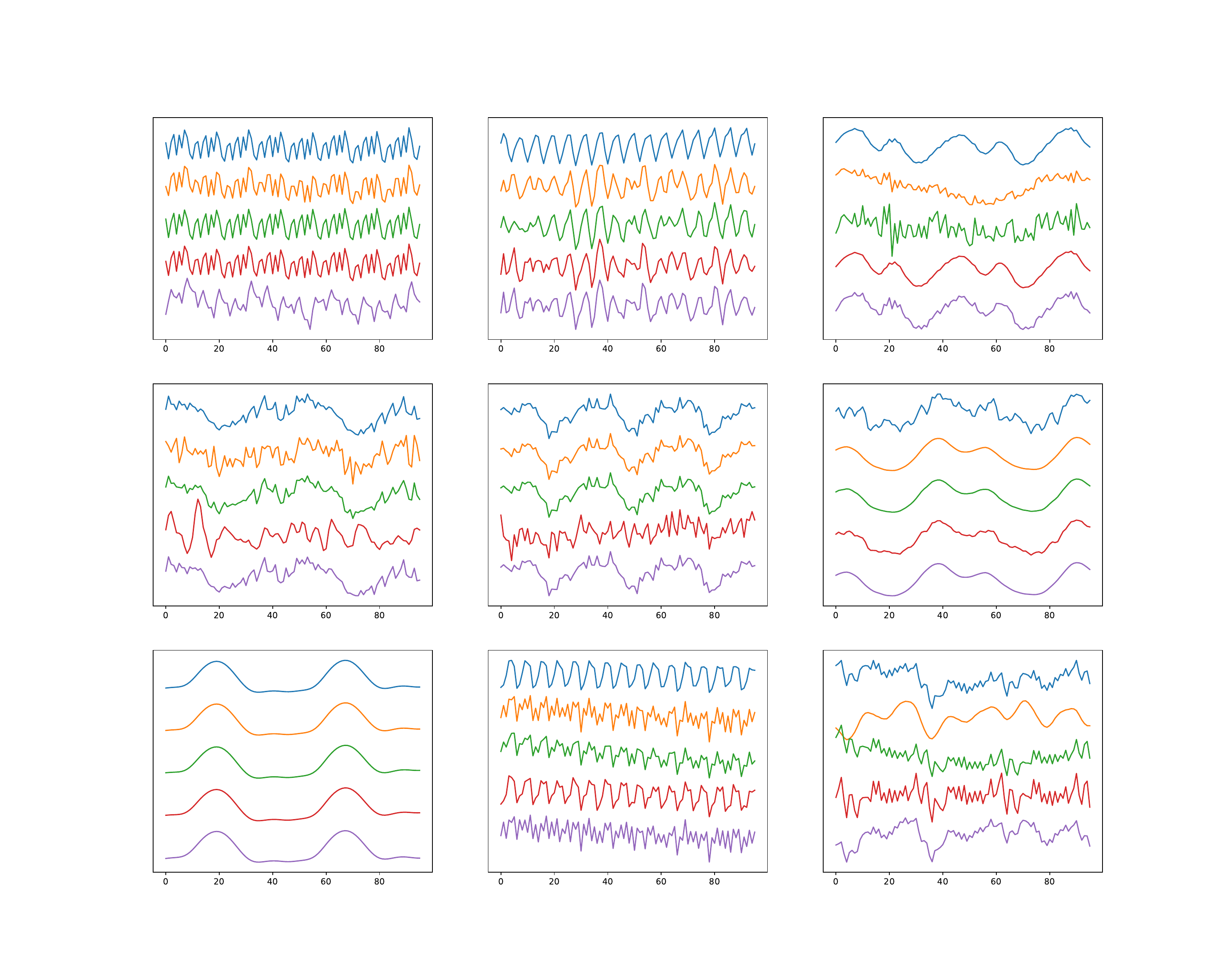}
    \caption{Examples of synthetic multivariate time time-series data generated by LMC-Synth. For the ease of understanding, we took C=5 and sequence lenght = 96. Dirichlet concentration parameter controls the diversity of variates from one another. }    \label{fig:LMC-synth3}
\end{figure*}

\begin{figure*}
    \centering
    \includegraphics[width=1\linewidth]{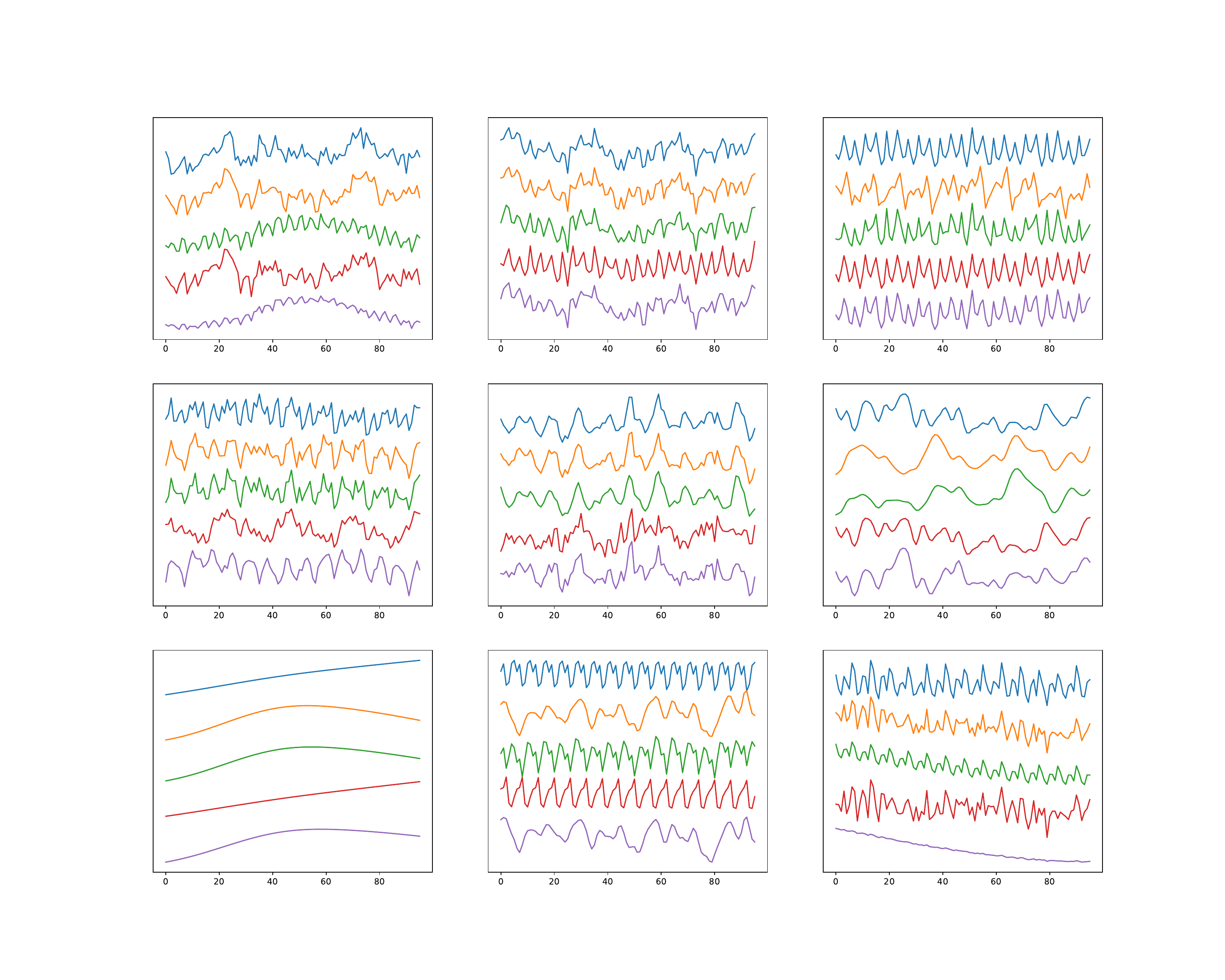}
    \caption{Examples of synthetic multivariate time time-series data generated by LMC-Synth. For the ease of understanding, we took C=5 and sequence lenght = 96. Dirichlet concentration parameter controls the diversity of variates from one another. }    \label{fig:LMC-synth4}
\end{figure*}

\begin{figure*}[htp]
\centering

\begin{subfigure}{0.32\textwidth}
\includegraphics[width=\linewidth]{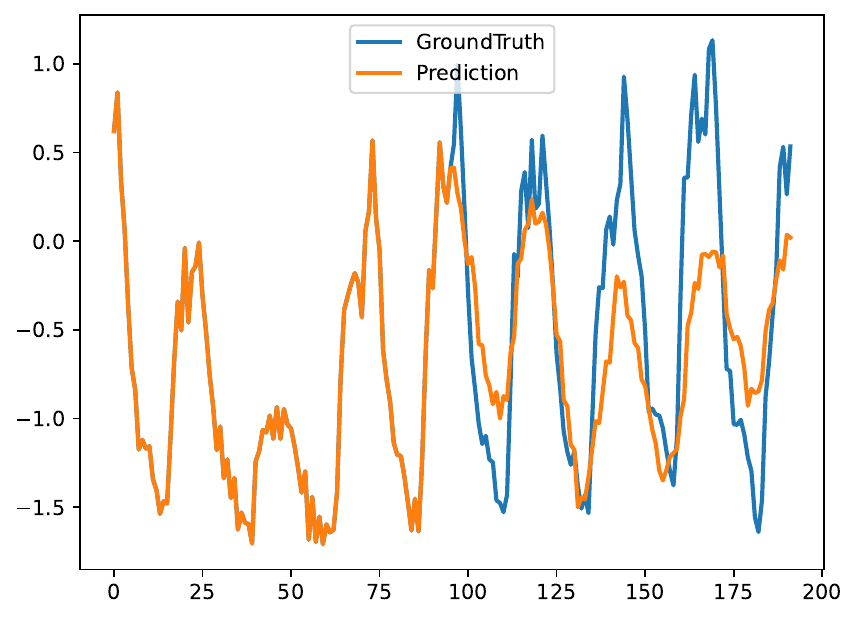}
\caption{Zero Shot}
\end{subfigure}\hfill
\begin{subfigure}{0.32\textwidth}
\includegraphics[width=\linewidth]{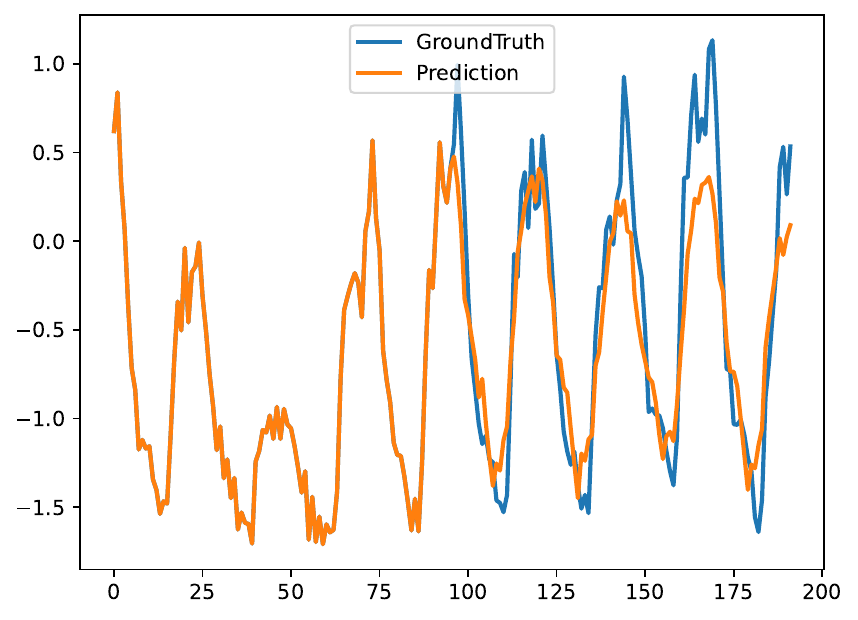}
\caption{Budget = 50}
\end{subfigure}\hfill
\begin{subfigure}{0.32\textwidth}
\includegraphics[width=\linewidth]{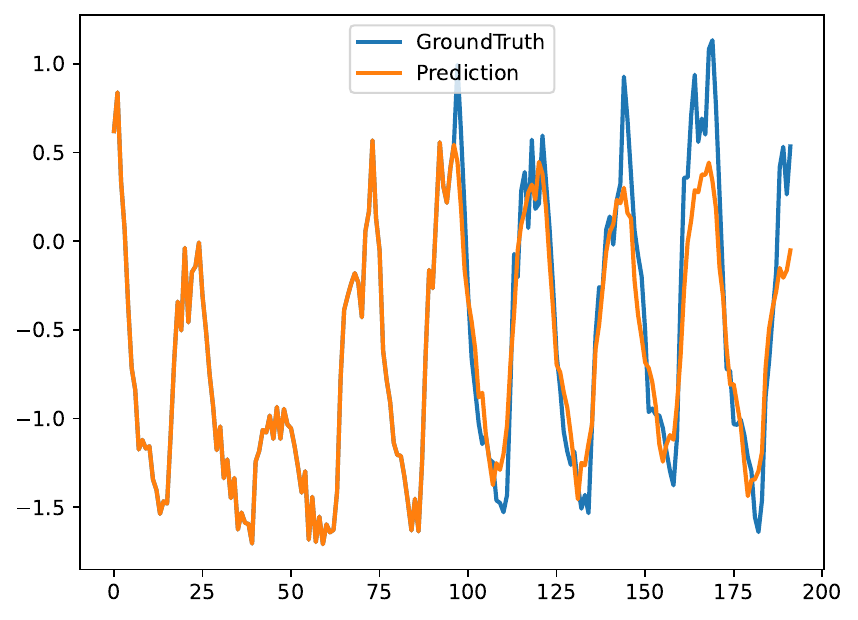}
\caption{Budget = 100}
\end{subfigure}

\begin{subfigure}{0.32\textwidth}
\includegraphics[width=\linewidth]{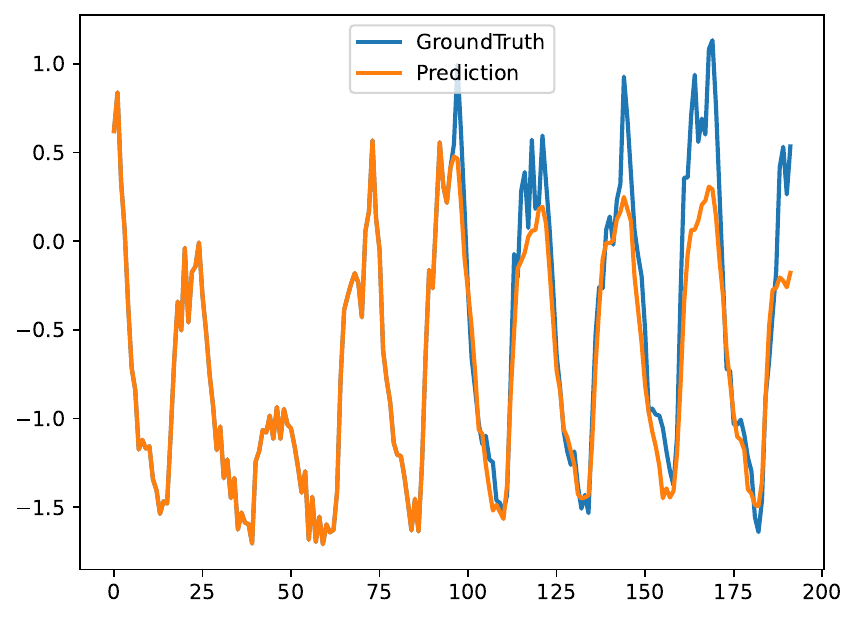}
\caption{Few-shot with budget 500}
\end{subfigure}\hfill
\begin{subfigure}{0.32\textwidth}
\includegraphics[width=\linewidth]{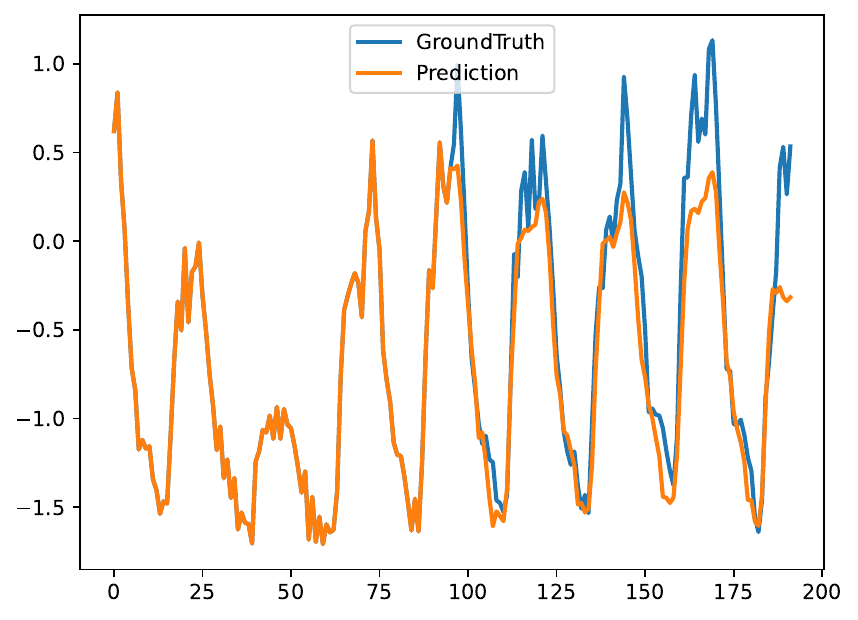}
\caption{Budget = 1000}
\end{subfigure}\hfill
\begin{subfigure}{0.32\textwidth}
\includegraphics[width=\linewidth]{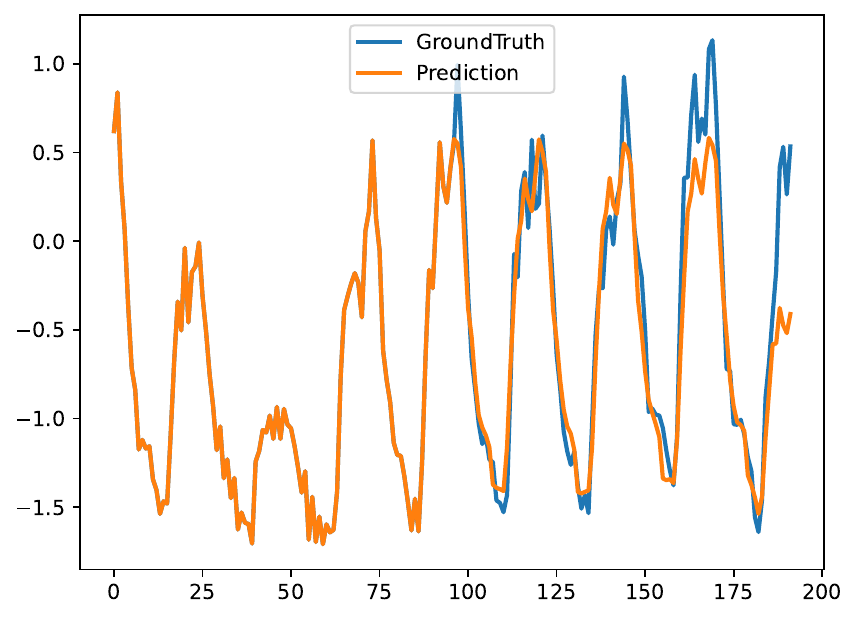}
\caption{Budget = All}
\end{subfigure}

\caption{The forecasts of TimePFN with various data budgets on ECL dataset. }
\label{fig:ECL20}
\end{figure*}
\begin{figure*}[htp]
\centering

\begin{subfigure}{0.32\textwidth}
\includegraphics[width=\linewidth]{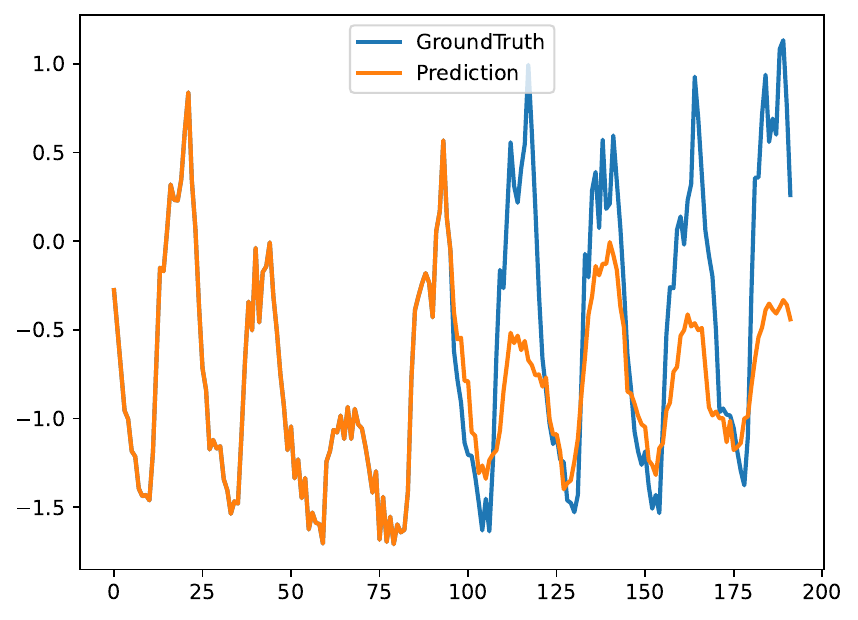}
\caption{Zero Shot}
\end{subfigure}\hfill
\begin{subfigure}{0.32\textwidth}
\includegraphics[width=\linewidth]{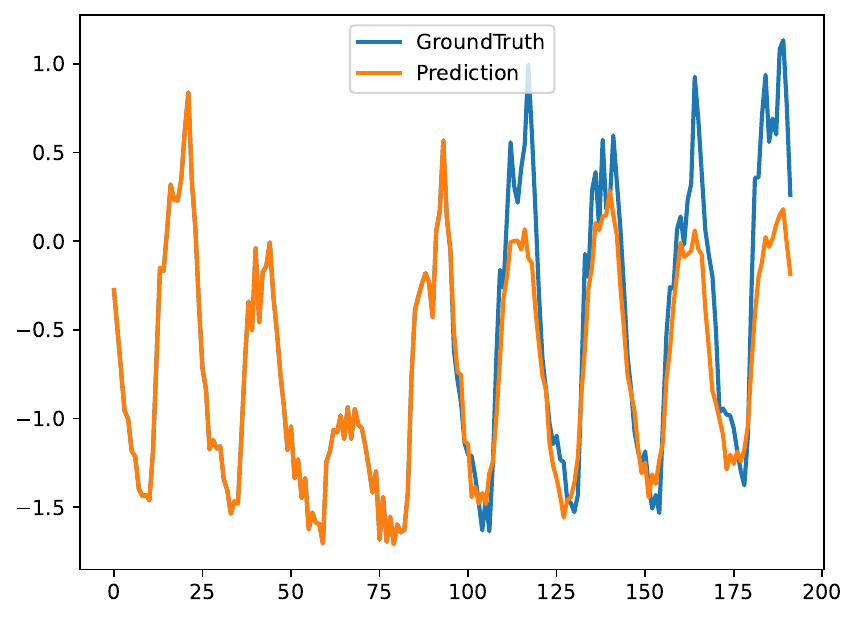}
\caption{Budget = 50}
\end{subfigure}\hfill
\begin{subfigure}{0.32\textwidth}
\includegraphics[width=\linewidth]{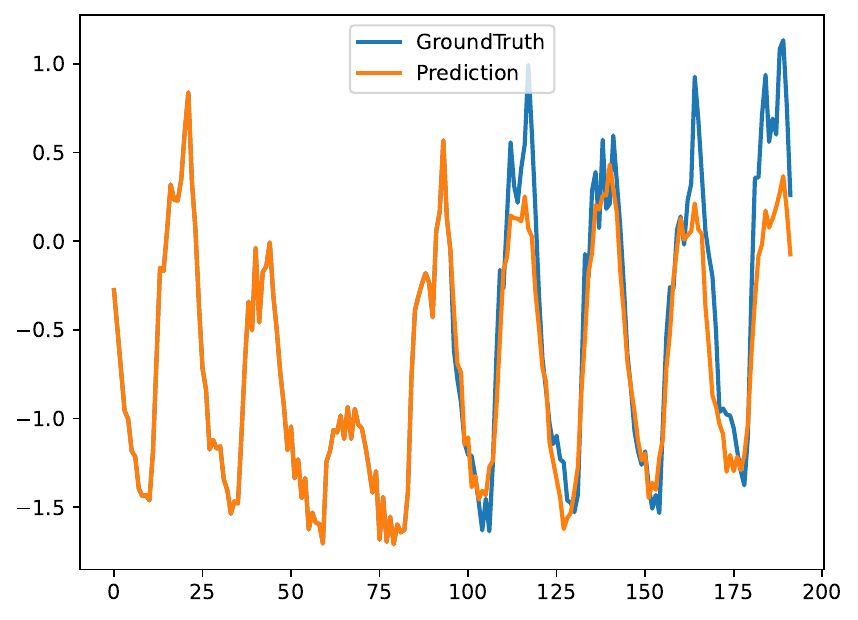}
\caption{Budget = 100}
\end{subfigure}

\begin{subfigure}{0.32\textwidth}
\includegraphics[width=\linewidth]{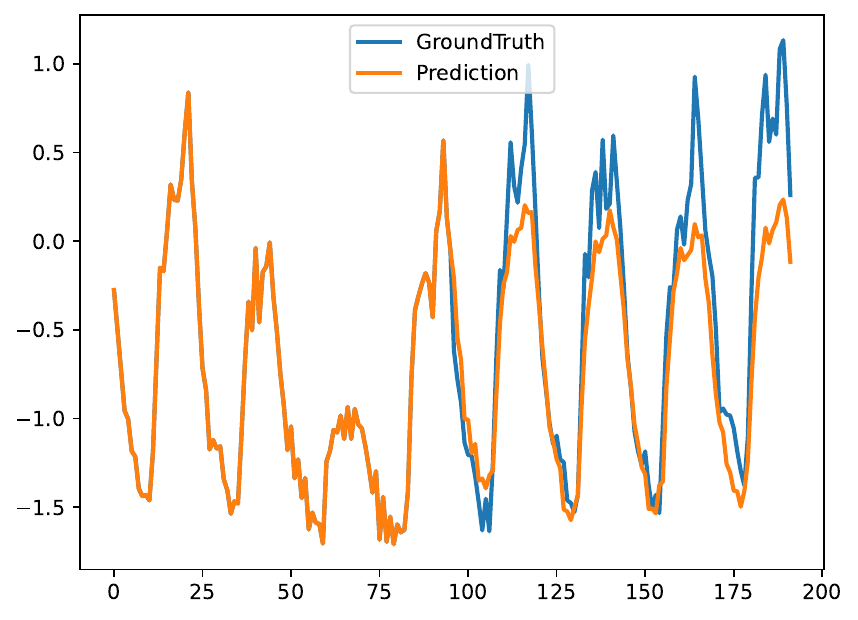}
\caption{Budget = 500}
\end{subfigure}\hfill
\begin{subfigure}{0.32\textwidth}
\includegraphics[width=\linewidth]{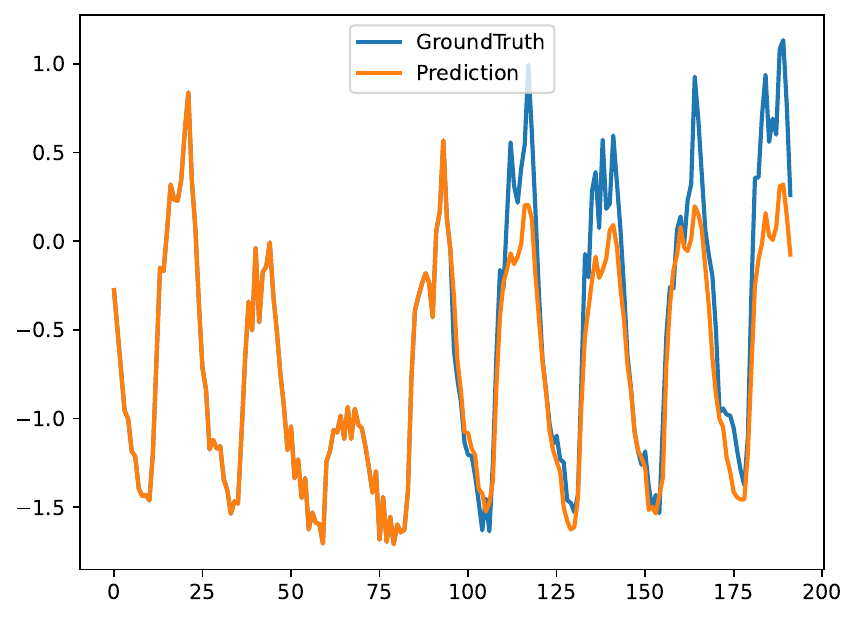}
\caption{Budget = 1000}
\end{subfigure}\hfill
\begin{subfigure}{0.32\textwidth}
\includegraphics[width=\linewidth]{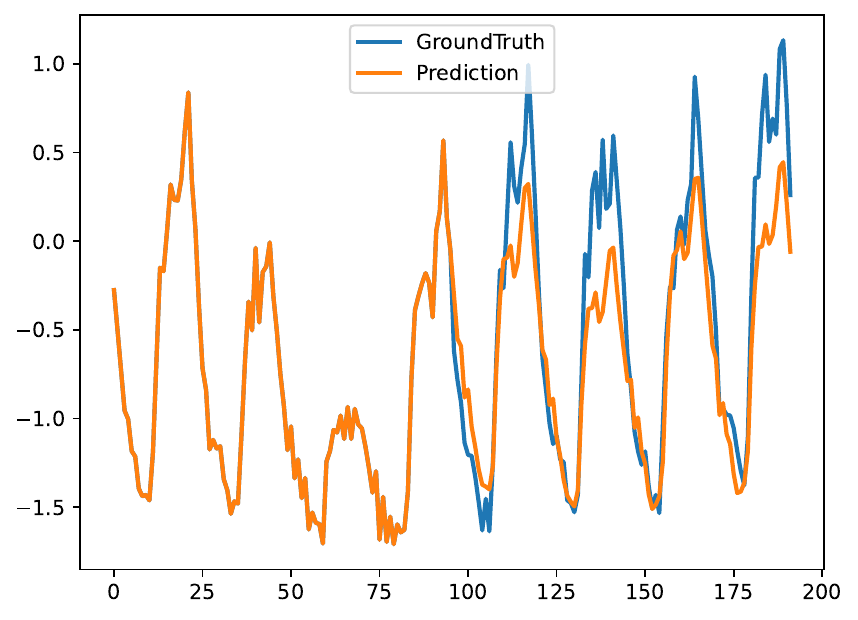}
\caption{Budget = All}
\end{subfigure}

\caption{The forecasts of TimePFN with various data budgets on ECL dataset. }

\label{fig:ECL0}
\end{figure*}

\begin{figure*}[htp]
\centering

\begin{subfigure}{0.32\textwidth}
\includegraphics[width=\linewidth]{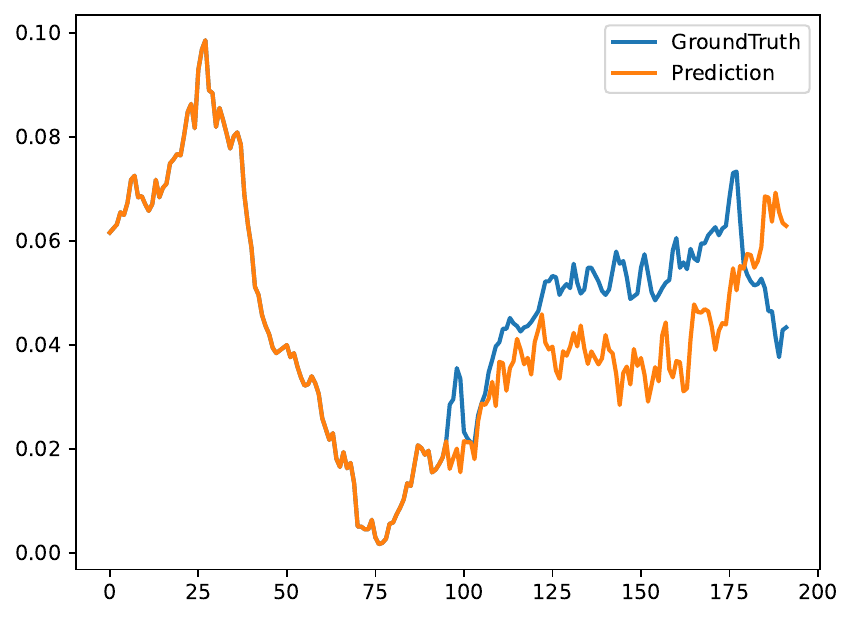}
\caption{Zero Shot}
\end{subfigure}\hfill
\begin{subfigure}{0.32\textwidth}
\includegraphics[width=\linewidth]{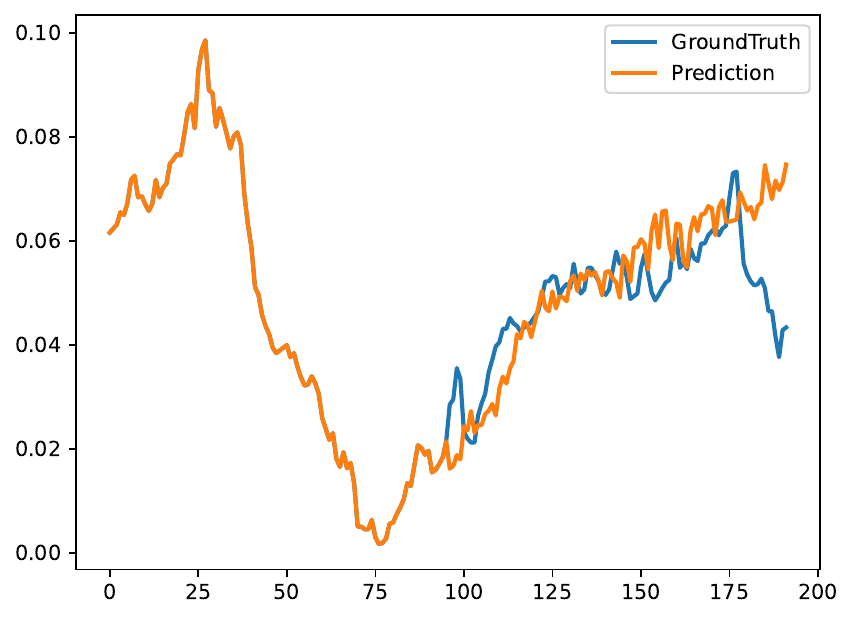}
\caption{Budget = 50}
\end{subfigure}\hfill
\begin{subfigure}{0.32\textwidth}
\includegraphics[width=\linewidth]{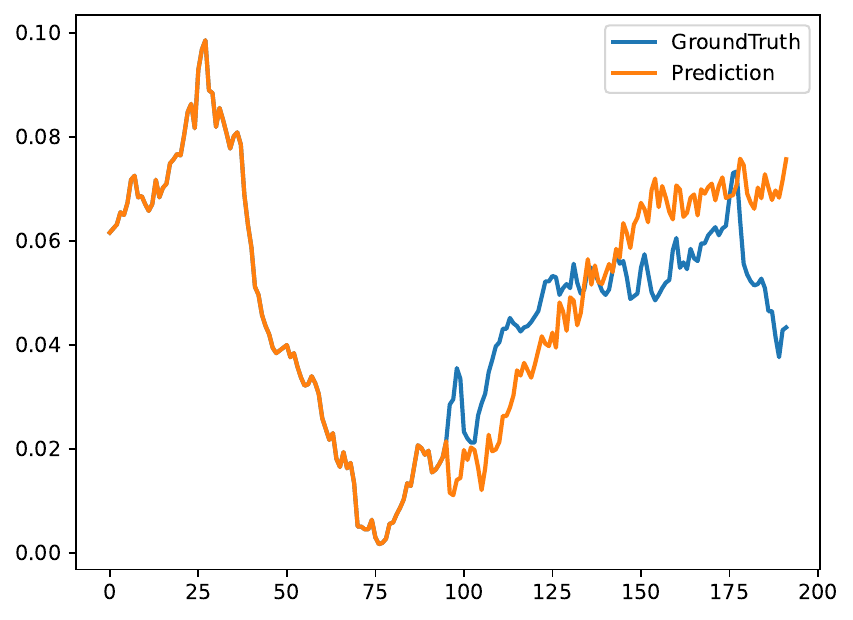}
\caption{Budget = 100}
\end{subfigure}

\begin{subfigure}{0.32\textwidth}
\includegraphics[width=\linewidth]{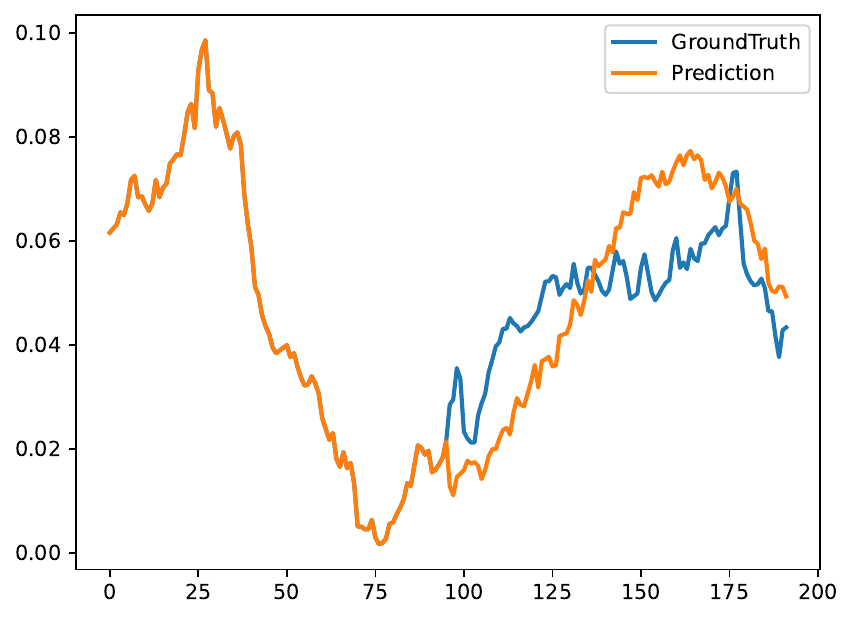}
\caption{Few-shot with budget 500}
\end{subfigure}\hfill
\begin{subfigure}{0.32\textwidth}
\includegraphics[width=\linewidth]{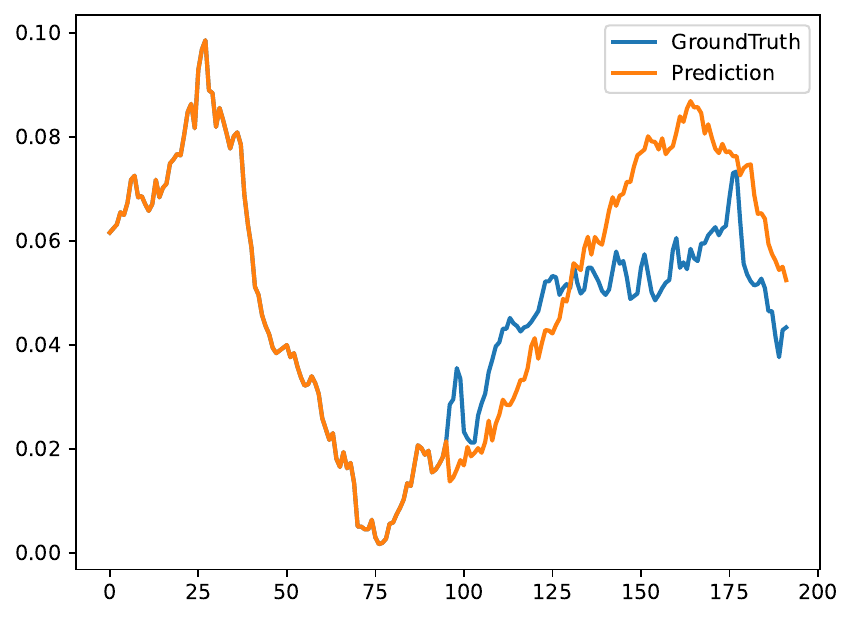}
\caption{Budget = 1000}
\end{subfigure}\hfill
\begin{subfigure}{0.32\textwidth}
\includegraphics[width=\linewidth]{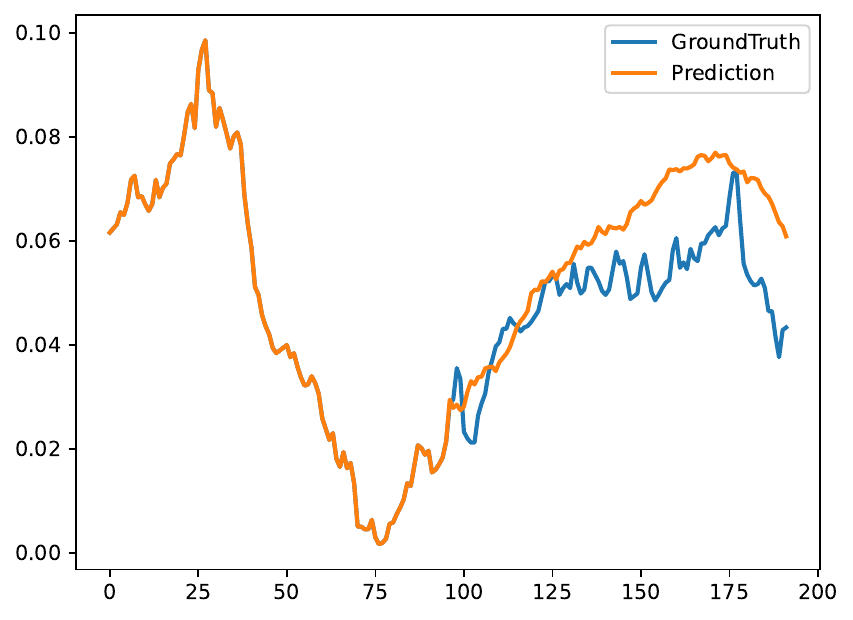}
\caption{Budget = All}
\end{subfigure}

\caption{The forecasts of TimePFN with various data budgets on weather dataset. }
\label{fig:weather280}
\end{figure*}
\begin{figure*}[htp]
\centering

\begin{subfigure}{0.32\textwidth}
\includegraphics[width=\linewidth]{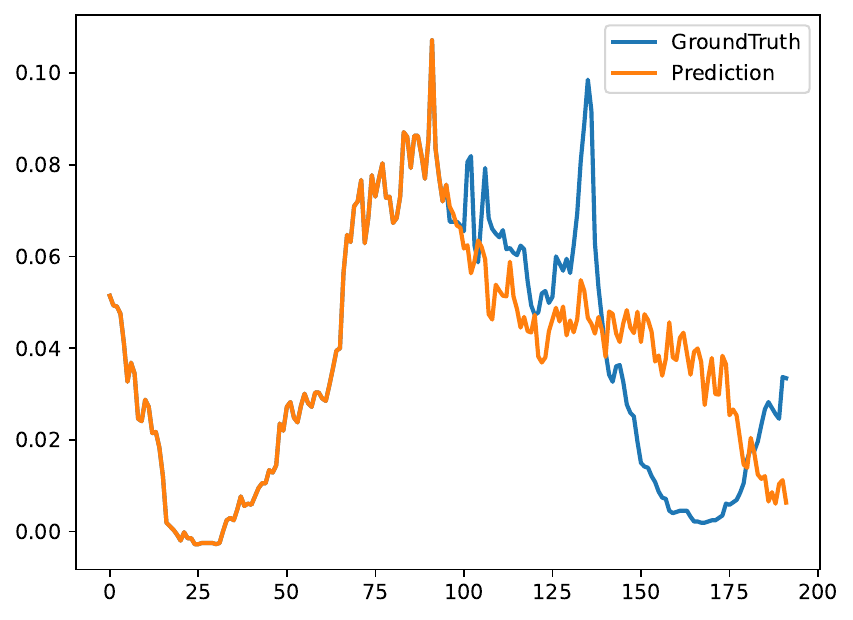}
\caption{Zero Shot}
\end{subfigure}\hfill
\begin{subfigure}{0.32\textwidth}
\includegraphics[width=\linewidth]{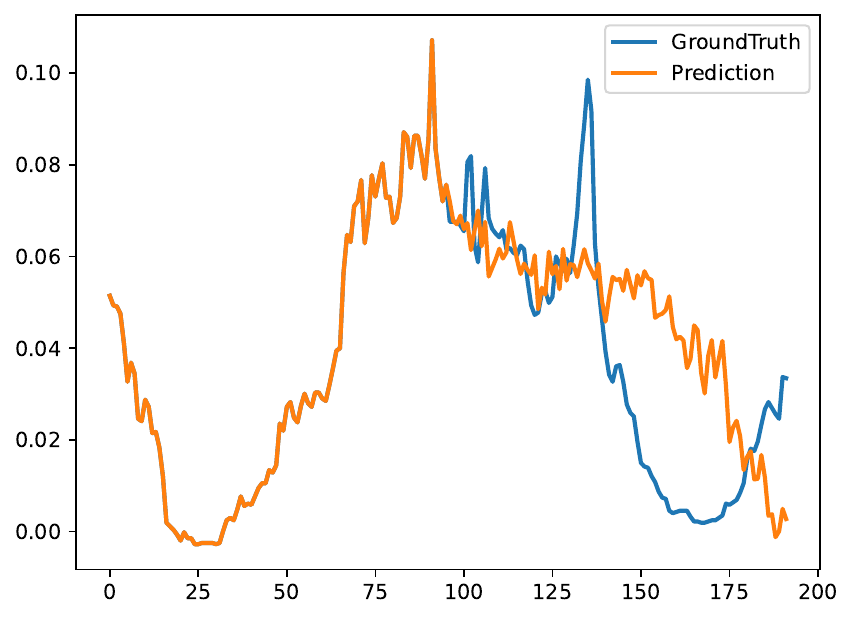}
\caption{Budget = 50}
\end{subfigure}\hfill
\begin{subfigure}{0.32\textwidth}
\includegraphics[width=\linewidth]{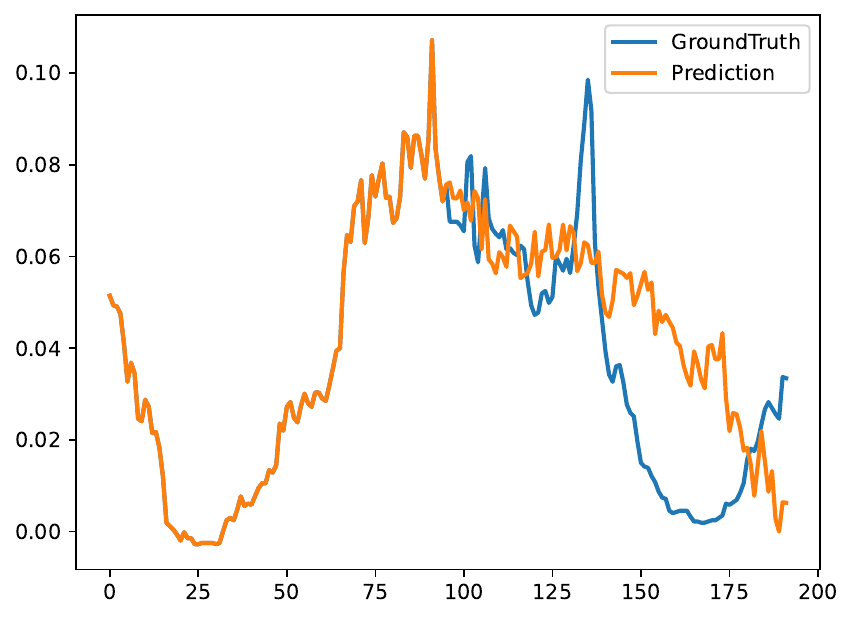}
\caption{Budget = 100}
\end{subfigure}

\begin{subfigure}{0.32\textwidth}
\includegraphics[width=\linewidth]{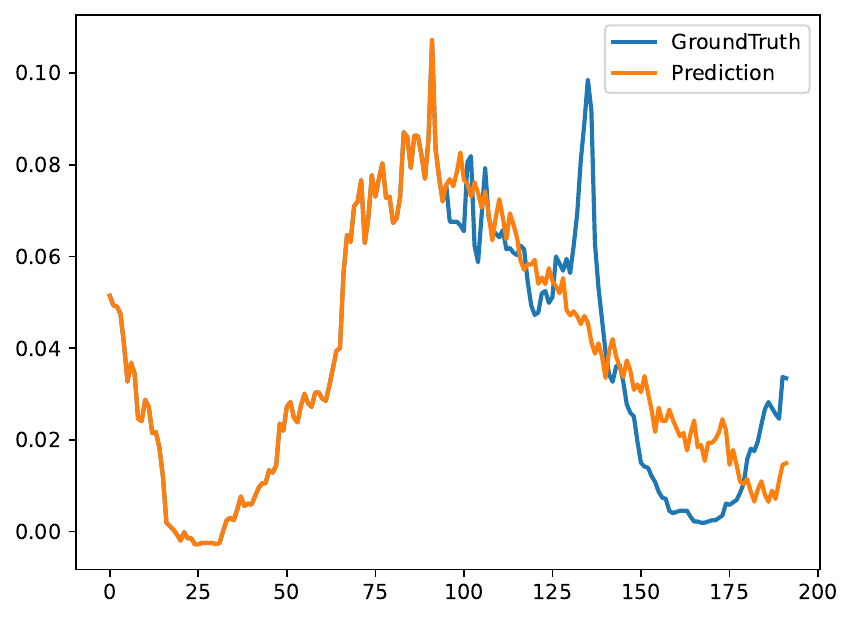}
\caption{Few-shot with budget 500}
\end{subfigure}\hfill
\begin{subfigure}{0.32\textwidth}
\includegraphics[width=\linewidth]{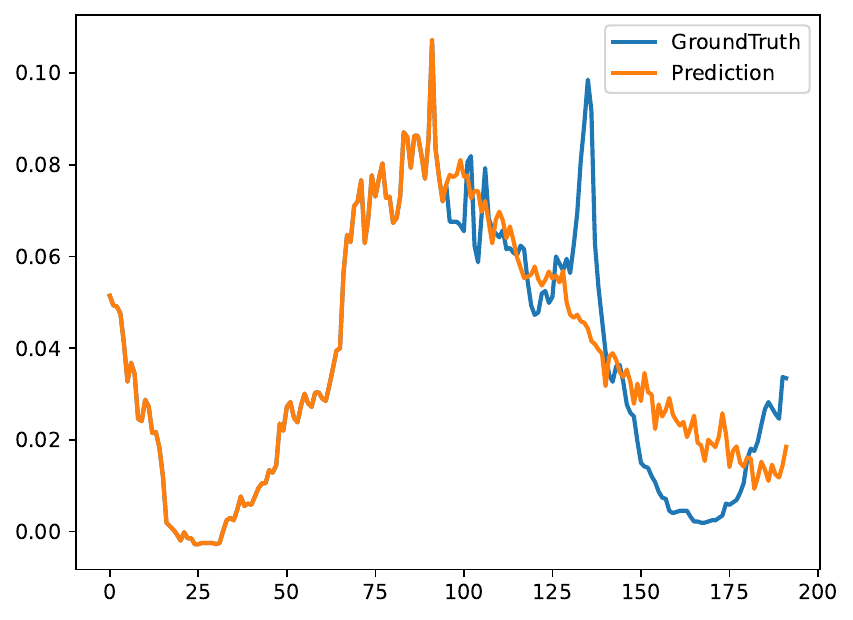}
\caption{Budget = 1000}
\end{subfigure}\hfill
\begin{subfigure}{0.32\textwidth}
\includegraphics[width=\linewidth]{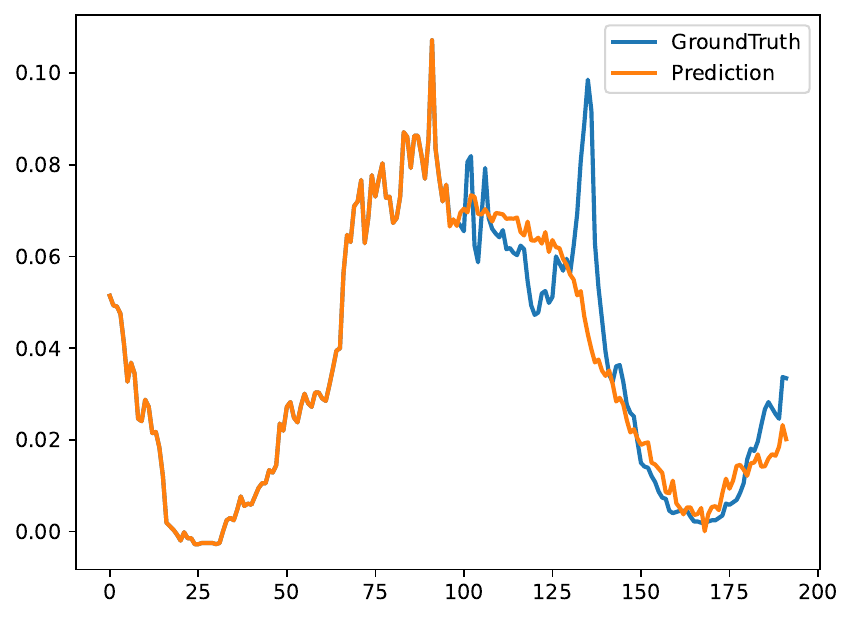}
\caption{Budget = All}
\end{subfigure}

\caption{The forecasts of TimePFN with various data budgets on weather dataset. }
\label{fig:weather40}
\end{figure*}

\begin{figure*}[htp]
\centering

\begin{subfigure}{0.32\textwidth}
\includegraphics[width=\linewidth]{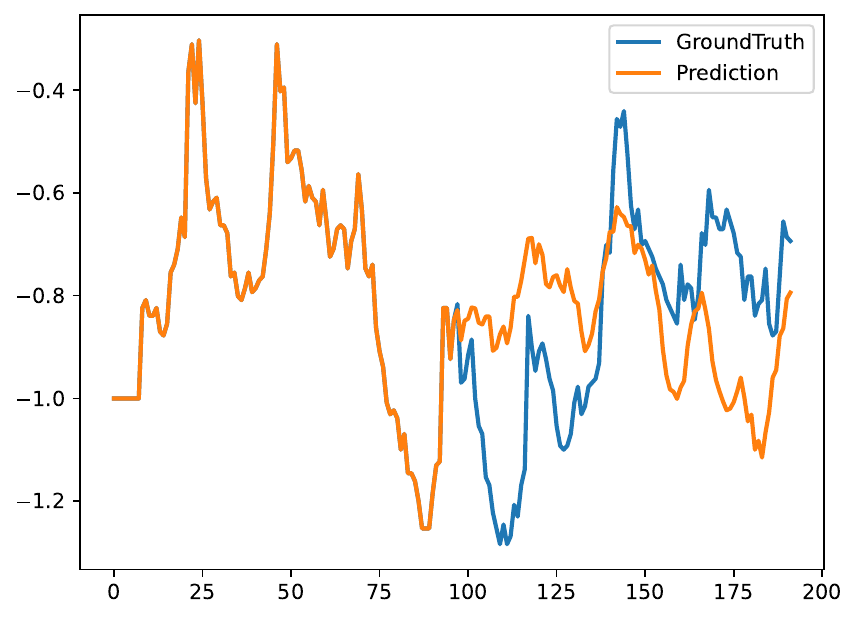}
\caption{Zero Shot}
\end{subfigure}\hfill
\begin{subfigure}{0.32\textwidth}
\includegraphics[width=\linewidth]{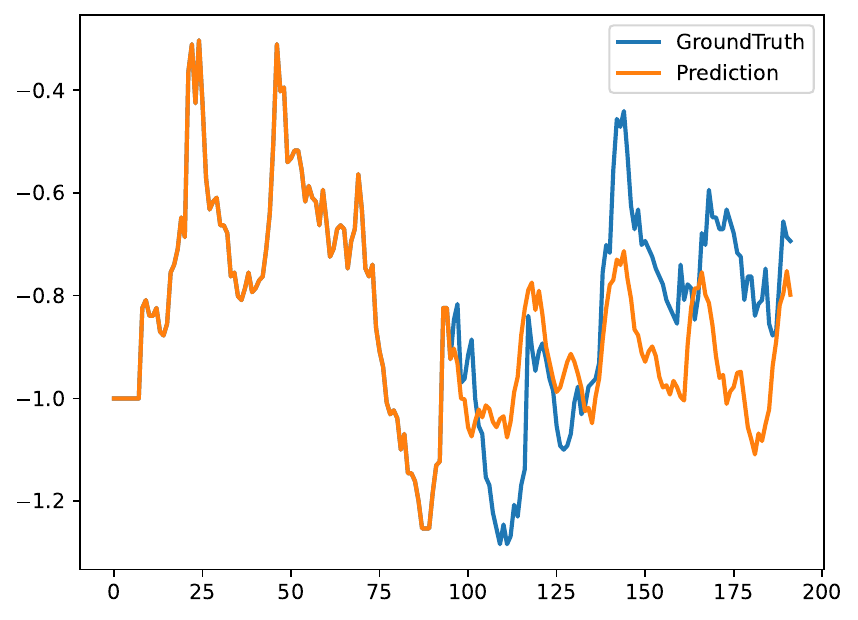}
\caption{Budget = 50}
\end{subfigure}\hfill
\begin{subfigure}{0.32\textwidth}
\includegraphics[width=\linewidth]{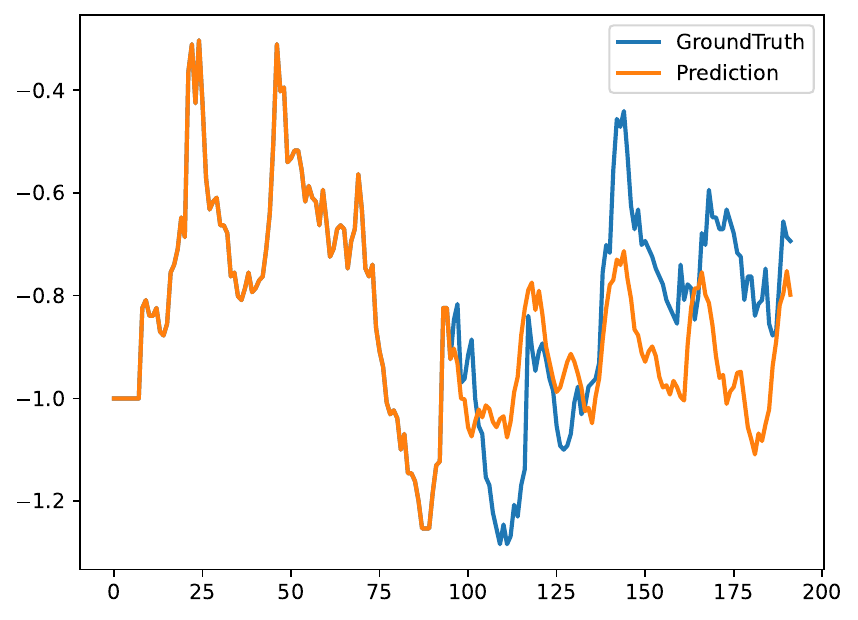}
\caption{Budget = 100}
\end{subfigure}

\begin{subfigure}{0.32\textwidth}
\includegraphics[width=\linewidth]{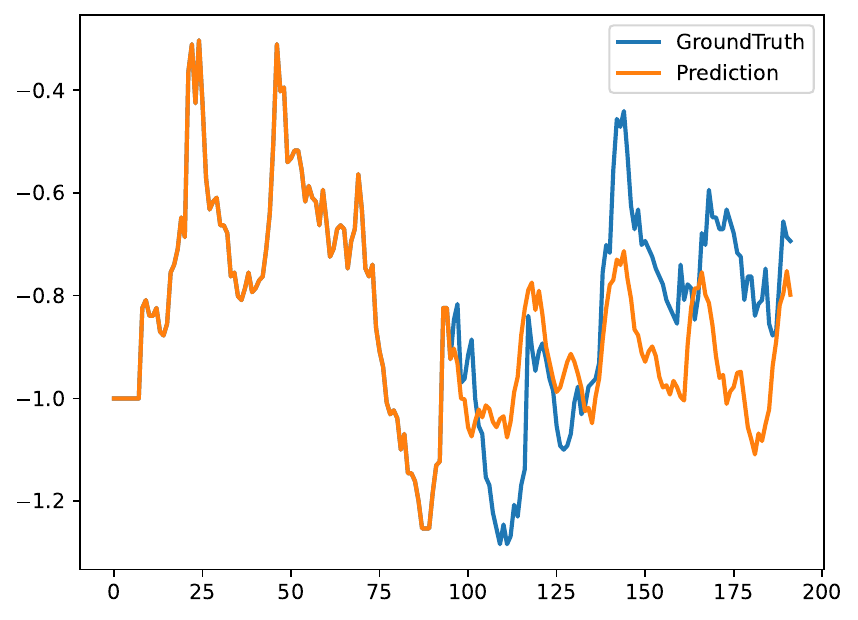}
\caption{Few-shot with budget 500}
\end{subfigure}\hfill
\begin{subfigure}{0.32\textwidth}
\includegraphics[width=\linewidth]{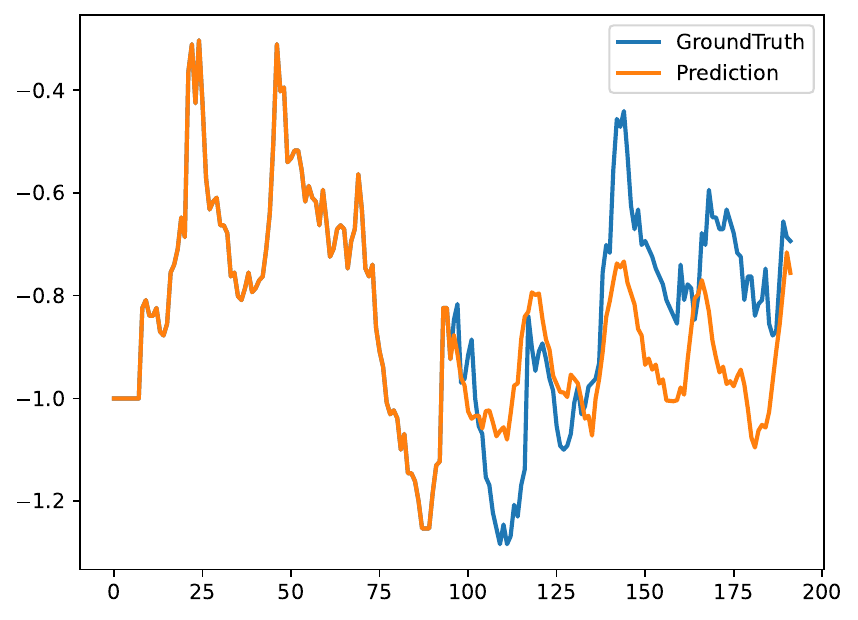}
\caption{Budget = 1000}
\end{subfigure}\hfill
\begin{subfigure}{0.32\textwidth}
\includegraphics[width=\linewidth]{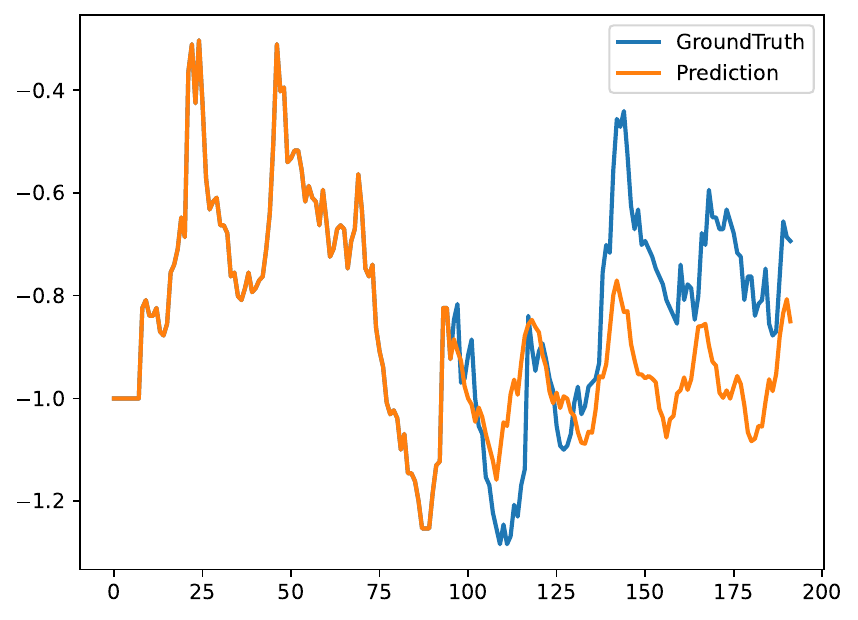}
\caption{Budget = All}
\end{subfigure}

\caption{The forecasts of TimePFN with various data budgets on ETTh1 dataset. }
\label{fig:etth1_280}
\end{figure*}
\begin{figure*}[htp]
\centering

\begin{subfigure}{0.32\textwidth}
\includegraphics[width=\linewidth]{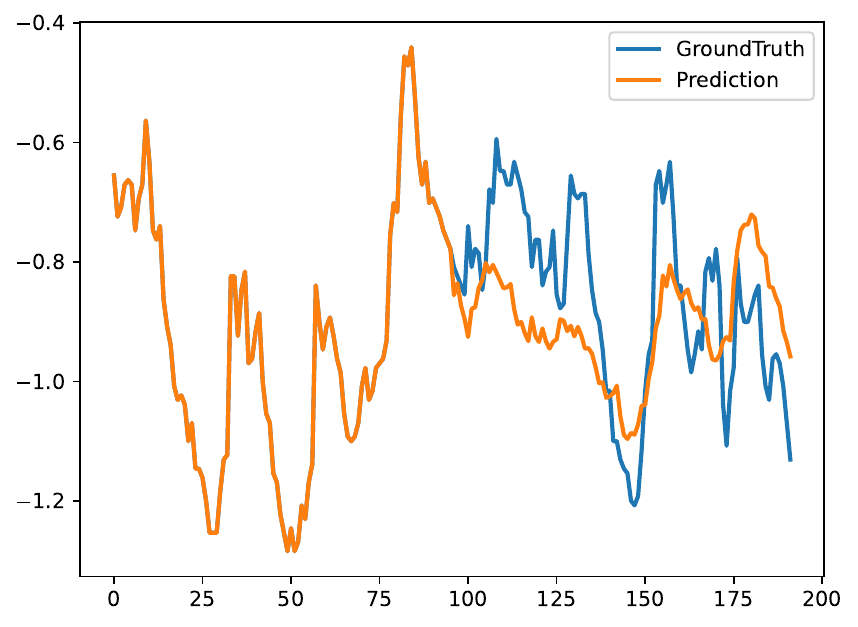}
\caption{Zero Shot}
\end{subfigure}\hfill
\begin{subfigure}{0.32\textwidth}
\includegraphics[width=\linewidth]{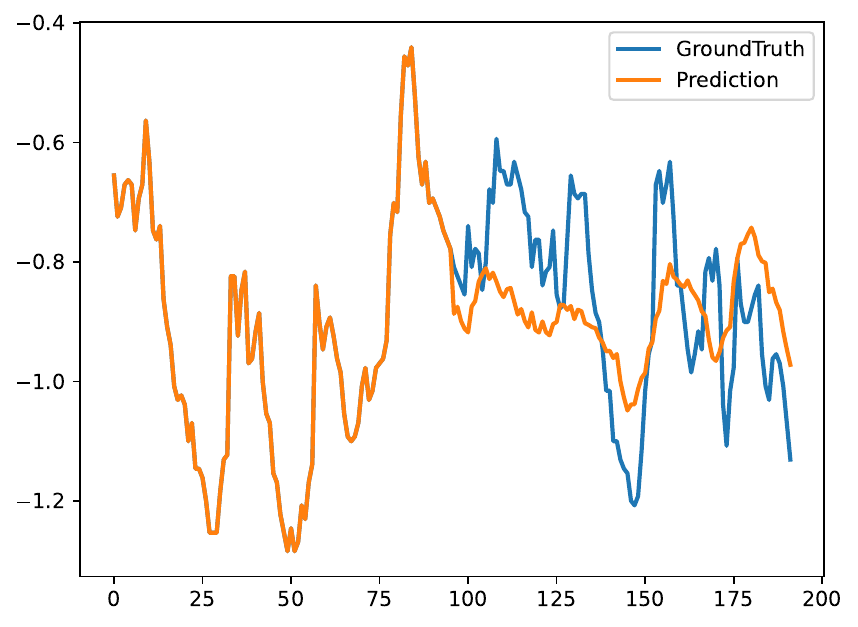}
\caption{Budget = 50}
\end{subfigure}\hfill
\begin{subfigure}{0.32\textwidth}
\includegraphics[width=\linewidth]{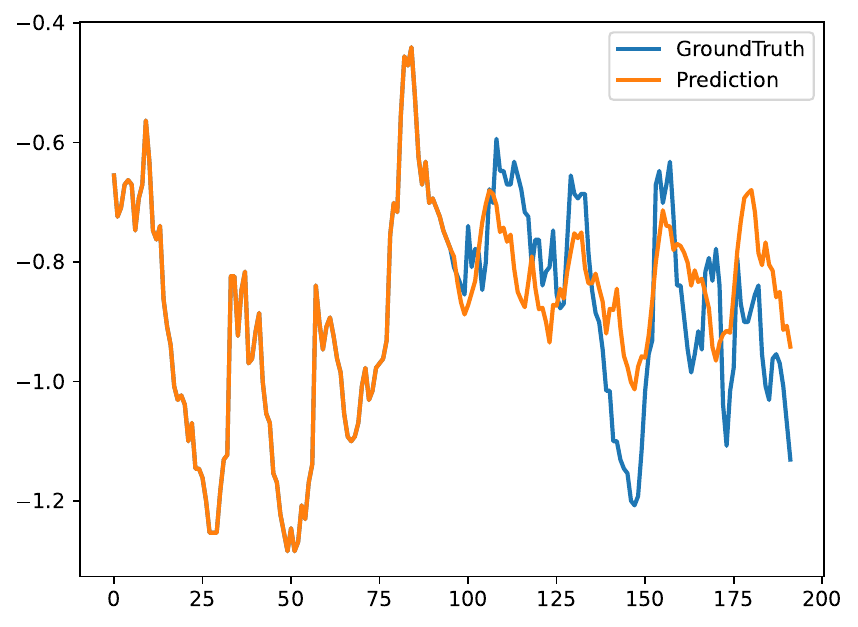}
\caption{Budget = 100}
\end{subfigure}

\begin{subfigure}{0.32\textwidth}
\includegraphics[width=\linewidth]{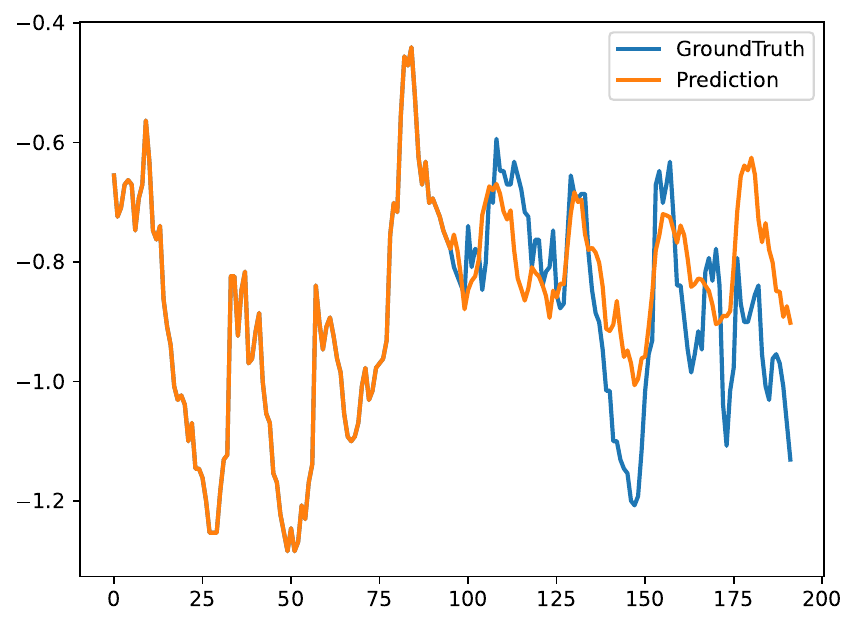}
\caption{Few-shot with budget 500}
\end{subfigure}\hfill
\begin{subfigure}{0.32\textwidth}
\includegraphics[width=\linewidth]{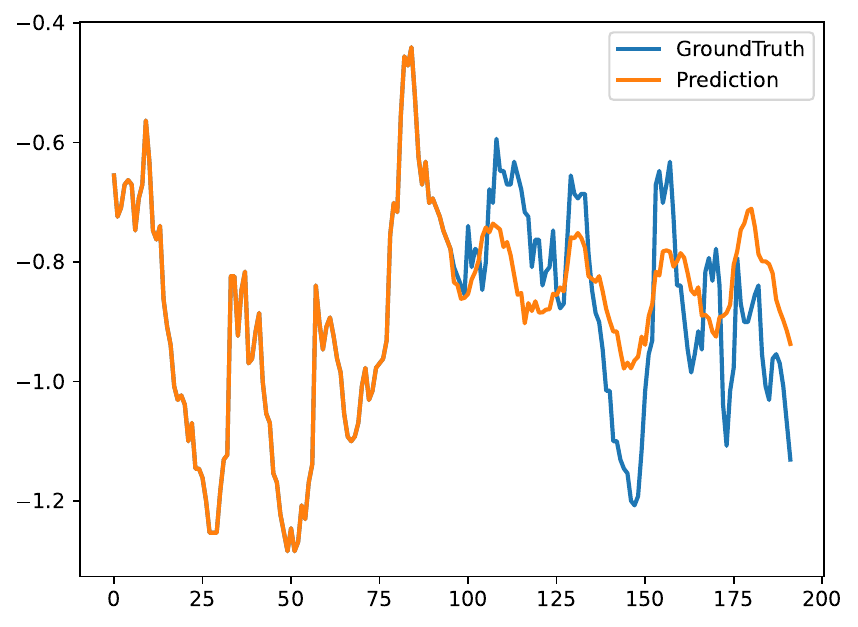}
\caption{Budget = 1000}
\end{subfigure}\hfill
\begin{subfigure}{0.32\textwidth}
\includegraphics[width=\linewidth]{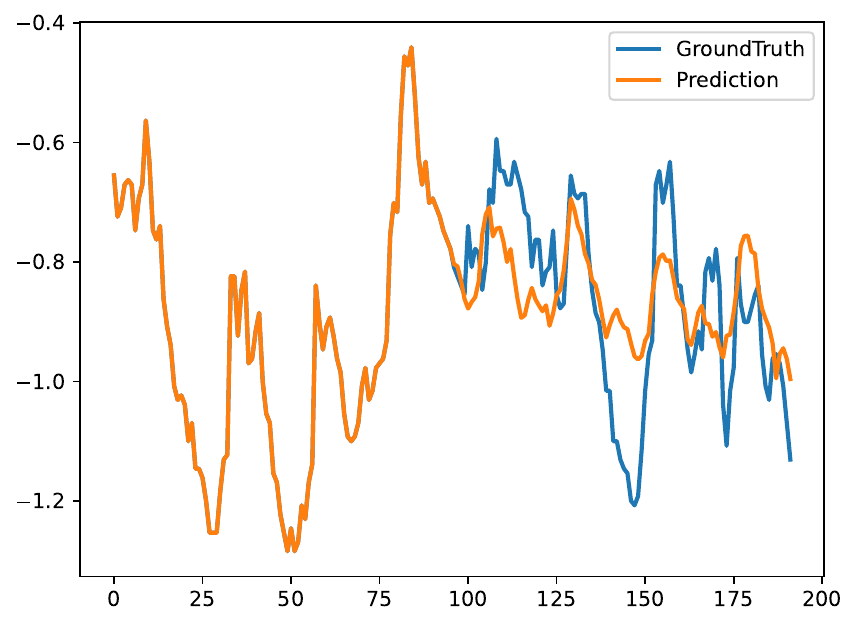}
\caption{Budget = All}
\end{subfigure}

\caption{The forecasts of TimePFN with various data budgets on ETTh1 dataset. }
\label{fig:etth1_340}
\end{figure*}

\begin{figure*}[htp]
\centering

\begin{subfigure}{0.32\textwidth}
\includegraphics[width=\linewidth]{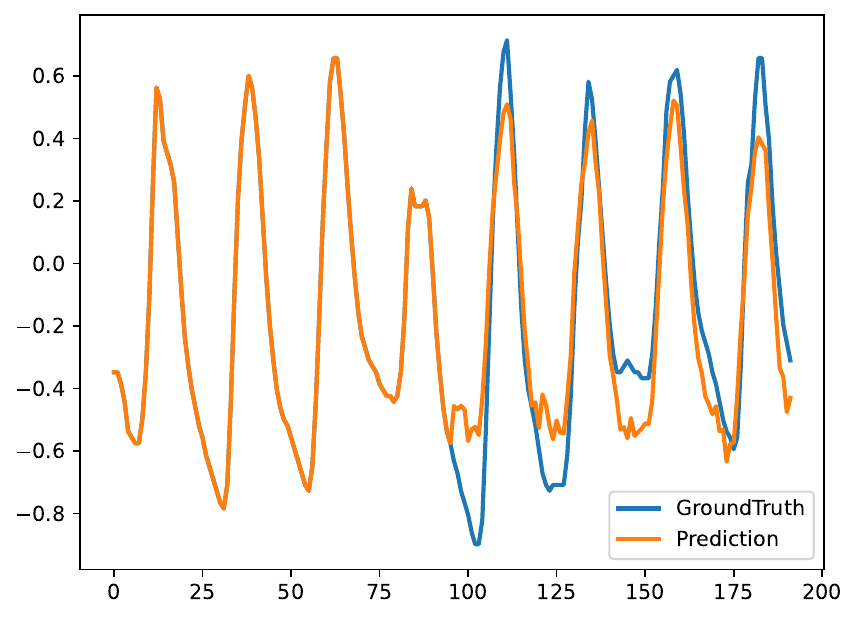}
\caption{Zero Shot}
\end{subfigure}\hfill
\begin{subfigure}{0.32\textwidth}
\includegraphics[width=\linewidth]{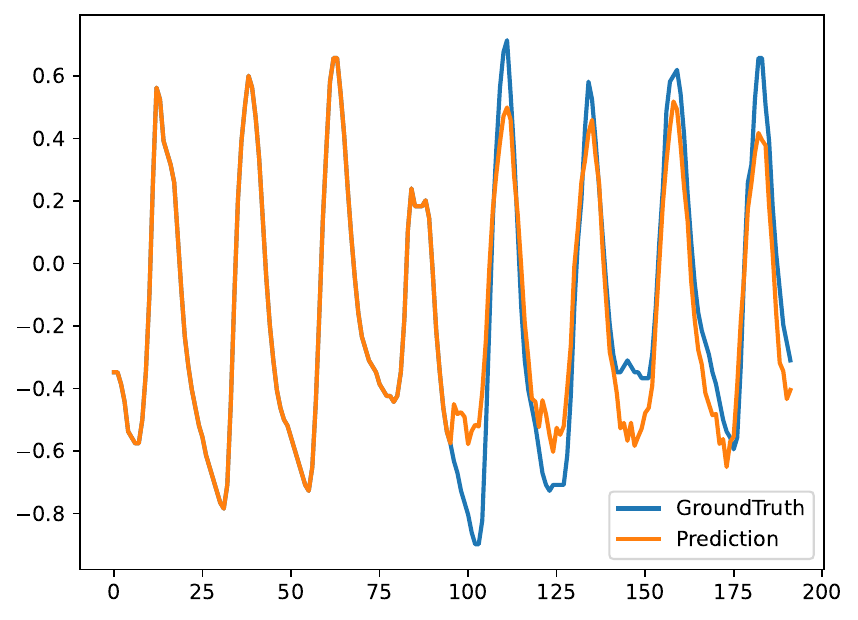}
\caption{Budget = 50}
\end{subfigure}\hfill
\begin{subfigure}{0.32\textwidth}
\includegraphics[width=\linewidth]{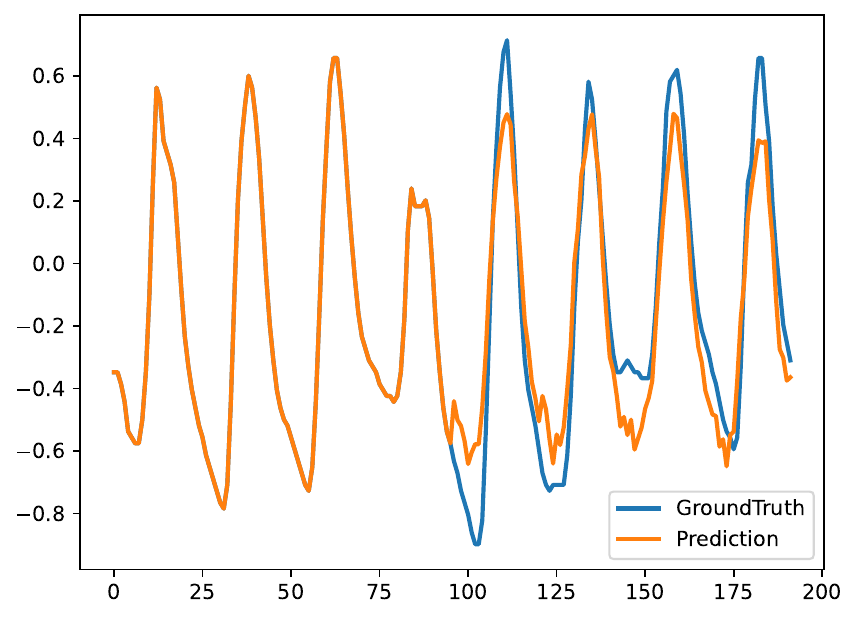}
\caption{Budget = 100}
\end{subfigure}

\begin{subfigure}{0.32\textwidth}
\includegraphics[width=\linewidth]{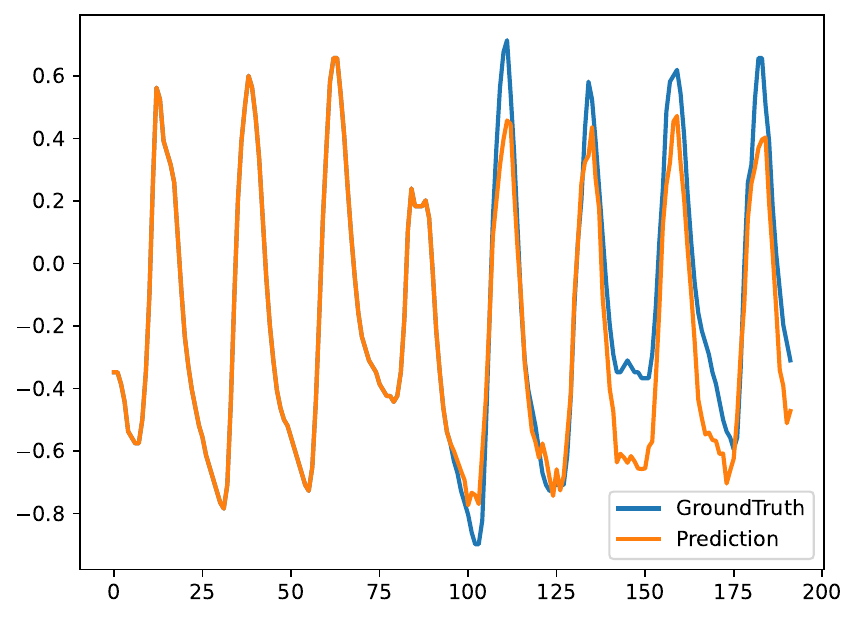}
\caption{Budget = 500}
\end{subfigure}\hfill
\begin{subfigure}{0.32\textwidth}
\includegraphics[width=\linewidth]{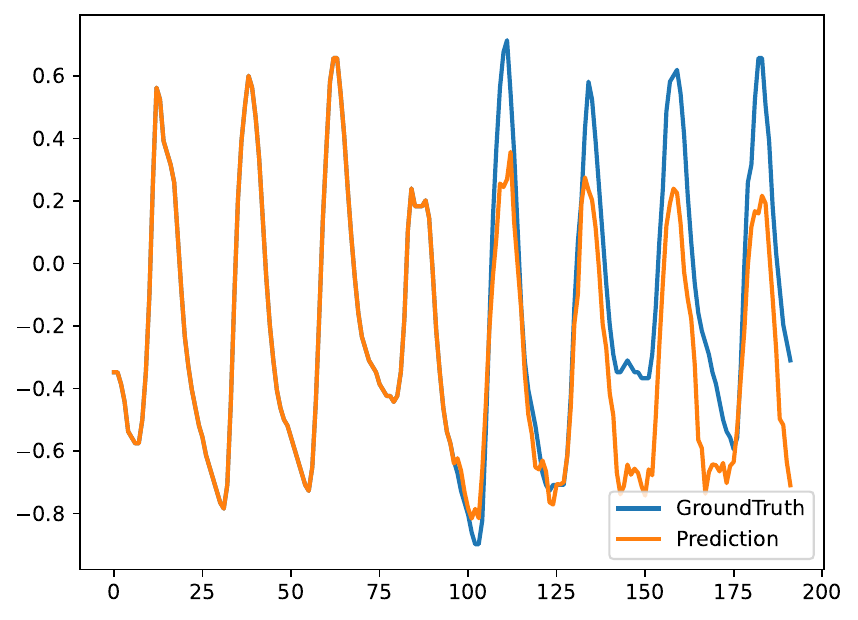}
\caption{Budget = 1000}
\end{subfigure}\hfill
\begin{subfigure}{0.32\textwidth}
\includegraphics[width=\linewidth]{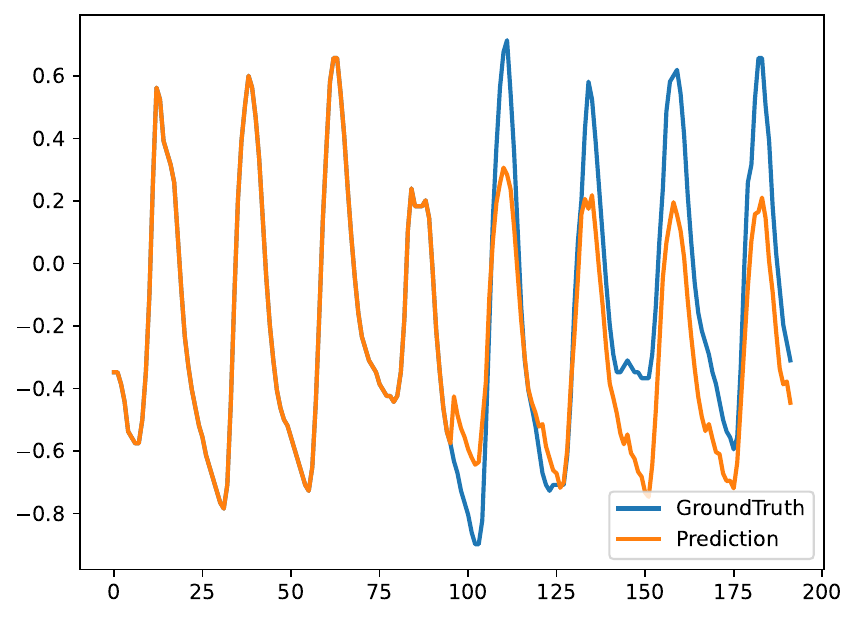}
\caption{Budget = All}
\end{subfigure}

\caption{The forecasts of TimePFN with various data budgets on ETTh2 dataset. }

\label{fig:etth2_0}
\end{figure*}
\begin{figure*}[htp]
\centering

\begin{subfigure}{0.32\textwidth}
\includegraphics[width=\linewidth]{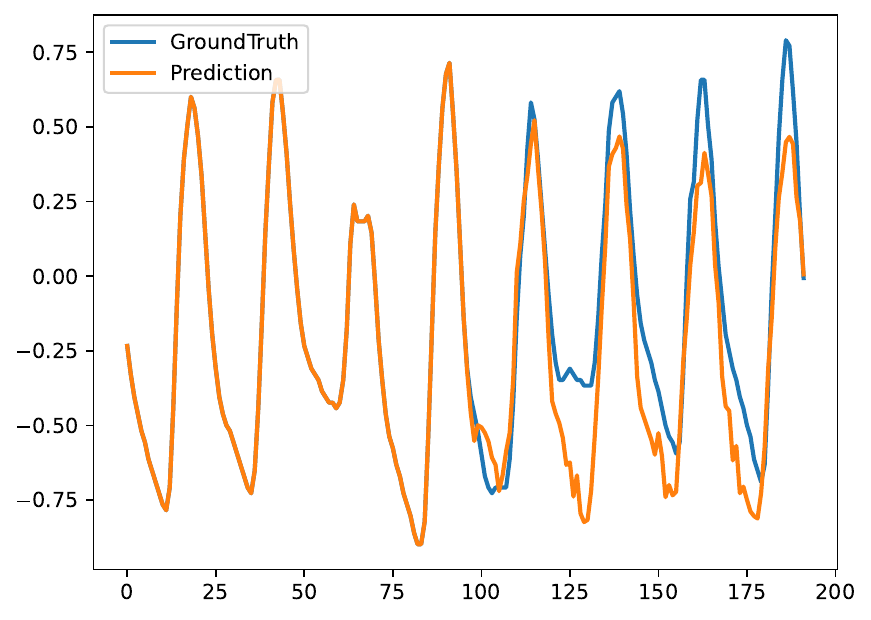}
\caption{Zero Shot}
\end{subfigure}\hfill
\begin{subfigure}{0.32\textwidth}
\includegraphics[width=\linewidth]{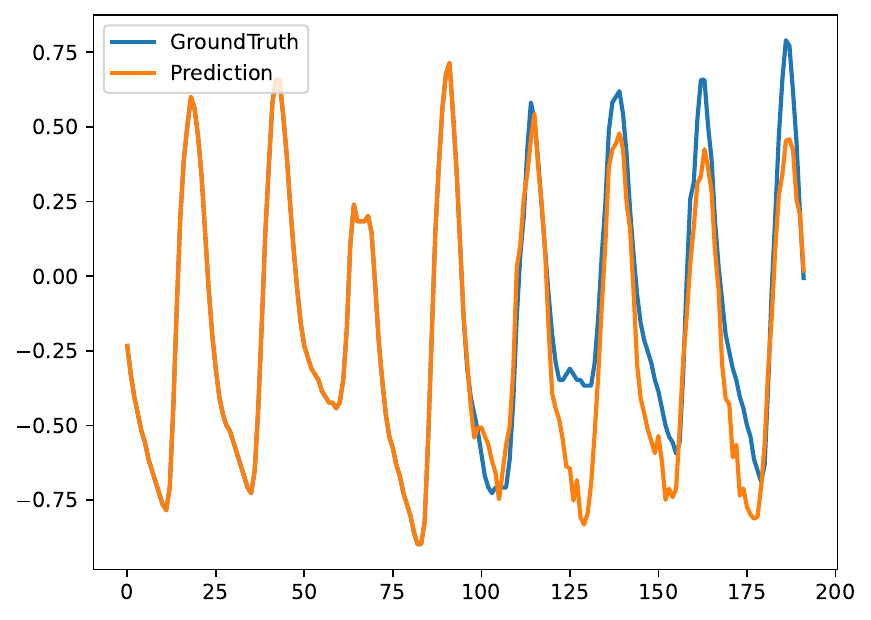}
\caption{Budget = 50}
\end{subfigure}\hfill
\begin{subfigure}{0.32\textwidth}
\includegraphics[width=\linewidth]{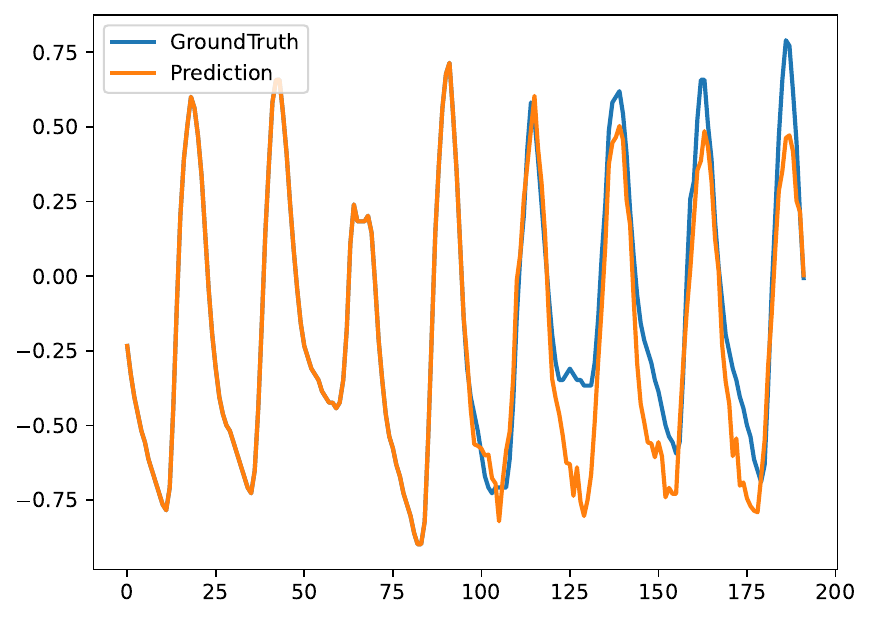}
\caption{Budget = 100}
\end{subfigure}

\begin{subfigure}{0.32\textwidth}
\includegraphics[width=\linewidth]{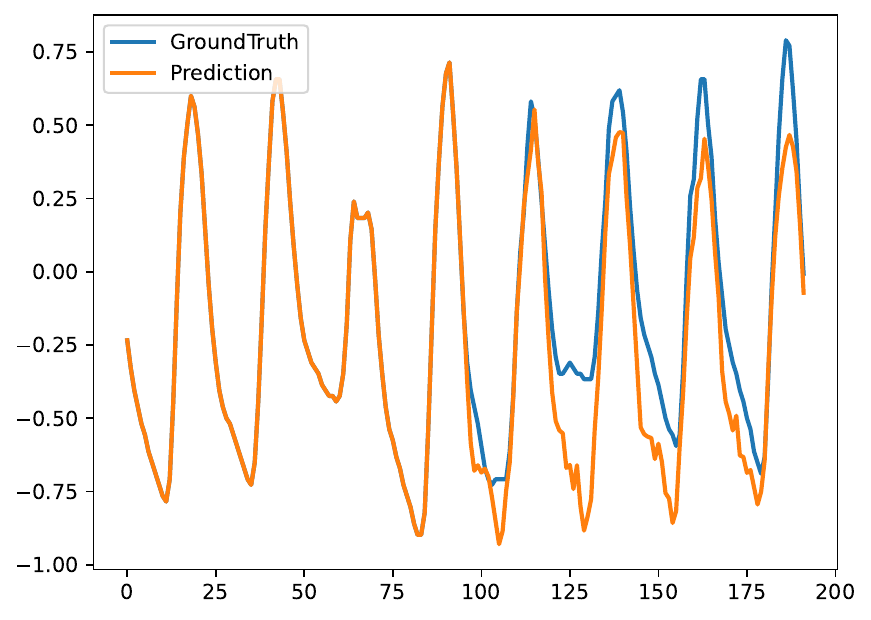}
\caption{Budget = 500}
\end{subfigure}\hfill
\begin{subfigure}{0.32\textwidth}
\includegraphics[width=\linewidth]{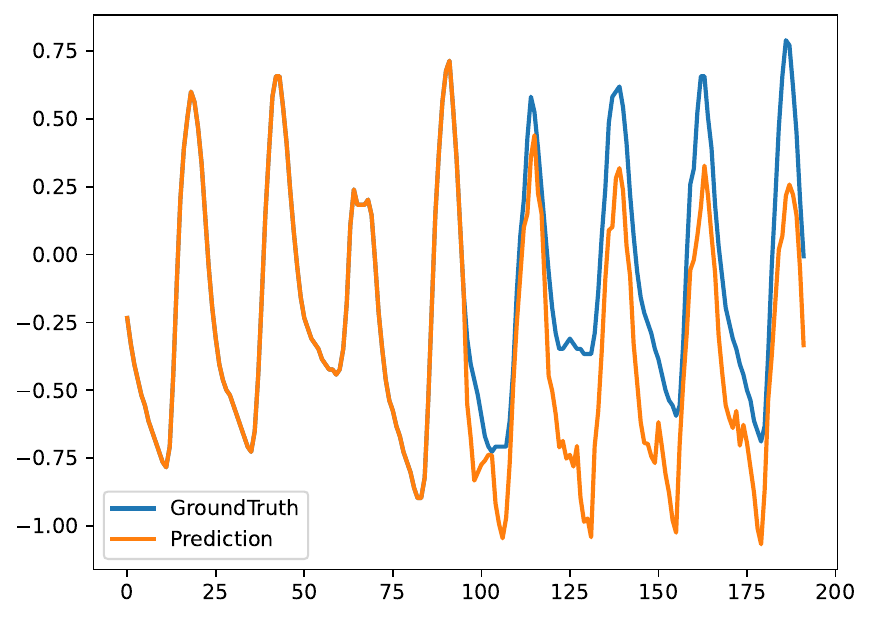}
\caption{Budget = 1000}
\end{subfigure}\hfill
\begin{subfigure}{0.32\textwidth}
\includegraphics[width=\linewidth]{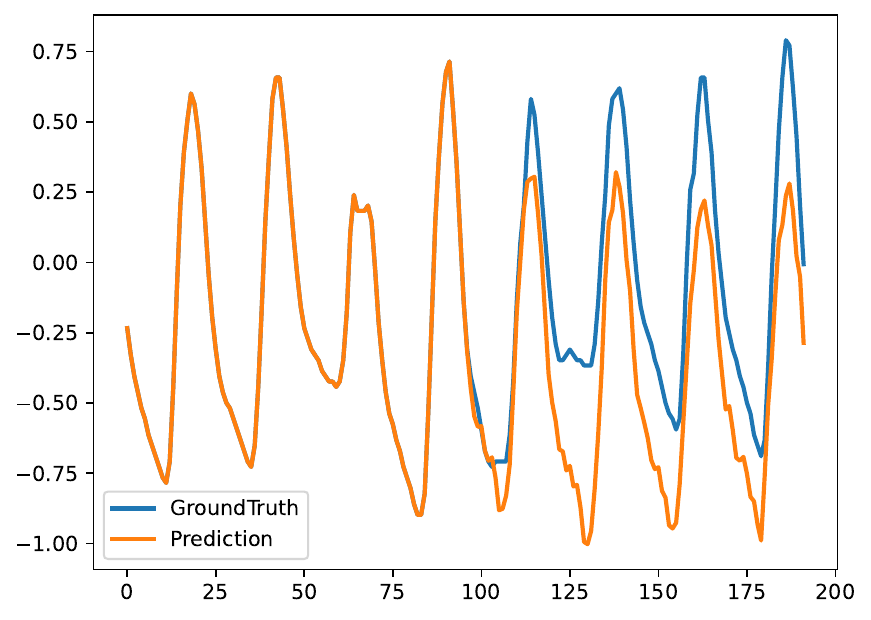}
\caption{Budget = All}
\end{subfigure}

\caption{The forecasts of TimePFN with various data budgets on ETTh2 dataset. }

\label{fig:etth2_20}
\end{figure*}

\begin{figure*}[htp]
\centering

\begin{subfigure}{0.32\textwidth}
\includegraphics[width=\linewidth]{figures/ECL/zero_shot/0.pdf}
\caption{Zero Shot}
\end{subfigure}\hfill
\begin{subfigure}{0.32\textwidth}
\includegraphics[width=\linewidth]{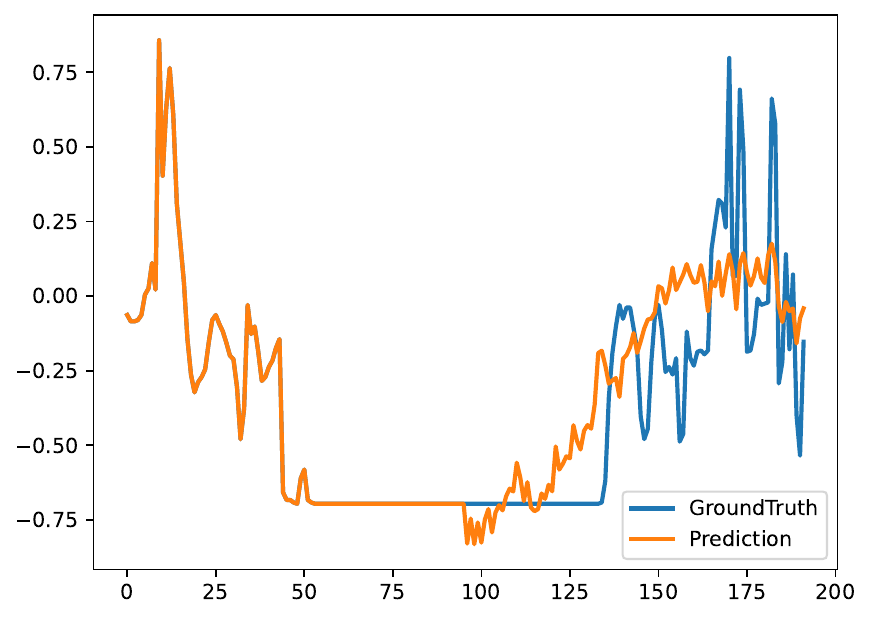}
\caption{Budget = 50}
\end{subfigure}\hfill
\begin{subfigure}{0.32\textwidth}
\includegraphics[width=\linewidth]{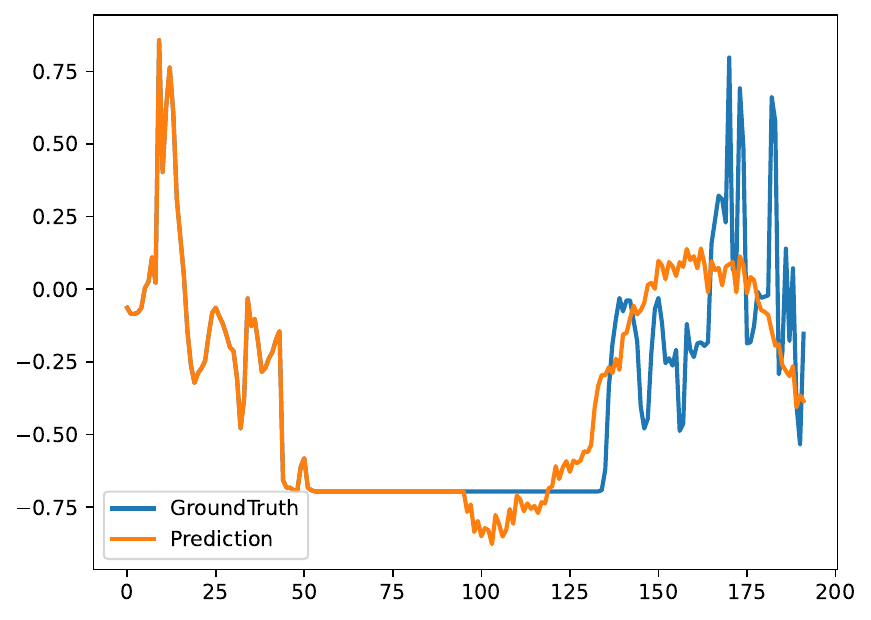}
\caption{Budget = 100}
\end{subfigure}

\begin{subfigure}{0.32\textwidth}
\includegraphics[width=\linewidth]{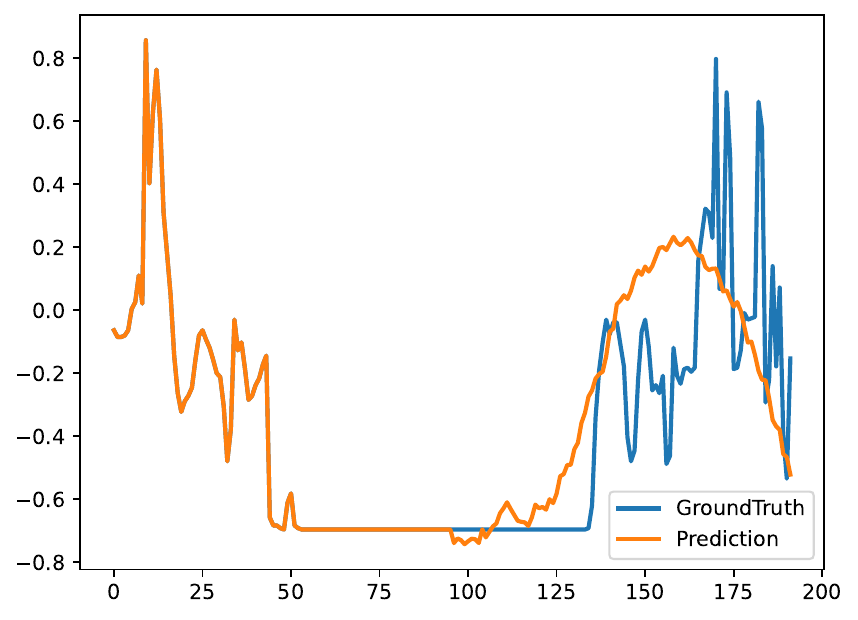}
\caption{Budget = 500}
\end{subfigure}\hfill
\begin{subfigure}{0.32\textwidth}
\includegraphics[width=\linewidth]{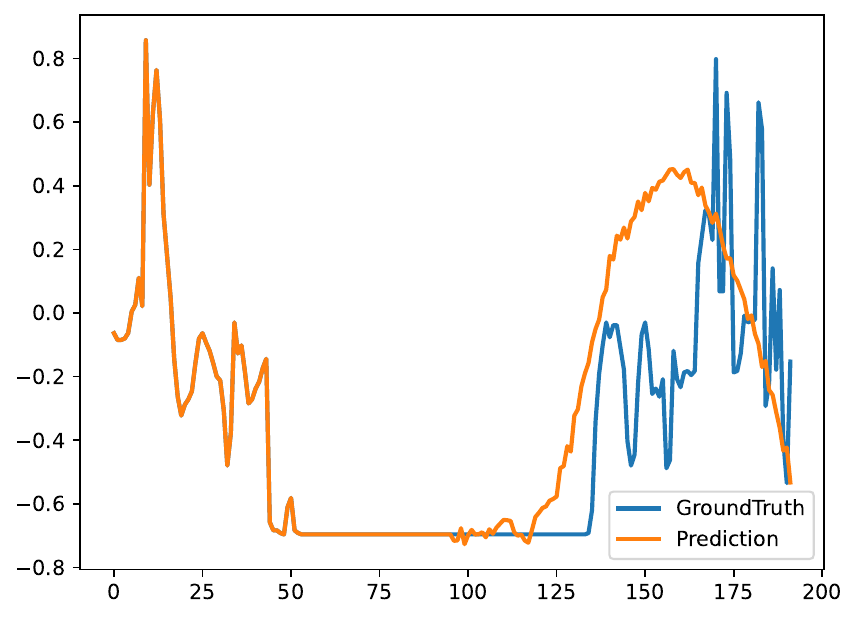}
\caption{Budget = 1000}
\end{subfigure}\hfill
\begin{subfigure}{0.32\textwidth}
\includegraphics[width=\linewidth]{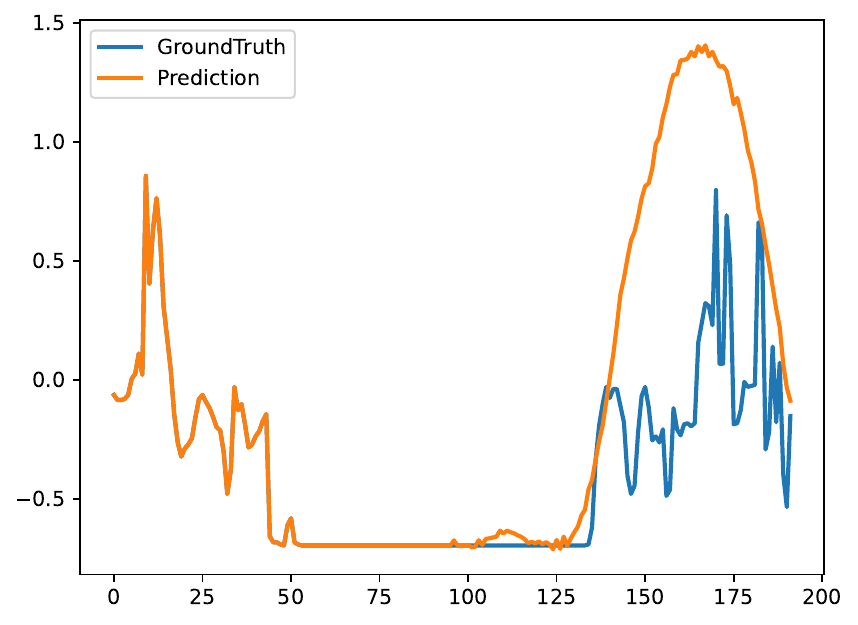}
\caption{Budget = All}
\end{subfigure}

\caption{The forecasts of TimePFN with various data budgets on Solar dataset. }

\label{fig:solar0}
\end{figure*}
\begin{figure*}[htp]
\centering

\begin{subfigure}{0.32\textwidth}
\includegraphics[width=\linewidth]{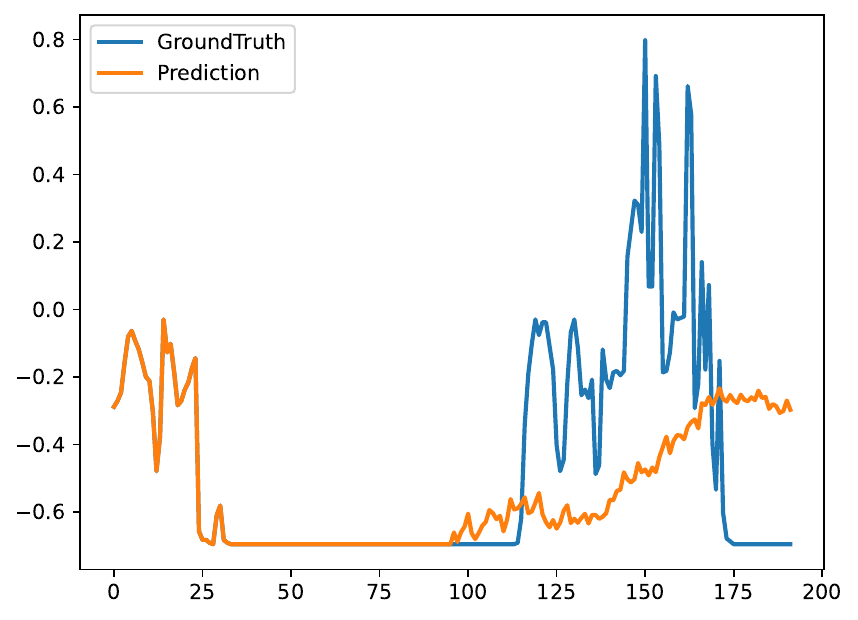}
\caption{Zero Shot}
\end{subfigure}\hfill
\begin{subfigure}{0.32\textwidth}
\includegraphics[width=\linewidth]{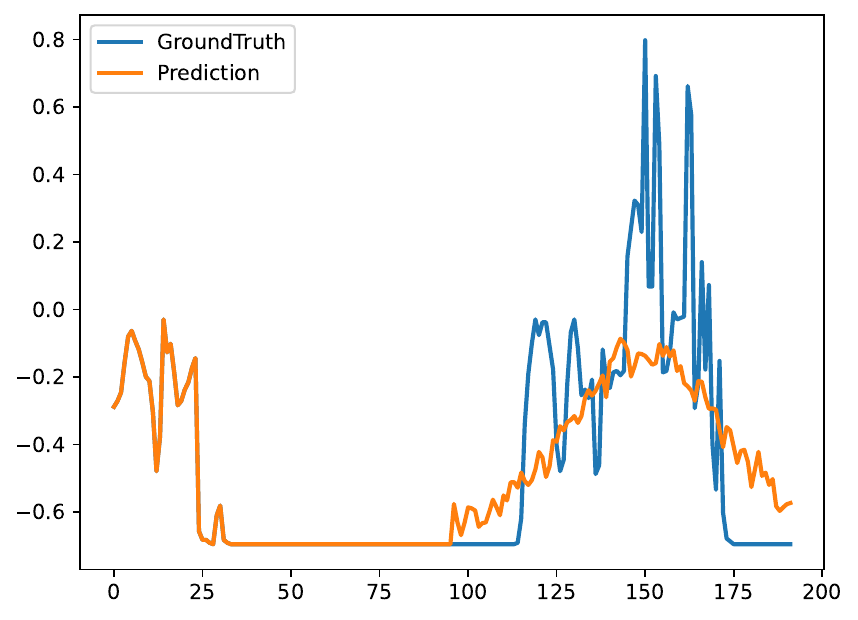}
\caption{Budget = 50}
\end{subfigure}\hfill
\begin{subfigure}{0.32\textwidth}
\includegraphics[width=\linewidth]{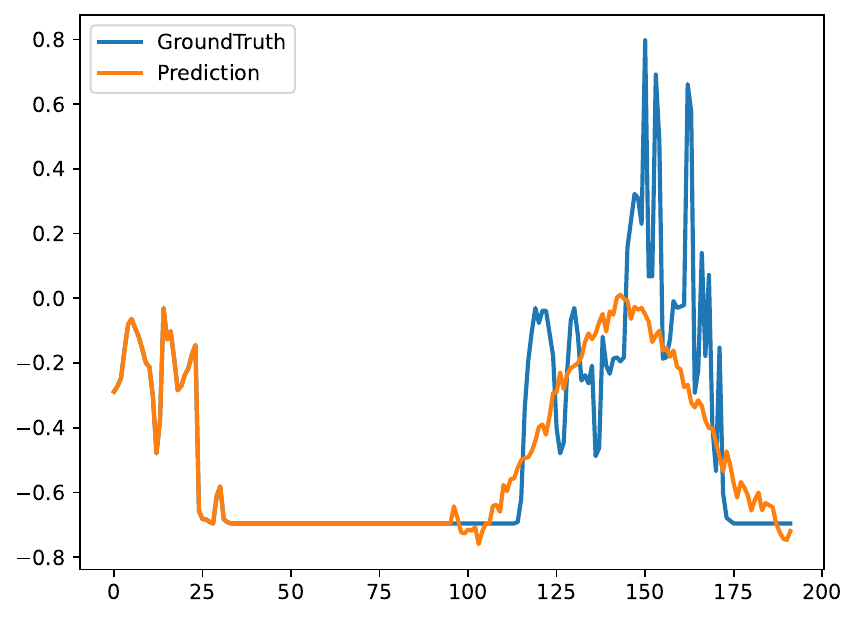}
\caption{Budget = 100}
\end{subfigure}

\begin{subfigure}{0.32\textwidth}
\includegraphics[width=\linewidth]{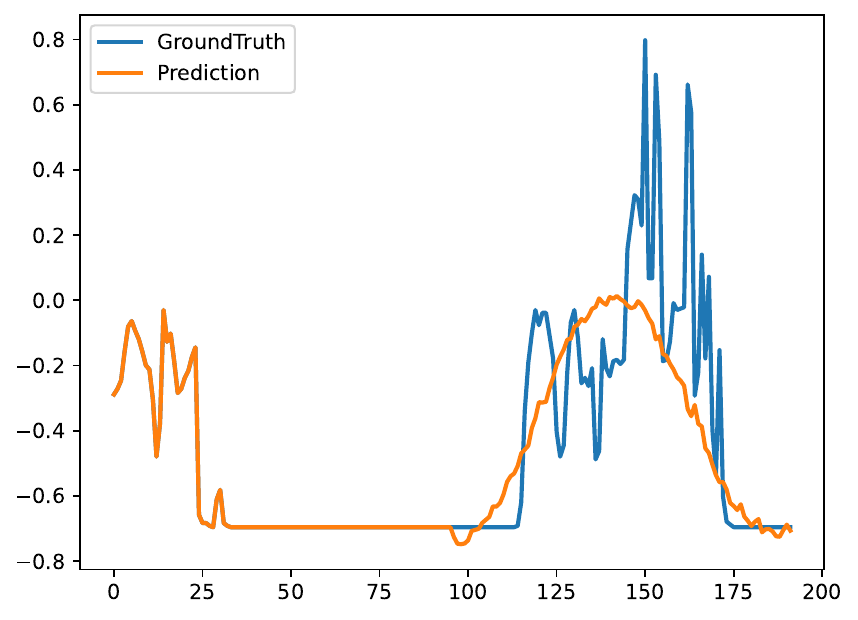}
\caption{Few-shot with budget 500}
\end{subfigure}\hfill
\begin{subfigure}{0.32\textwidth}
\includegraphics[width=\linewidth]{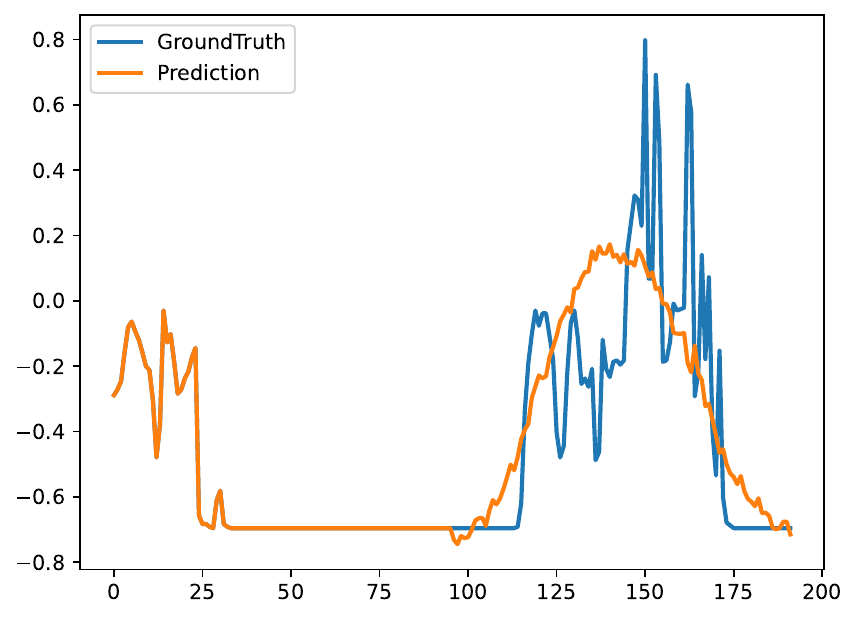}
\caption{Budget = 1000}
\end{subfigure}\hfill
\begin{subfigure}{0.32\textwidth}
\includegraphics[width=\linewidth]{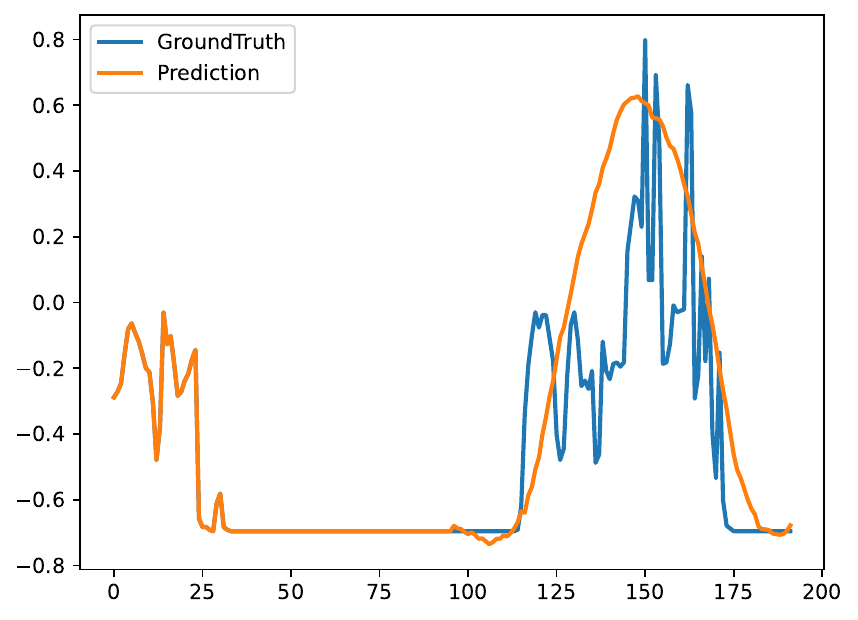}
\caption{Budget = All}
\end{subfigure}

\caption{The forecasts of TimePFN with various data budgets on Solar dataset. }
\label{fig:solar20}
\end{figure*}

\begin{figure*}[htp]
\centering

\begin{subfigure}{0.32\textwidth}
\includegraphics[width=\linewidth]{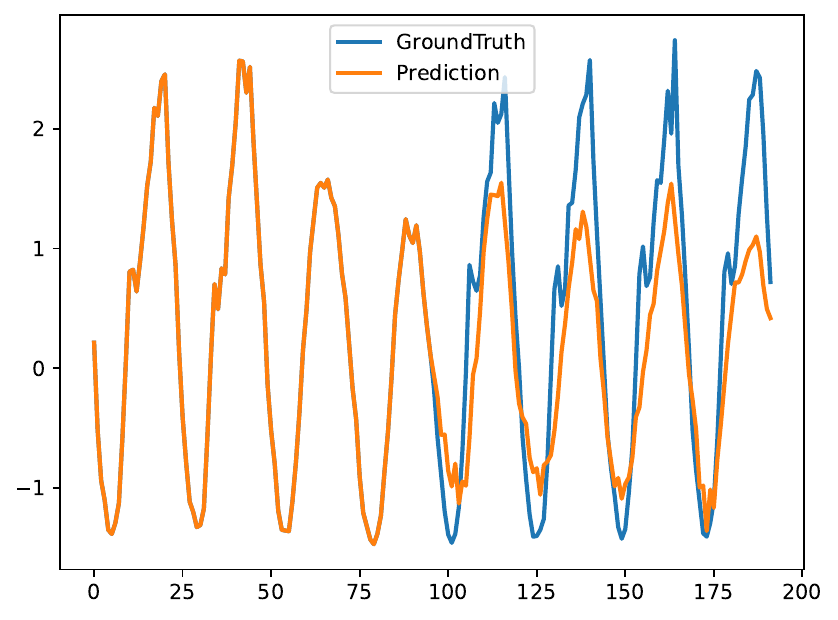}
\caption{Zero Shot}
\end{subfigure}\hfill
\begin{subfigure}{0.32\textwidth}
\includegraphics[width=\linewidth]{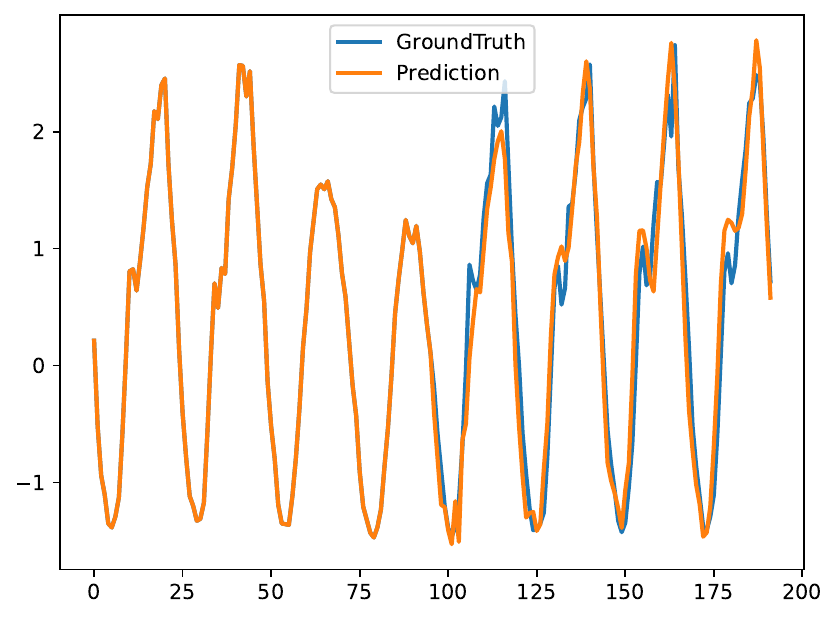}
\caption{Budget = 50}
\end{subfigure}\hfill
\begin{subfigure}{0.32\textwidth}
\includegraphics[width=\linewidth]{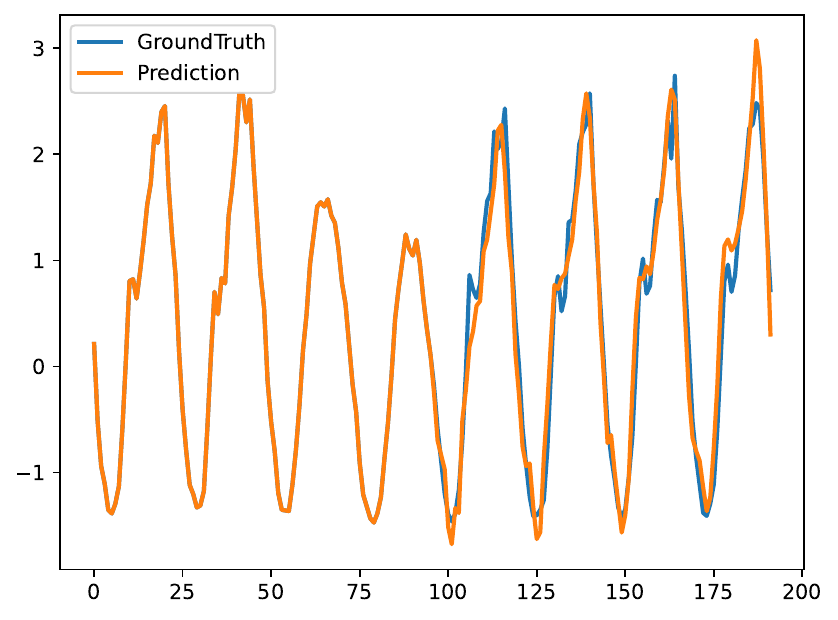}
\caption{Budget = 100}
\end{subfigure}

\begin{subfigure}{0.32\textwidth}
\includegraphics[width=\linewidth]{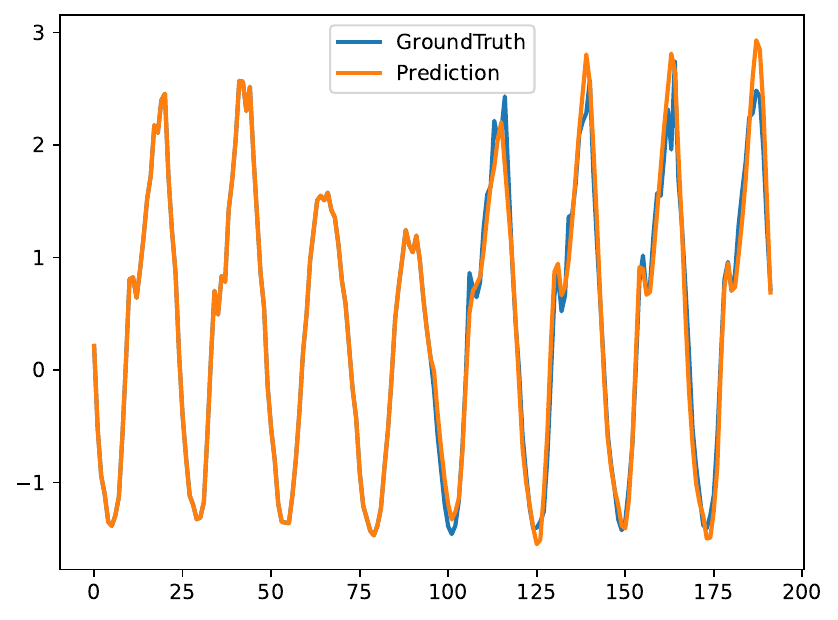}
\caption{Budget = 500}
\end{subfigure}\hfill
\begin{subfigure}{0.32\textwidth}
\includegraphics[width=\linewidth]{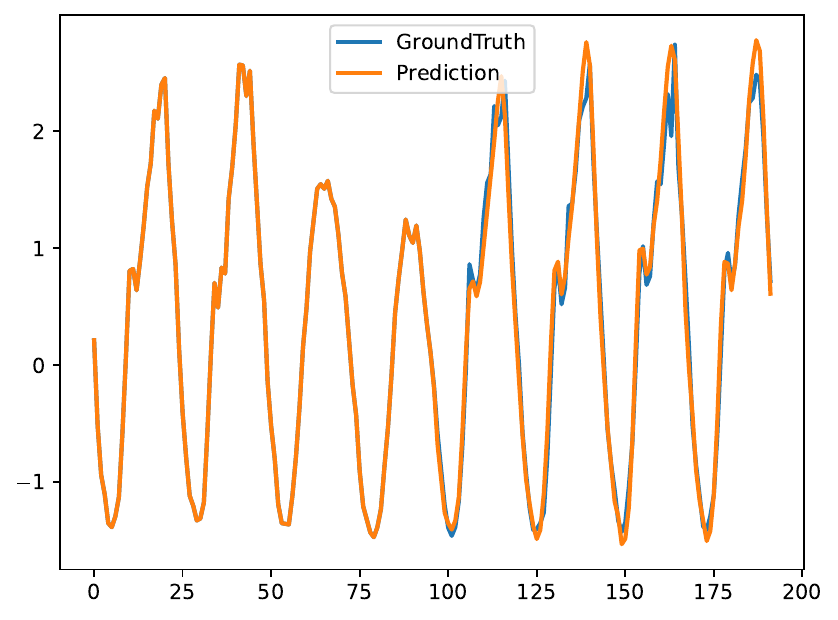}
\caption{Budget = 1000}
\end{subfigure}\hfill
\begin{subfigure}{0.32\textwidth}
\includegraphics[width=\linewidth]{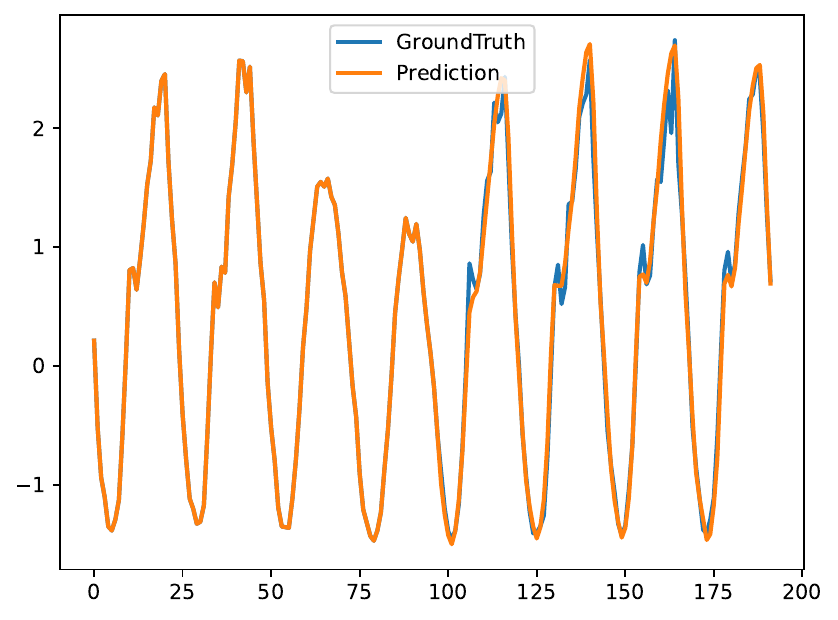}
\caption{Budget = All}
\end{subfigure}

\caption{The forecasts of TimePFN with various data budgets on traffic dataset. }

\label{fig:traffic0}
\end{figure*}
\begin{figure*}[htp]
\centering

\begin{subfigure}{0.32\textwidth}
\includegraphics[width=\linewidth]{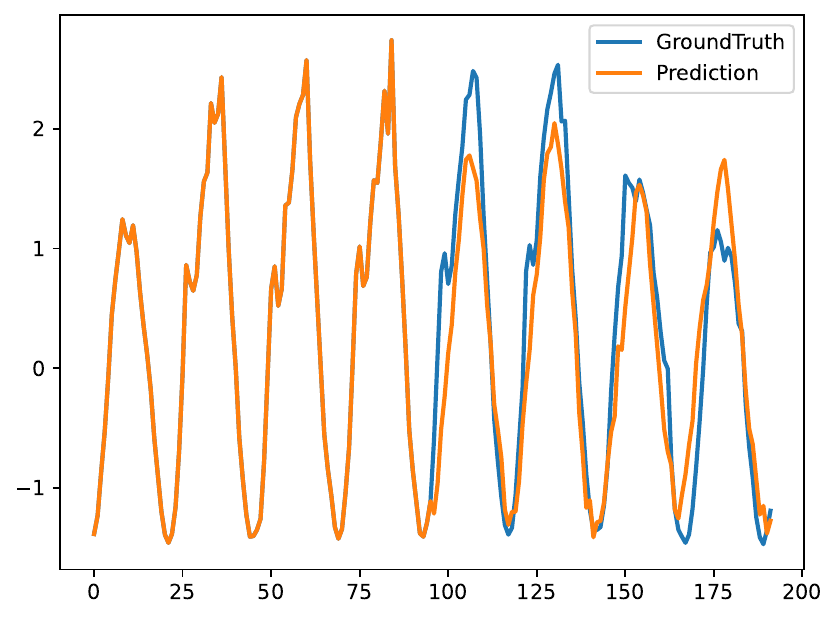}
\caption{Zero Shot}
\end{subfigure}\hfill
\begin{subfigure}{0.32\textwidth}
\includegraphics[width=\linewidth]{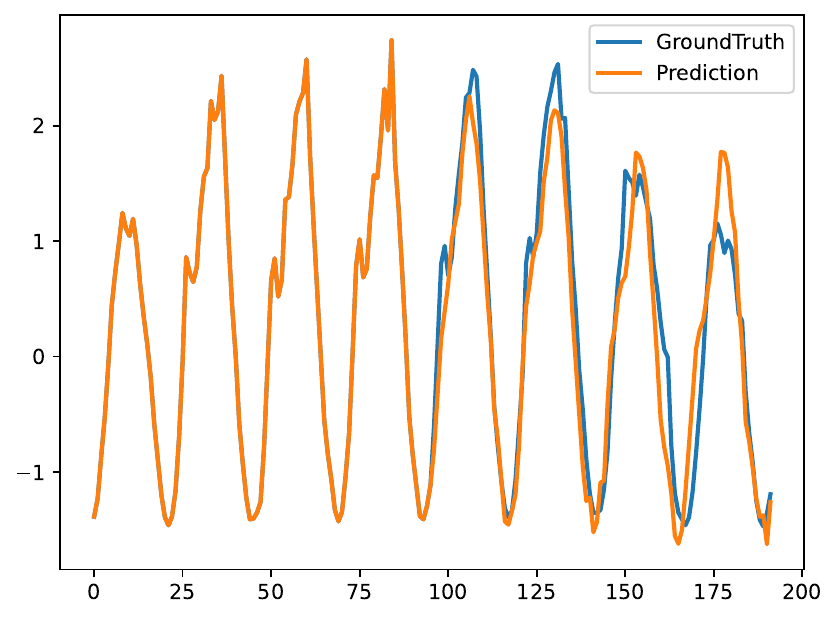}
\caption{Budget = 50}
\end{subfigure}\hfill
\begin{subfigure}{0.32\textwidth}
\includegraphics[width=\linewidth]{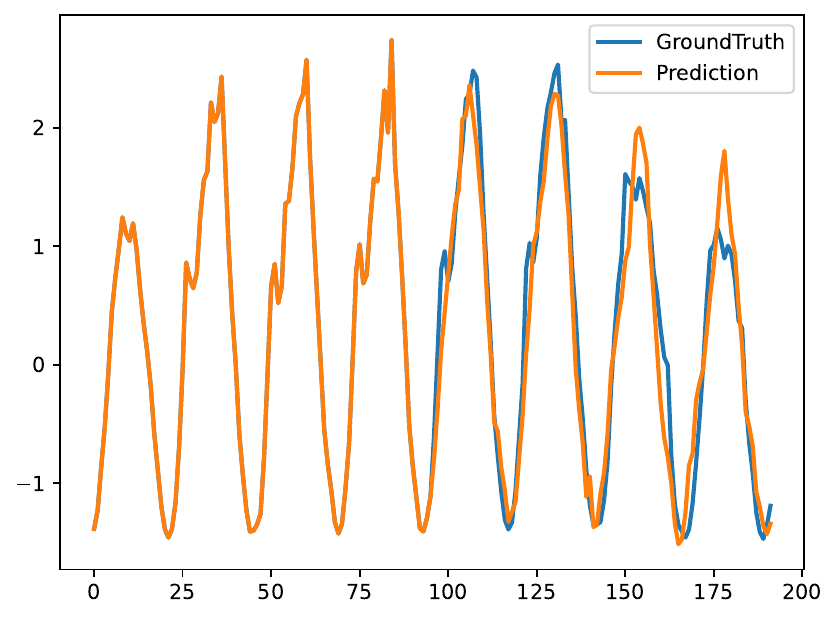}
\caption{Budget = 100}
\end{subfigure}

\begin{subfigure}{0.32\textwidth}
\includegraphics[width=\linewidth]{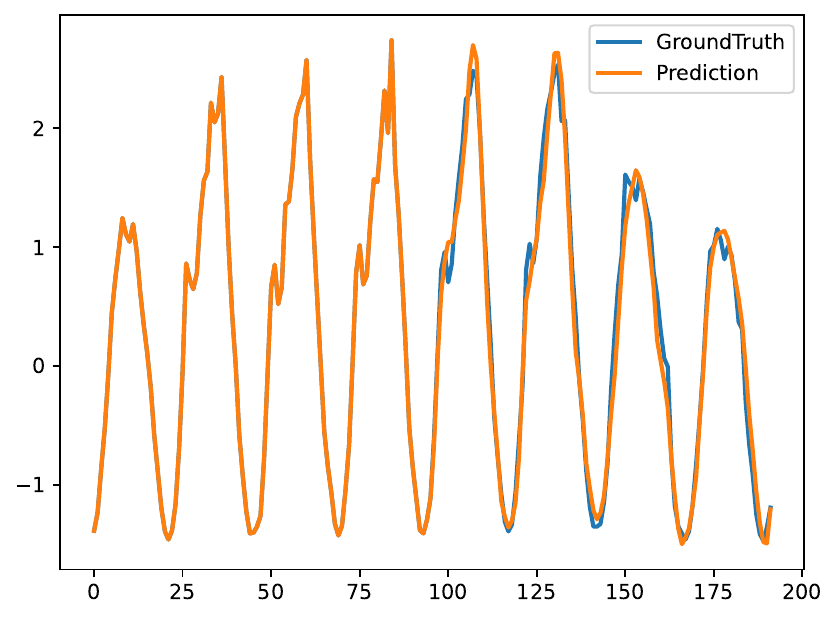}
\caption{Budget = 500}
\end{subfigure}\hfill
\begin{subfigure}{0.32\textwidth}
\includegraphics[width=\linewidth]{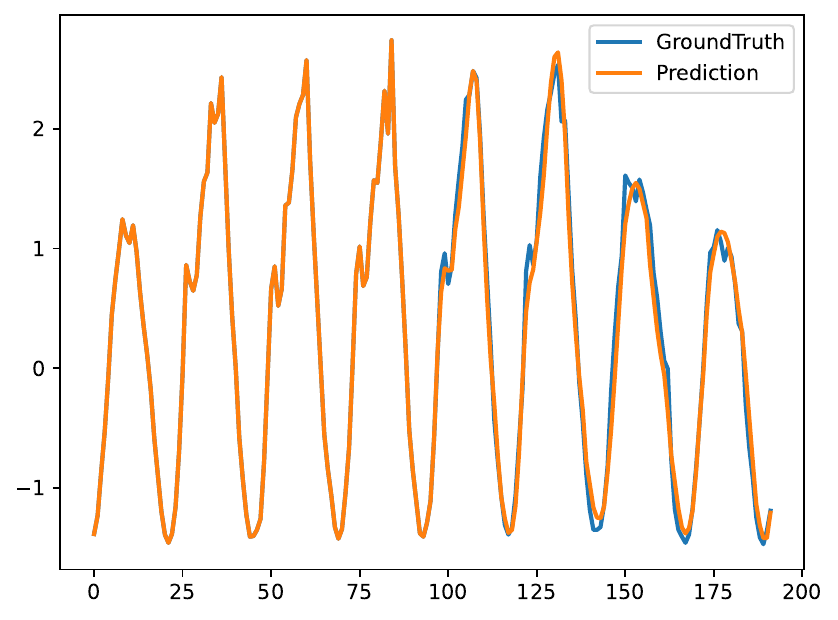}
\caption{Budget = 1000}
\end{subfigure}\hfill
\begin{subfigure}{0.32\textwidth}
\includegraphics[width=\linewidth]{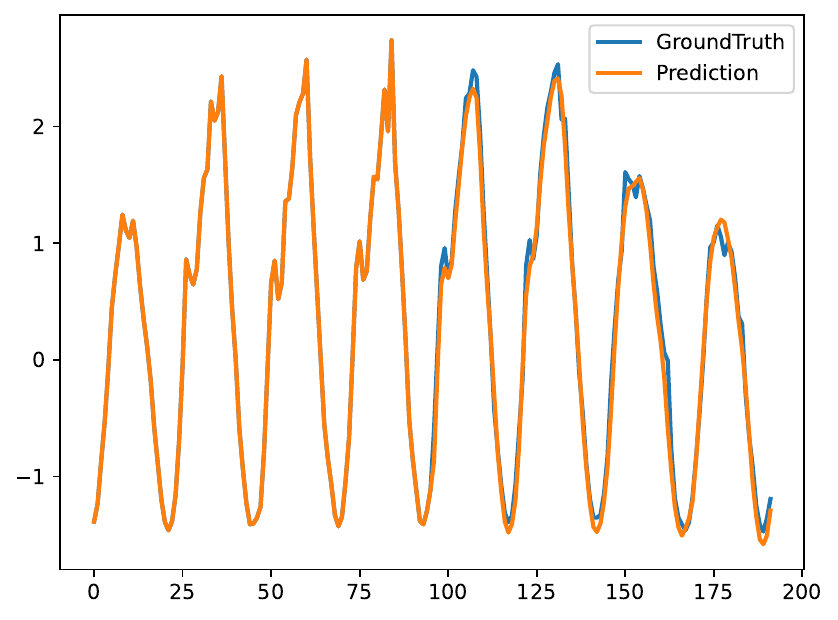}
\caption{Budget = All}
\end{subfigure}

\caption{The forecasts of TimePFN with various data budgets on traffic dataset. }

\label{fig:traffic80}
\end{figure*}

\end{document}